\documentclass{article} %
\usepackage{iclr2027_conference,times}

\usepackage{amsmath,amsfonts,bm}

\def\eqref#1{equation~\ref{#1}}

\def\1{\bm{1}}

\DeclareMathAlphabet{\mathsfit}{\encodingdefault}{\sfdefault}{m}{sl}
\SetMathAlphabet{\mathsfit}{bold}{\encodingdefault}{\sfdefault}{bx}{n}

\usepackage{hyperref}
\usepackage{url}

\usepackage[utf8]{inputenc} %
\usepackage[T1]{fontenc}    %
\usepackage{hyperref}       %
\usepackage{url}            %
\usepackage{booktabs}       %
\usepackage{amsfonts}       %
\usepackage{nicefrac}       %
\usepackage{microtype}      %
\usepackage{xcolor}         %

\usepackage[normalem]{ulem}
\usepackage{amsmath}
\usepackage{amssymb}
\usepackage[capitalize,noabbrev]{cleveref}
\usepackage{tabularx}
\usepackage{pifont}%
\newcommand{\cmark}{\textcolor{black!80}{\ding{51}}}%
\newcommand{\xmark}{\textcolor{gray!60}{\ding{55}}}
\usepackage{multirow}
\usepackage{makecell}
\usepackage{sansmath}
\usepackage{siunitx}
\usepackage{wrapfig}
\usepackage{titletoc}
\usepackage[normalem]{ulem}
\usepackage{fix-cm}
\usepackage[symbol]{footmisc}

\usepackage{xspace}
\newcommand{\ours}{DUNE\xspace}

\newcommand{\eg}{\emph{e.g.}}
\newcommand{\Eg}{\emph{E.g.}}
\newcommand{\ie}{\emph{i.e.}}
\newcommand{\cf}{\emph{cf.}}
\newcommand{\vsiclr}{\emph{vs.}}

\newcommand{\tablestd}[1]{\tiny\mdseries\color{gray}{\ \textpm\ \num{#1}}}

\newcommand{\colorindicator}[1]{%
	{\textcolor{#1}{$\blacksquare\!\!\!\!\!\blacksquare$}}%
}

\usepackage{tikz}
\usetikzlibrary{positioning}
\usetikzlibrary{backgrounds,calc}
\usetikzlibrary{shapes.geometric}
\usetikzlibrary{calc,shapes.symbols}

\usepackage{pgfplots}
\pgfplotsset{compat=1.18} %
\usepgfplotslibrary{statistics}

\usetikzlibrary{fit}
\usetikzlibrary{matrix}

\usetikzlibrary{shapes.misc}

\usepackage{xcolor}
\definecolor{myblue}{HTML}{648FFF}
\definecolor{mypurple}{HTML}{785EF0}
\definecolor{mypink}{HTML}{DC267F}
\definecolor{myorange}{HTML}{FE6100}
\definecolor{myyellow}{HTML}{FFB000}
\definecolor{lightgray}{gray}{0.92}

\DeclareRobustCommand{\colorcircle}[1]{\tikz\draw[#1, fill=#1] (0,0) circle (0.5ex);}
\DeclareRobustCommand{\colorsquare}[1]{%
  \tikz[baseline=-0.5ex]\node[rectangle,fill=#1,inner sep=0pt,
    minimum size=1ex] {};%
}
\DeclareRobustCommand{\colordiamond}[1]{%
  \tikz[baseline=-0.5ex]\node[diamond,fill=#1,inner sep=0pt,
    minimum size=1ex] {};%
}
\DeclareRobustCommand{\colortriangle}[1]{%
  \tikz[baseline=-0.5ex]\node[regular polygon,regular polygon sides=3,
    fill=#1,inner sep=0pt,minimum size=1ex] {};%
}

\DeclareRobustCommand{\colorcirclecross}[2]{%
  \tikz[baseline=-0.5ex]{
    \draw[#1, fill=#1] (0,0) circle (0.5ex);
    \draw[#2, line width=0.6pt, line cap=round]
  (-0.25ex,-0.25ex) -- (0.25ex,0.25ex)
  (-0.25ex,0.25ex) -- (0.25ex,-0.25ex);
  }
}

\usetikzlibrary{shapes.arrows, shapes.geometric, decorations.pathreplacing}

\newcommand{\myparagraph}[1]{\vspace{3pt}\noindent{\bf #1}}

\title{\ours: Disentangling Polysemantic Neurons Without Losing Predictive Performance}

\newcommand{\myand}{\hspace{0.5em}}

\author{%
  Do\u{g}ukan Ba\u{g}c\i\textsuperscript{\textnormal{\,1}}\myand
  Bernt Schiele\textsuperscript{\textnormal{ 1}} \myand
  Simone Schaub-Meyer\textsuperscript{\textnormal{\,2\,3}} \myand
  Jonas Fischer\textsuperscript{\textnormal{\,1}} \myand
  Robin Hesse\textsuperscript{\textnormal{\,1}} \\
    \and 
  {\small
  \textsuperscript{1}Max Planck Institute for Informatics, SIC \hspace{0.25em}
  \textsuperscript{2}Department of Computer Science, TU Darmstadt \hspace{0.25em}
  \textsuperscript{3}hessian.AI
} \\
}

\iclrfinalcopy %
\begin{document}

\maketitle\lhead{}

\begin{abstract}
Deep neural networks (DNNs) are widely used, but understanding what they actually learn remains difficult. A major obstacle is that individual neurons often respond to multiple unrelated concepts, obscuring the network's internal decision process. Existing approaches, such as sparse autoencoders, can disentangle these mixed signals into more meaningful, ``monosemantic'' features, but typically introduce approximation errors that may degrade downstream performance. Consequently, these methods exhibit a Pareto frontier between monosemanticity and functional faithfulness.
To address this, we introduce \textbf{\ours} (\textbf{d}isentangling \textbf{u}nits with \textbf{n}o \textbf{e}rror), a method that decomposes DNN neurons into more monosemantic subunits while preserving their original function by design. \ours identifies concept-specific contributions from the preceding layer and re-routes them into separate subunits whose aggregation exactly recovers the original activation. \ours can be applied directly to pretrained models, without labels or training of an auxiliary model.
Across several models, including DINOv2, \ours achieves state-of-the-art or competitive monosemanticity while exactly preserving functional faithfulness, substantially advancing the existing Pareto frontier between these properties. \ours operates efficiently and enables fine-grained interventions such as concept-specific steering.
\end{abstract}

\section{Introduction}\label{sec:intro}

\begin{wrapfigure}{r}{0.45\textwidth}
  \centering
  \vspace{-4.1em}
  \begin{sansmath}
\begin{tikzpicture}

\scriptsize
\sffamily

\begin{axis}[
    width=1\linewidth,
    height=0.75\linewidth,
    xmin=0.20, xmax=0.75,
    ymin=0.50, ymax=1.06,
    xlabel={Monosemanticity ($\uparrow$)},
    ylabel={Faithfulness ($\uparrow$)},
    xtick={0.2,0.3,0.4,0.5,0.6,0.7},
    ytick={0.50,0.60,0.70,0.80,0.90,1.00},
    label style={font=\scriptsize},
    axis line style={thick},
    enlargelimits=false,
    clip=false,
    axis background/.style={fill=gray!5,rounded corners=1pt},
    grid=both,
    major grid style={gray!30},
    minor grid style={gray!15},
    axis lines=left,
    trim axis left,
    trim axis right,
]

\addplot[
    myyellow,
     thick,
    opacity=0.8,
    no marks,
] coordinates {
    (0.6198,0.8365)
    (0.5156,0.8475)
    (0.4303,0.8526)
};

\addplot[
    only marks,
    scatter,
    scatter/use mapped color={draw=myyellow,fill=myyellow},
    mark=*,
    draw=myyellow,
    fill=myyellow,
    opacity=0.8,
    visualization depends on={\thisrow{s}\as\perpointmarksize},
    scatter/@pre marker code/.append style={
        /tikz/mark size=\perpointmarksize
    },
] table {
x       y       s
0.6198 0.8365 0.5
0.5156 0.8475 1.3
0.4303 0.8526 2.1
};

\node[myyellow, opacity=0.8, anchor=north] at (axis cs:0.4303,0.8526) [yshift=0pt] {ASAE};

\addplot[
    myyellow,
    very thick,
    opacity=0.8,
    no marks,
] coordinates {
    (0.5636,0.8758)
    (0.5214,0.8871)
    (0.4748,0.8938)
};

\addplot[
    only marks,
    scatter,
    scatter/use mapped color={draw=myyellow,fill=myyellow},
    mark=*,
    draw=myyellow,
    fill=myyellow,
    opacity=0.8,
    visualization depends on={\thisrow{s}\as\perpointmarksize},
    scatter/@pre marker code/.append style={
        /tikz/mark size=\perpointmarksize
    },
] table {
x       y       s
0.5636 0.8758 0.5
0.5214 0.8871 1.3
0.4748 0.8938 2.1
};

\node[myyellow, opacity=0.8, anchor=west] at (axis cs:0.5636,0.8758) [xshift=-14pt, yshift=-3pt]
{MSAE};

\addplot[
    myyellow,
    very thick,
    opacity=0.8,
    no marks,
] coordinates {
    (0.5367,0.8796)
    (0.5064,0.8915)
    (0.4413,0.8965)
};

\addplot[
    only marks,
    scatter,
    scatter/use mapped color={draw=myyellow,fill=myyellow},
    mark=*,
    draw=myyellow,
    fill=myyellow,
    opacity=0.8,
    visualization depends on={\thisrow{s}\as\perpointmarksize},
    scatter/@pre marker code/.append style={
        /tikz/mark size=\perpointmarksize
    },
] table {
x       y       s
0.5367 0.8796 0.5
0.5064 0.8915 1.3
0.4413 0.8965 2.1
};

\node[myyellow, opacity=0.8, anchor=east] at (axis cs:0.4413,0.8965) [xshift=0pt]
{BatchTopK};

\addplot[
    myyellow,
    very thick,
    opacity=0.8,
    no marks,
] coordinates {
    (0.5840,0.8877)
    (0.5692,0.9028)
    (0.5535,0.9137)
};

\addplot[
    only marks,
    scatter,
    scatter/use mapped color={draw=myyellow,fill=myyellow},
    mark=*,
    draw=myyellow,
    fill=myyellow,
    opacity=0.8,
    visualization depends on={\thisrow{s}\as\perpointmarksize},
    scatter/@pre marker code/.append style={
        /tikz/mark size=\perpointmarksize
    },
] table {
x       y       s
0.5840 0.8877 0.5
0.5692 0.9028 1.3
0.5535 0.9137 2.1
};

\node[myyellow, opacity=0.8, anchor=east] at (axis cs:0.5535,0.9137) [xshift=-4pt, yshift=2pt]
{MpSAE};

\addplot[
    myyellow,
    very thick,
    opacity=0.8,
    no marks,
] coordinates {
    (0.3295,0.8417)
    (0.2750,0.9288)
    (0.2338,1)
};

\addplot[
    only marks,
    scatter,
    scatter/use mapped color={draw=myyellow,fill=myyellow},
    mark=*,
    draw=myyellow,
    fill=myyellow,
    opacity=0.8,
    visualization depends on={\thisrow{s}\as\perpointmarksize},
    scatter/@pre marker code/.append style={
        /tikz/mark size=\perpointmarksize
    },
] table {
x       y       s
0.3295 0.8417 0.25
0.2750 0.9288 0.65
0.2338 1 1.05
};

\node[myyellow, opacity=0.8, anchor=east] at (axis cs:0.2750,0.9288) 
{PCA};

\addplot[
    myyellow,
    very thick,
    opacity=0.8,
    no marks,
] coordinates {
    (0.7328,0.5222)
    (0.5118,0.5717)
    (0.2567,0.7945)
};

\addplot[
    only marks,
    scatter,
    scatter/use mapped color={draw=myyellow,fill=myyellow},
    mark=*,
    draw=myyellow,
    fill=myyellow,
    opacity=0.8,
    visualization depends on={\thisrow{s}\as\perpointmarksize},
    scatter/@pre marker code/.append style={
        /tikz/mark size=\perpointmarksize
    },
] table {
x       y       s
0.7328 0.5222 0.5
0.5118 0.5717 1.3
0.2567 0.7945 2.1
};

\node[myyellow, opacity=0.8, anchor=south west] at (axis cs:0.5118,0.5717) [xshift=-4pt]
{SemiNMF};

\addplot[
    only marks,
    mark=square*,
    mark options={fill=white,draw=black,thick},
    black,
] coordinates {(0.3620,1.0)};

\node[black, opacity=0.8, anchor=south,yshift=1pt] at (axis cs:0.3620,1.0) {DINOv2-B};

\addplot[
    mypink,
    opacity=0.8,
    very thick,
    no marks,
] coordinates {
    (0.6272,1.0000)
    (0.6485,1.0000)
    (0.6677,1.0000)
};

\addplot[
    only marks,
    scatter,
    scatter/use mapped color={draw=mypink,fill=mypink},
    mark=*,
    draw=mypink,
    fill=mypink,
    opacity=0.8,
    visualization depends on={\thisrow{s}\as\perpointmarksize},
    scatter/@pre marker code/.append style={
        /tikz/mark size=\perpointmarksize
    },
] table {
x       y       s
0.6272 1.0000 0.5
0.6485 1.0000 1.3
0.6748 1.0000 2.1
};

\node[mypink, opacity=0.8, anchor=south] at (axis cs:0.6399,1.0000) [yshift=0pt,xshift=-5pt] {\ours \textit{(ours)}};

\addplot[
    gray!70!black,
    densely dashed,
    line width=0.5pt,
    smooth,
    tension=0.6,
    opacity=0.8,
    no marks,
] coordinates {
    (0.3620,1.0000)
    (0.5535,0.9137)
    (0.5692,0.9028)
    (0.5840,0.8877)
    (0.6198,0.8365)
    (0.7328, 0.5222) 
};
\node[gray!70!black, opacity=0.8, anchor=south, rotate=-29] at (axis cs:0.47,0.963) [yshift=2pt, xshift=35pt] {existing Pareto front};

\end{axis}

\pgfresetboundingbox
\path ([xshift=-20pt,yshift=-20pt]current axis.south west) rectangle (current axis.north east);

\end{tikzpicture}
\end{sansmath}
  
  \caption{\textbf{Monosemanticity--faithfulness tradeoff.} We compare faithfulness and monosemanticity across disentanglement methods; marker size indicates the expansion factor. Existing approaches exhibit a clear Pareto front, where higher monosemanticity typically comes at the cost of lower faithfulness. In contrast, \ours improves monosemanticity without sacrificing faithfulness, substantially advancing the Pareto front. See \cref{sec:experiments:interp_faith} for experimental details and detailed descriptions of the metrics.}
  \label{fig:tradeoff}
  \vspace{-1.5em}
\end{wrapfigure}
Deep neural networks (DNNs) have achieved remarkable performance across a wide range of domains, from language~\citep{Brown:2020:LMF} over vision~\citep{Oquab:2024:DLR} to scientific modeling~\citep{Jumper:2021:HAP}. Despite their success, these models remain difficult to understand and control, posing significant challenges for their deployment in safety-critical applications~\citep{Bommasani:2021:OTO}. Issues such as hallucinations~\citep{Ji:2023:SOH,Liu:2024:ASO}, unsafe outputs~\citep{Wang:2023:DNA, Zhang:2024:TGO}, and failures under distribution shift~\citep{Hendrycks:2021:TMF, Hesse:2026:BAW} highlight the need for methods that provide insight into model internals and enable reliable intervention~\citep{Bereska:2024:MIF,Sharkey:2925:OPM}.

A central goal in this context is \emph{interpretability}: understanding how models represent and process information. However, DNNs are complex systems, where individual units (\eg, neurons or channels) do not correspond cleanly to human-interpretable concepts. Instead, they often exhibit \emph{polysemanticity}, responding to multiple unrelated concepts~\citep{Elhage:2022:TMO}. This obscures the internal structure of representations and makes it difficult to attribute model behavior to specific components.

To overcome this limitation, recent work has explored disentangling internal representations into more interpretable components, aiming for \emph{monosemanticity}, where individual units correspond to distinct semantic concepts. Among these, sparse autoencoders (SAEs)~\citep{Bricken:2023:TMD, Huben:2024:SAF} and transcoders~\citep{Dunefsky:2024:TFI} have emerged as particularly prominent approaches, decomposing internal representations into sparse, interpretable features.

However, many recent approaches exhibit a practical tradeoff between monosemanticity and faithfulness, as visualized in \cref{fig:tradeoff}. As these methods rely on \emph{learned} decompositions, they may introduce approximation errors and thus fail to exactly preserve the original model's behavior, limiting their applicability~\citep{Gao:2025:SAE}. In contrast, approaches that preserve functional faithfulness typically offer limited improvements in monosemanticity, do not yield explicit disentanglements that support fine-grained manipulation~\citep{OMahony:2023:DNR,Dreyer:2024:PTP}, or require supervised training~\citep{Hesse:2025:DPC}, limiting their applicability or scalability.

\myparagraph{Contributions.} In this work, we introduce \emph{\ours} (\textbf{d}isentangling \textbf{u}nits with \textbf{n}o \textbf{e}rror), a method that mitigates this tradeoff between monosemanticity and faithfulness by achieving an explicit, lossless, and unsupervised disentanglement. Specifically, \ours produces monosemantic and directly manipulable units \emph{(explicit)}, exactly preserves the original model's behavior \emph{(lossless)}, and requires no annotations \emph{(unsupervised)}. %
It identifies monosemantic components within polysemantic units by uncovering structure in previous-layer attributions and restructuring model connections to isolate these components into independent, function-preserving subunits (\cf~\cref{fig:teaser}). %
Due to its simplicity and lossless nature, \ours can be applied post-hoc to a wide range of models, effectively providing improved monosemanticity without sacrificing downstream predictive performance.

Empirically, \ours achieves state-of-the-art or competitive monosemanticity across a wide range of models and settings, substantially advancing the monosemanticity--faithfulness Pareto front (\cf~\cref{fig:tradeoff}). We further show that its explicit, lossless decomposition enables fine-grained interventions, which we demonstrate through model steering in LLaVA-OneVision~\citep{Li:2025:LOE}.

\begin{figure}[t!]
  \centering
  \input{figures/teaser_new.tex}
  \vspace{-1.75em}
  \caption{\textbf{\ours for monosemantic disentanglement.} \ours decomposes a polysemantic unit at layer $L$ into more monosemantic subunits by restructuring incoming weights so that each sub-unit captures only a specific semantic concept. The sum of all subunits exactly recovers the original unit activation, ensuring perfect faithfulness. On the right, we provide a concrete example of this process by showing highly activating images for an actual neuron from the last layer of a DINOv2 model alongside its disentangled \ours subunits. While the original neuron reacts to multiple semantically unrelated images, the disentangled units are more coherent and interpretable.}
  \label{fig:teaser}
\end{figure} 

\section{Related Work}\label{sec:related_work}

\begin{table}[t]
\scriptsize
\centering
\caption{\textbf{Properties of existing disentanglement approaches.} We compare \ours (shaded) against representative SAE variants and prior disentanglement methods across five properties: \emph{stable} (gives a similar disentanglement on repeated runs), \emph{explicit} (subunits can be inspected and modified directly, rather than only analyzed), \emph{lossless} (the disentangled model produces exactly the same outputs as the original), \emph{unsupervised} (no class labels needed) and \emph{effective disentanglement} (produces subunits that are more semantically coherent than the original units). \ours satisfies all five properties. 
} 
\label{tab:related_work}
\tabcolsep = 8.0pt
\begin{tabularx}{\linewidth}{@{}X c  c c c c@{}}
\toprule
{\bfseries Method} &
{\bfseries Stable} &
{\bfseries Explicit} &
{\bfseries Lossless} &
{\bfseries Unsupervised} &
{\bfseries Effective Disentanglement} \\
\midrule
PCA~\citep{Pearson:1901:OLA} & \cmark &   \cmark & \cmark & \cmark & \xmark \\ 
Semi-NMF~\citep{Ding:2010:CAS} & \xmark  &  \cmark & \xmark & \cmark & \cmark \\ 
Vanilla SAE~\citep{Bricken:2023:TMD} & \xmark &  \cmark & \xmark & \cmark & \cmark \\
Archetypal SAE~\citep{Fel:2025:ASA} & \cmark & \cmark & \xmark & \cmark & \cmark \\
DNRwCV~\citep{OMahony:2023:DNR} & \cmark &  \xmark & \cmark & \cmark & \cmark \\
PURE~\citep{Dreyer:2024:PTP} & \cmark & \xmark & \cmark & \cmark & \cmark \\
DPCiCNN~\citep{Hesse:2025:DPC} & \cmark & \cmark & \cmark & \xmark  & \cmark\\
\cmidrule(){1-6}
\tikz[remember picture,baseline] \node[inner sep=0pt,outer sep=0pt,minimum size=0pt] (row2beg) {};\ours \emph{(ours)}& \cmark  &  \cmark & \cmark & \cmark & \cmark {\rlap{\tikz[remember picture,baseline] \node[inner sep=0pt,outer sep=0pt,minimum size=0pt] (row2end) {};}}\\
\bottomrule
\end{tabularx}
\begin{tikzpicture}[remember picture,overlay]
  \begin{scope}[on background layer]
      \fill[black,opacity=0.10]
        ([xshift=0,yshift=2.7ex]row2beg.north west)
        rectangle
        ([xshift=11.6ex,yshift=-1.4ex]row2end.south east);
    \end{scope}
\end{tikzpicture}
\end{table}

In the pursuit of gaining a deeper model understanding, \emph{mechanistic interpretability}~\citep{Bereska:2024:MIF, Sharkey:2925:OPM, Rai:2024:APR}
aims to reverse-engineer the internal computations of neural networks. %
By exposing the units, directions, and circuits that carry meaningful computation, it promises a level of understanding that even supports targeted intervention, such as \emph{steering}~\citep{Turner:2023:SLM, Templeton:2024:SME, Pach:2025:SAL}.
However, a central challenge in mechanistic interpretability is \emph{polysemanticity}~\citep{Elhage:2022:TMO, Olah:2020:ZIA, Huben:2024:SAF}: individual units of modern neural networks frequently respond to multiple, semantically unrelated concepts rather than to a single coherent concept. This entanglement, often attributed to the superposition of more concepts than the layer has dimensions to allocate~\citep{Elhage:2022:TMO}, %
makes it difficult to attribute, explain, or steer model behavior at the level of individual units.

Recent work thus addresses polysemanticity by decomposing internal activations into more interpretable components. \emph{Linear decomposition methods}, such as PCA, NMF, and Semi-NMF, decompose internal activations into linear components that can expose more monosemantic structure in the representation~\citep{Fel:2023:CCR, Collins:2018:DFF,Zhang:2021:ICE, Fel:2025:ASA}.
\emph{Sparse autoencoders} (SAEs)~\citep{Bricken:2023:TMD} train an overcomplete autoencoder to reconstruct activations as sparse combinations of learned features, and \emph{transcoders}~\citep{Dunefsky:2024:TFI} extend this idea to mappings between layers. Both produce an explicit disentanglement and are unsupervised but introduce a learned approximation that does not exactly preserve the original model's behavior. 
\emph{DNRwCV}~\citep{OMahony:2023:DNR} and its successor \emph{PURE}~\citep{Dreyer:2024:PTP} instead disentangle polysemantic units \emph{``virtually''}: DNRwCV by finding concept-aligned directions in the activation space and PURE by clustering circuits. While both keep the model unchanged, they only yield ``virtual'' decompositions rather than explicit ones; hence, many applications such as feature-visualization and steering cannot be applied directly on the disentangled representations. Finally, \emph{DPCiCNN}~\citep{Hesse:2025:DPC} restructures the parameters of convolutional channels to produce explicit, function-preserving subunits, but requires supervision, introduces an uninterpretable residual neuron, and has so far been demonstrated only on convolutional architectures. \Cref{tab:related_work} summarizes these approaches across the axes of being stable, explicit, lossless, unsupervised, and effective disentanglement. 

To conclude, many existing approaches enable explicit and scalable unsupervised disentanglement, but incur information loss. In contrast, lossless disentanglement methods either lack an explicit disentanglement, achieve only mediocre monosemanticity, or do not scale to large vision foundation models, as they depend on supervision. Our method bridges this gap by providing an explicit, lossless, and unsupervised disentanglement while achieving strong monosemanticity.

\section{\ours: Disentangling Units with No Error}\label{sec:method}

Given a probing dataset of $N$ images $\mathcal{D} = \{\mathbf{x}^{(i)}\}_{i=1}^N$ without annotations, a pre-trained vision backbone $f(\cdot)$, and a layer of interest $L$, our goal is to improve monosemanticity by disentangling polysemantic activation units $a_u$ (\eg, channels or neurons) into multiple, more monosemantic units, as previewed in \cref{fig:teaser}. Crucially, this is achieved without altering the overall function of $f(\cdot)$. %

Our method decomposes a potentially polysemantic unit into a set of more interpretable subunits whose weights form an additive partition of the original unit, ensuring that their aggregated activation is exactly equal to the original (\cf~\cref{fig:teaser}). By leveraging contribution structure in the preceding layer, we first identify inputs that strongly activate the unit to characterize its behavior. We then analyze the corresponding contribution patterns in layer $L{-}1$ to uncover distinct groups of upstream units associated with different concepts. Using this structure, we disentangle the original unit into multiple subunits, each selectively connected to a concept-specific group of upstream units. Finally, we merge these subunits such that their aggregation exactly reconstructs the original unit’s activation, preserving the model’s function by construction.

More specifically, our method consists of four steps, which are illustrated in \cref{fig:method_overview}:
\textit{(i)} collecting the top-$k$ activation instances $\mathcal{D}_u$ from $\mathcal{D}$ for unit $a_u$ that expose the concepts driving its activation (\cref{sec:method:1});
\textit{(ii)} clustering contribution patterns in layer $L{-}1$ to identify distinct concepts (\cref{sec:method:2});
\textit{(iii)} restructuring the weights between layers $L{-}1$ and $L$ to produce a new layer $L^{*}$ in which $a_u$ is replaced by a set of subunits, each primarily connected to a concept-specific group of upstream units (\cref{sec:method:3}); and
\textit{(iv)} merging the subunits to recover the original unit $a_u$ (\cref{sec:method:4}).

\begin{figure}[t]
    \centering
    \resizebox{\linewidth}{!}{\input{figures/method_overview}}
    \vspace{-0.5em}
    \caption{
    \textbf{Overview of \ours.}
    \textit{(1)}~Embed every image $\mathbf{x}^{(i)}$ in the dataset $\mathcal{D}$ through layer $L$ and consider the resulting activation instances $\ell$ (\eg, spatial locations or tokens); select the top-$k$ instances that activate unit $a_u$ most strongly.
    \textit{(2)}~For each selected activation instance, compute a normalized contribution vector ${\mathbf{c}}^{(\ell)}$ from layer $L{-}1$ to $a_u^{(\ell)}$, revealing which upstream units drive the activation. Cluster these contribution vectors into concept groups, yielding $K$ clusters represented by $\tilde{\mathbf{r}}^{(u,c)}$, one per concept encoded by $a_u$.
    \textit{(3)}~Replace $a_u$ with $K$ concept-specific units whose weights are constructed from the cluster representatives.
    \textit{(4)}~A summation operator reconstructs the original activation exactly.
}
    \label{fig:method_overview}
\end{figure}

\subsection{Collection of the top-$k$ activation instances}\label{sec:method:1}

To disentangle a unit of interest $a_u$ in layer $L$, we first need to find the set of image patterns (respectively concepts) $\mathcal{D}_u$ for which it activates. To this end, we collect the top-$k$ activation instances from a probing dataset $\mathcal{D}$ that activate $a_u$ most strongly, as illustrated in \cref{fig:method_overview}~\emph{(1)}. %
Depending on the architecture, an activation instance is either a spatial location in a CNN feature map or a token in a Vision Transformers (ViT).
The choice of the probing dataset $\mathcal{D}$ is task- and user-specific and defines the potential concepts that can be extracted; a common choice in related work is, \eg, the ImageNet-1k dataset~\citep{Russakovsky:2015:ILS,Dreyer:2024:PTP}.

Formally, we index all activation instances across the dataset by a single index $\ell \in \{1, \dots, M\}$, where each $\ell$ corresponds to a specific pair $(i, p)$ of image index $i$ and spatial location or token index $p$; $M$ is the total number of activation instances across the dataset. Let $\mathbf{a}^{(\ell)} \in \mathbb{R}^{d_L}$ denote the corresponding activation vector at target layer $L$, and let $a_u^{(\ell)}$ be the activation of unit $u$ for that instance. We then select the top-$k$ activation instances
$\mathcal{D}_u =
\left\{
\ell \;\middle|\;
a_u^{(\ell)} \in \mathrm{TopK}\!\left(\{a_u^{(\ell')}\}_{\ell'},\, k\right)
\right\}$.
Each $\ell \in \mathcal{D}_u$ is associated with an underlying image $\mathbf{x}^{(i)}$, which provides semantic context. Higher values of $k$ include a broader range of activation strengths and may capture a wider variety of concepts encoded by the unit. Note that several tokens of the same image can appear in the top-$k$ activation instances since different regions of an image may represent multiple concepts. Such top-$k$ set constructions have been proven successful in prior interpretability research. \Eg, PURE~\citep{Dreyer:2024:PTP} and \citet{OMahony:2023:DNR} both build similar reference sets before clustering, and maximally activating images are a common starting point in feature visualization more broadly~\citep{Olah:2017:FVZ}. %

The following steps leverage that for polysemantic units different concepts will appear among the image regions that activate it most strongly. For example, as in \cref{fig:teaser}, a unit might respond to image regions of dogs, violins, and oceans -- visualized in yellow \colorindicator{myyellow}, orange \colorindicator{myorange}, and pink \colorindicator{mypink} in \cref{fig:method_overview}.

\subsection{Unsupervised clustering of the contribution patterns}\label{sec:method:2}
Given a polysemantic unit, we now aim to identify the distinct concepts present in $\mathcal{D}_u$. To this end, we follow a simple assumption that has been established in prior work~\citep{Dreyer:2024:PTP,Hesse:2025:DPC}: if two activation instances corresponding to distinct concepts activate the same unit in layer $L$, then different upstream circuits should have been used by the model. Instead of looking at the full circuits, common practice is to only consider the part of the circuit in layer $L{-}1$~\citep{Dreyer:2024:PTP,Hesse:2025:DPC}. 
Hence, for each of the selected activation instances $\ell$, we now identify which units in the previous layer $L{-}1$ were responsible for activating unit $a_u^{(\ell)}$ of layer $L$.
Concretely, we compute a \emph{contribution vector} $\mathbf{c}^{(\ell)} \in \mathbb{R}^{d_{L{-}1}}$, where $d_{L{-}1}$ is the number of units in layer $L{-}1$ and each entry measures how much one of those units contributed to the pre-activation of $a_u^{(\ell)}$. %

Let $\mathbf{q}^{(\ell)} \in \mathbb{R}^{d_{L{-}1}}$ denote the activation vector at the preceding layer $L{-}1$.  Since the contribution of the activation vector $\mathbf{q}^{(\ell)} \in \mathbb{R}^{d_{L{-}1}}$ in layer $L{-}1$ to the pre-activation $\tilde{a}_u^{(\ell)}$ reduces to a linear transformation for convolutional layers or MLP layers in ViTs, \ie, $\tilde{a}_u^{(\ell)} = \mathbf{W}_u^\top \mathbf{q}^{(\ell)} + \mathbf{b}_u$, we can compute the exact contribution via %
$\mathbf{c}^{(\ell)} =  \mathbf{W}_u \odot \mathbf{q}^{(\ell)}$ where $\odot$ denotes element-wise multiplication. %
We then $L_2$-normalize each contribution vector from all $\ell \in \mathcal{D}_u$ so that clustering in the next step focuses on \emph{which} upstream units were involved, rather than \emph{how strongly} they contributed. %
Doing this for all $k$ selected activation instances gives us the contribution set $\mathcal{C}^{(u)} = \{{\mathbf{c}}^{(\ell)} \mid \ell \in \mathcal{D}_u\}$ for unit $a_u$.

If unit $a_u$ is polysemantic, activation instances corresponding to different concepts will produce contribution vectors pointing in different directions, since they are driven by different upstream units.
Coming back to our example of a unit activating for image regions corresponding to dogs, violins, and oceans, we would expect three distinct contribution clusters corresponding to each of the concepts.
Hence, we next group the contribution vectors into clusters, with each cluster corresponding to one concept that unit $a_u$ responds to. %
To this end, we cluster $\mathcal{C}^{(u)}$ using HDBSCAN~\citep{Campello:2013:DBC}, a density-based clustering algorithm that finds groups of points that are tightly packed together while leaving sparse or ambiguous points unclustered. Crucially, in contrast to $k$-means clustering~\citep{Hartigan:1979:AKC}, used for example in PURE~\citep{Dreyer:2024:PTP}, HDBSCAN does not require specifying the number of clusters $K$ in advance. Instead, it infers $K$ directly from the data. It also assigns each point a membership probability $p({\mathbf{c}}^{(\ell)}) \in [0, 1]$, reflecting how confidently it belongs to its cluster.  Each point is assigned to the cluster for which it has the highest membership probability, and this probability weights its contribution to the cluster representative (we ablate the effect of using HDBSCAN instead of $k$-means in \cref{tab:cluster_ablation}).%

For each cluster index $c \in [1, \dots, K]$, we compute a single representative direction  $\tilde{\mathbf{r}}^{(u,c)}$  by taking a normalized membership-probability-weighted average of the contribution vectors in that cluster:
\begin{equation}
\label{eq:weighted_rep}
\tilde{\mathbf{r}}^{(u,c)} =
\frac{\mathbf{r}^{(u,c)}}{\|\mathbf{r}^{(u,c)}\|_2 + \epsilon}\ , \qquad \text{where} \qquad
\mathbf{r}^{(u,c)} =
\sum_{{\mathbf{c}}^{(\ell)} \in \mathcal{C}^{(u,c)}} p({\mathbf{c}}^{(\ell)})\,{\mathbf{c}}^{(\ell)}.
\end{equation}
The representative $\tilde{\mathbf{r}}^{(u,c)}$ summarizes which upstream units are most responsible for cluster (or concept) $c$, and will be used in the next section to construct the new concept-specific weights. The above steps are illustrated in \cref{fig:method_overview} \emph{(2)}.

\subsection{Disentangling through restructuring of the weights}\label{sec:method:3}
We now realize the discovered clusters as explicit subunits in a new layer $L^{*}$. The goal is to replace the polysemantic unit $a_u$ with $K$ new subunits $\{a_{u}^{(c)}\}_{c=1}^{K}$, one per concept, such that each new subunit responds selectively to its own concept. At the same time, we ensure that the behavior of the network remains unchanged, yielding a faithful decomposition.

Each concept subunit $a_{u}^{(c)}$ gets its own incoming weight vector $\mathbf{w}^{(u,c)}$ and bias $b_{u}^{(c)}$, constructed from the original weight vector $\mathbf{W}_u$, the original bias $b_u$, and the cluster representative $\tilde{\mathbf{r}}^{(u,c)}$.
Specifically, for each upstream dimension $j$ in layer $L{-}1$, we check whether one group of concept subunits clearly dominates the others in $\tilde{\mathbf{r}}^{(u,c)}_j$. If so, we assign that dimension exclusively to the winning group. If not, we treat it as shared and split the original weight equally across all concept subunits.

A set of concept indices $S \subseteq \{1, \dots, K\}$ wins dimension $j$ if the minimum score among its subunits exceeds the score of every concept subunit outside the set by a multiplicative margin $\rho
\in (0, 1]$:
\begin{equation}
    \rho \cdot \min_{c \in S} |\tilde{r}^{(u,c)}_j| > \max_{c' \notin S} |\tilde{r}^{(u,c')}_j|.
    \label{eq:winner_margin}
\end{equation}
The smallest set of top-scoring subunits $S$ that satisfies this condition is assigned dimension $j$ exclusively, and the corresponding weight is distributed equally among the subunits in $S$. If no set wins clearly, the weight is split equally across all $a_{u}^{(c)}$ concept subunits:
\begin{equation}
    w^{(u,c)}_j = \begin{cases}
        \dfrac{1}{|S|}\, W_{u,j} & \text{if } c \in S \text{ and } S \text{ wins dimension } j \text{ exclusively,} \\
        \dfrac{1}{K}\, W_{u,j} & \text{otherwise.}
    \end{cases}
    \label{eq:weight_construction}
\end{equation}
By construction, the concept subunit weights always sum back to the original:
$
    \sum_{c=1}^{K} w^{(u,c)}_j = W_{u,j},
$
and hence, for each upstream dimension $j$, subunit $c$ receives a fraction
$w^{(u,c)}_j / W_{u,j}$ of the original weight. We split the bias in
proportion to the average of these fractions across all $d_{L{-}1}$
dimensions: $b_u^{(c)} = \omega_c \, b_u$ with $\omega_c =
\tfrac{1}{d_{L{-}1}} \sum_j w^{(u,c)}_j / W_{u,j}$, so that
$\sum_{c=1}^{K} b_u^{(c)} = b_u$.

The margin $\rho$ controls how aggressively dimensions are assigned exclusively to a group. We select it automatically by searching over all values in $\{0.05, 0.10, \dots, 1.00\}$ and choosing the one for which each concept subunit activates most strongly on its own cluster and most weakly on the others (see \cref{tab:rho_distribution,tab:rho_ablation} for ablations). %

Returning to our illustrative example, we might find that the third unit from the top in the preceding layer $L{-}1$ (\colorindicator{mypink} in
\cref{fig:method_overview}) contributes particularly strongly on activation instances corresponding to oceans, and consequently, we connect it to the subunit corresponding to the ocean concept. Similarly, the fourth unit (bottom) in $L{-}1$ has no particularly high contribution to any activation instances, so we connect it to all the subunits. The process is visualized in \cref{fig:method_overview} \emph{(3)}.

\subsection{Merging}\label{sec:method:4}

Because the disentangled weights and biases form an additive partition of the originals, the activations of the concept subunits sum exactly to the pre-activation of the original unit, \ie, 
$\sum_{c=1}^{K} a_{u}^{(c)} = a_u$ for all inputs.
This allows us to directly reconstruct the original unit by summing the concept-specific subunits, ensuring that the overall function of the network remains unchanged, as illustrated in \cref{fig:method_overview} \emph{(4)}.

\section{Experiments}\label{sec:experiments}

We evaluate \ours, focusing on the monosemanticity--faithfulness tradeoff (\cref{sec:experiments:interp_faith}), model steering (\cref{sec:experiments:steering}), and practical considerations, such as stability and efficiency (\cref{sec:experiments:applications}).

\subsection{Monosemanticity and faithfulness}\label{sec:experiments:interp_faith}

\begin{table}[t]
  \caption{
    \textbf{Monosemanticity \vsiclr~faithfulness.} We compare \ours (shaded) against various linear, SAE, and transcoder baselines across three monosemanticity metrics %
    and two faithfulness metrics %
    on DINOv2-B/14 at expansion factor $8$. We report mean and standard deviation across three seeded token probing datasets. The best method is shown in \textbf{bold} and the second best is \underline{underlined}. \ours simultaneously improves monosemanticity while preserving faithfulness up to numerical precision (R$^2=100\%$ and matching the backbone's accuracy). \textdagger{} denotes an expansion factor of $0.25$.
  }
  \label{tab:interpretability_vs_faithfulness}
  \centering
  \setlength{\tabcolsep}{4.0pt}
  \scriptsize
  \begin{tabularx}{\linewidth}{@{}rXS[table-format=1]S[table-format=2.2(.2)]S[table-format=2.2(.1)]S[table-format=2.2(.1)]S[table-format=3.2(.2)]@{}l@{}}
    \toprule
    &  & \multicolumn{3}{c}{\bfseries Monosemanticity} & \multicolumn{2}{c}{\bfseries Faithfulness}\\
    \cmidrule(lr){3-5}\cmidrule(l){6-7}
    {\bfseries Model} & {\bfseries Method} & {\bfseries LLM Rank $(\downarrow)$} & {\bfseries C-Score $(\uparrow)$} & {\bfseries SIM@1 $(\uparrow)$} & {\bfseries Accuracy $(\uparrow)$} & {\bfseries R$^2$ $(\uparrow)$}\\
    \midrule
    \parbox[t]{2mm}{\multirow{17}{*}{\rotatebox[origin=c]{90}{DINOv2 ViT-B/14}}} & \emph{Backbone only} & \num{9} & \num{36.2} & \num{10.49}\tablestd{0.06} & \num{84.3} & \num{100.00}\\
    \cmidrule(l){2-7}
    & {\textcolor{black!55}{\emph{Linear baselines}}} & \\
    & PCA\textsuperscript{\textdagger}~\citep{Pearson:1901:OLA} & \num{5} & \num{32.95}\tablestd{0.60} & \num{14.49}\tablestd{0.33} & \num{81.83}\tablestd{0.04} & \num{84.17}\tablestd{0.00}\\
    & Semi-NMF~\citep{Ding:2010:CAS} & \textbf{\num{1}} & \num{51.18}\tablestd{0.00} & \num{25.67}\tablestd{0.00} & \num{78.28}\tablestd{0.00} & \num{57.17}\tablestd{0.00}\\
    \cmidrule(l){2-7}
    & {\textcolor{black!55}{\emph{Nonlinear baselines}}} & \\
   & Vanilla SAE~\citep{Bricken:2023:TMD} & \num{10} & \num{19.32}\tablestd{0.26} & \num{10.79}\tablestd{1.34} & \num{77.63}\tablestd{0.01} & \num{47.76}\tablestd{0.05}\\
   & TopK-SAE~\citep{Makhzani:2014:KSA} & \num{8} & \num{47.63}\tablestd{0.30} & \num{26.65}\tablestd{0.34} & \num{82.19}\tablestd{0.07} & \num{88.58}\tablestd{0.05}\\
   & JumpReLU~\citep{Rajamanoharan:2024:JAI} & \num{12} & \num{22.59}\tablestd{0.16} & \num{8.13}\tablestd{0.36} & \uline{\num{84.27}}\tablestd{0.02} & \uline{\num{99.97}}\tablestd{0.01}\\
   & BatchTopK-SAE~\citep{Bussmann:2024:BTK} & \num{6} & \num{50.64}\tablestd{0.29} & \num{29.96}\tablestd{0.16} & \num{82.06}\tablestd{0.14} & \num{89.15}\tablestd{0.01}\\
   & Matryoshka SAE~\citep{Bussmann:2025:LML} & \num{7} & \num{52.14}\tablestd{0.13} & \num{30.17}\tablestd{0.20} & \num{82.13}\tablestd{0.04} & \num{88.71}\tablestd{0.02}\\
   & Archetypal SAE~\citep{Fel:2025:ASA} & \num{4} & \num{51.56}\tablestd{0.52} & \uline{\num{33.01}}\tablestd{0.08} & \num{81.26}\tablestd{0.08} & \num{84.75}\tablestd{0.07}\\
   & Matching Pursuit~\citep{Costa:2025:FFH} & \num{3} & \uline{\num{56.92}}\tablestd{0.13} & \num{31.80}\tablestd{0.21} & \num{82.86}\tablestd{0.05} & \num{90.28}\tablestd{0.01}\\
   & Transcoder~\citep{Dunefsky:2024:TFI} & \num{11} & \num{35.79}\tablestd{2.05} & \num{17.16}\tablestd{1.10} & \num{79.70}\tablestd{0.81} & \num{82.17}\tablestd{0.69}\\
    \cmidrule(l){2-7}
    & \tikz[remember picture,baseline] \node[inner sep=0pt,outer sep=0pt,minimum size=0pt] (row2beg) {}; \ours \emph{(ours)} & \uline{\num{2}} & \bfseries \num{64.85}\tablestd{0.52} & \bfseries \num{34.09}\tablestd{0.85} & \bfseries \num{84.29}\tablestd{0.00} & \bfseries \num{100.00}\tablestd{0.00} & {\rlap{\tikz[remember picture,baseline] \node[inner sep=0pt,outer sep=0pt,minimum size=0pt] (row2end) {};}}\\
  \bottomrule
  \end{tabularx}
  \begin{tikzpicture}[remember picture,overlay]
    \begin{scope}[on background layer]
      \fill[black,opacity=0.10]
        ([xshift=0,yshift=2.7ex]row2beg.north west)
        rectangle
        ([xshift=0,yshift=-1.4ex]row2end.south east);
    \end{scope}
  \end{tikzpicture}
\end{table}

\begin{figure}[t]
  \centering
  \hspace{-2em}
  \begin{minipage}[t]{0.5\linewidth}
    \centering
    \scalebox{0.22}{\input{figures/neuron_collage_2split.tex}}%
    \label{fig:appendix:collage_2split}
  \end{minipage}\hspace{-2.5em}
  \begin{minipage}[t]{0.5\linewidth}
    \centering
    \scalebox{0.22}{\input{figures/neuron_collage_4split.tex}}%
    \label{fig:appendix:collage_4split}
  \end{minipage}
  \vspace{-1em}
  \caption{\textbf{Qualitative examples of \ours disentanglements.} Unit $28$ of DINOv2-B/14 (yellow frame, top-$16$) decomposed by \ours into $2$ (left) and unit $286$ into $4$ (right) subunits (pink frames, top-$4$ each). Labels are annotated by hand. Please refer to \cref{sec:appendix:qualitative} for additional examples.} 
  \label{fig:qualitative_examples}
\end{figure}

\paragraph{Experimental setup.} We study the monosemanticity--faithfulness Pareto front of existing methods and \ours, as previewed in \cref{fig:tradeoff}. We use the final pre-activation-function representation of a ViT-B/14-based DINOv2 vision backbone as the starting representation and compare \ours to a broad set of linear, SAE, and transcoder baselines (see \cref{tab:interpretability_vs_faithfulness} for baselines). %
If applicable, we use an expansion factor of $8$, \ie, the size of the disentangled representation relative to the original one. For PCA~\citep{Pearson:1901:OLA} we report the best-performing factor of $0.25$ among $\{0.25, 0.5, 1\}$ (see \cref{tab:pca_table}), as its expansion factor cannot exceed $1$.
For all methods, we use the ImageNet-1k~\citep{Russakovsky:2015:ILS} evaluation split for evaluation as described below and the training split as the probing dataset. To keep the activation cache tractable for training/clustering, we follow~\citet{Pach:2025:SAL} and retain only $2$ randomly sampled tokens per image. We repeat the sampling three times to run each method independently on the resulting probing datasets, and report mean and standard deviation across runs.

We assess monosemanticity from two complementary perspectives to establish a clear neuron-concept correspondence: \emph{(i)} the samples that highly activate a neuron should be semantically coherent, reflecting a consistent concept and \emph{(ii)} given samples of the concept, the neuron should activate highly. %
We measure the first perspective, \ie\ the \emph{activation-to-concept} relationship, using a \emph{Coherence Score} (C-Score) and an \emph{LLM Rank}. Briefly, both metrics take the top-activating samples for a neuron and measure their semantic coherence (similarity) either using the model's internal representation (inspired by \citep{Pach:2025:SAL,Dreyer:2025:MUA}) or an ensemble of three VLLM judges.
For the second perspective, \ie\ the \emph{concept-to-activation} relationship, we use the \emph{simulation metric} (SIM@1) from \citet{Oikarinen:2024:LEF}. Here, each unit is described through a single concept to score how well this explanation predicts the unit's activation across the entire probing set. By considering the full probing set, this metric also captures lower activation ranges, providing a more comprehensive assessment of monosemanticity than metrics restricted to top-activating samples from the first perspective.
To assess faithfulness to the original function, we report the downstream classification \emph{accuracy} when representations are replaced with method reconstructions and the reconstruction \emph{R$^2$} relative to the original representation. Metric details can be found in \cref{sec:appendix:definitions}.

\myparagraph{Quantitative results.} Results are reported in \cref{tab:interpretability_vs_faithfulness}. %
As discussed in \cref{sec:intro} and previewed in \cref{fig:tradeoff}, there is a clear tradeoff between monosemanticity and faithfulness among prior methods. %
In contrast, \ours achieves competitive or even state-of-the-art results in the monosemanticity metrics while retaining perfect faithfulness as reflected by accuracy and R$^2$ scores that match the \emph{backbone only} baseline (up to numeric precision). Remarkably, \ours achieves the best SIM@1 score, despite this metric also considering lower activation regimes. Unlike TopK-based SAEs and Transcoders, \ours does not threshold activations and therefore produces more weakly activating subunits, which we expected to reduce SIM@1. The strong performance suggests that \ours remains interpretable even beyond its highest-activation regime. %
To conclude, \ours consistently retains perfect faithfulness and hence downstream classification accuracy of the opaque baseline, whereas existing methods lose up to \num{7}pp. At the same time, ours provides substantially more monosemantic neurons. As such, \ours provides better monosemanticity without losing predictive performance

To strengthen these claims, we extend our evaluation across more metrics, layers, backbone sizes, model families, expansion factors, modalities (language), and ablations in \cref{sec:appendix:results,sec:appendix:language_models}.

\myparagraph{Qualitative examples.} We qualitatively compare two polysemantic DINOv2-B/14 units with their \ours subunits in \cref{fig:qualitative_examples}. The original units respond to multiple concepts, whereas their subunits capture distinct, monosemantic concepts. Additional examples are provided in \cref{sec:appendix:qualitative}.

\subsection{Practical considerations}\label{sec:experiments:applications}

\paragraph{Computational efficiency.} \Cref{tab:throughput} shows the time and GPU memory used to process image batches with \ours and the baselines. \ours and Transcoder use the $3{,}072$-dim. input to the target layer whereas the SAEs use the $768$-dim. input. Compared to Transcoders, \ours takes $436$ \vsiclr~ ~$705$\,ms/batch and uses $577$ \vsiclr~$954$\,MB during inference. \ours also outperforms recent SAE variants. It uses less memory than Archetypal SAE ($577$ \vsiclr~$1{,}461$\,MB) and runs faster than Matching Pursuit SAE ($436$ \vsiclr~$1{,}083$\,ms/batch). Standard SAEs take $408$--$416$\,ms/batch and use about $558$\,MB.

\paragraph{Stability.} A robust disentanglement should recover similar concepts when applied repeatedly to the same data. For learned feature dictionaries, variation across runs is a known issue~\citep{Paulo:2025:SAT} that inspired stability-focused methods like Archetypal SAE~\citep{Fel:2025:ASA}. We assess stability by keeping the data fixed and comparing three runs of each method. The learned baselines use different initialization seeds; \ours has no random initialization, so its three runs instead test whether its clustering produces consistent subunits. We match concept vectors between each pair of runs and measure their average similarity (\cref{sec:appendix:dict_structure}).
\Cref{tab:stability} shows that \ours achieves 91.66\% similarity. This outperforms Archetypal SAE’s 86.85\%, all other SAEs and the Transcoder. Excluding the deterministic PCA (100\%), \ours recovers the most consistent subunits in this comparison.

\subsection{Model steering}\label{sec:experiments:steering}

\begin{figure*}[t]
  \centering
    \input{figures/steering_attribution_cake2}%
  \vspace{-0.5em}
  \caption{\textbf{Attribution patching for concept removal with \ours.} Given an image and the query ``Is there a \textcolor{myyellow!80}{cake}?'', we use the logit difference of ``Yes'' and ``No'' to identify \ours subunits that support the target concept. We steer the top cake-attributing subunits (\colorcirclecross{myyellow}{mypurple}) before merging their contributions into layer~$L$. The resulting caption removes the target concept while retaining the remaining scene.}
  \label{fig:steering_attribution}
\end{figure*}

An application of disentangled representations is model steering, which changes a model's output by intervening on subunits associated with a target concept. We study whether the subunits identified by \ours enable \emph{targeted} steering, \ie, removing a target concept from an image caption model while preserving the remaining scene description. We evaluate this task with LLaVA-OneVision~\citep{Li:2025:LOE} on the COCO validation set~\citep{Lin:2014:MCC}.  We use attribution patching~\citep{Syed:2023:APO,Kramar:2024:ATP} to identify the subunits linked to the target concept. Specifically, we ask a binary question such as ``Is there a cake in the image?'' and compute the attribution of the \ours subunits for the difference between the ``Yes'' and ``No'' logits. This difference measures the model's preference for the concept being present, quantifying each subunit's contribution. We suppress the highest-attributing subunits by setting their activations to $a_{\mathrm{new}}=-s\,a_{\max}$, where $s$ controls the intervention strength. \Cref{fig:steering_attribution} illustrates this method.  

\begin{wrapfigure}{r}{0.45\columnwidth}
  \centering
  \begin{sansmath}
\begin{tikzpicture}

\scriptsize
\sffamily

\begin{axis}[
    width=1\linewidth,
    height=0.75\linewidth,
    xmin=0.00, xmax=1.05,
    ymin=0.00, ymax=1.05,
    xlabel={\small Target Removal ($\uparrow$)},
    ylabel={\small Non-target retention ($\uparrow$)},
    xtick={0,0.2,0.4,0.6,0.8,1},
    ytick={0,0.2,0.4,0.6,0.8,1},
    label style={font=\scriptsize},
    axis line style={thick},
    enlargelimits=false,
    clip=false,
    axis background/.style={fill=gray!5,rounded corners=1pt},
    grid=both,
    major grid style={gray!30},
    minor grid style={gray!15},
    axis lines=left,
    trim axis left,
    trim axis right,
]

\addplot[myyellow, thick, opacity=0.8, no marks] coordinates {
(0.091032,0.797530) (0.296447,0.592105) (0.450423,0.459508)
    (0.730964,0.213017) (0.971235,0.091399)
};
\addplot[only marks, mark=*, myyellow, opacity=0.8]
coordinates { (0.091032,0.797530) (0.296447,0.592105) (0.450423,0.459508) (0.730964,0.213017) (0.971235,0.091399)};
\node[myyellow, opacity=0.8, anchor=north] at (axis cs:0.450423,0.459508) [yshift=-1pt, xshift=-2pt] {ASAE};

\addplot[myyellow, very thick, opacity=0.8, no marks] coordinates {
     (0.027073,0.925842) (0.180372,0.783918) (0.474788,0.640270)
    (0.839255,0.312351) (0.901861,0.171678)
};
\addplot[only marks, mark=*, myyellow, opacity=0.8]
coordinates {(0.027073,0.925842) (0.180372,0.783918) (0.474788,0.640270) (0.839255,0.312351) (0.901861,0.171678)};
\node[myyellow, opacity=0.8, anchor=south west] at (axis cs:0.839255,0.312351) [yshift=4pt, xshift=-9pt] {BatchTopK};

\addplot[myyellow, very thick, opacity=0.8, no marks] coordinates {
     (0.033503,0.916858) (0.234518,0.725943) (0.476819,0.599365)
    (0.791201,0.344404) (0.957699,0.101926)
};
\addplot[only marks, mark=*, myyellow, opacity=0.8]
coordinates { (0.033503,0.916858) (0.234518,0.725943) (0.476819,0.599365) (0.791201,0.344404) (0.957699,0.101926)};
\node[myyellow, opacity=0.8, anchor=east] at (axis cs:0.791201,0.344404) [yshift=-3pt, xshift=9pt] {MSAE};

\addplot[myyellow, very thick, opacity=0.8, no marks] coordinates {
     (0.021320,0.951149) (0.214213,0.594898) (0.528596,0.604295)
    (0.898816,0.315840) (0.978680,0.092710)
};
\addplot[only marks, mark=*, myyellow, opacity=0.8]
coordinates { (0.021320,0.951149) (0.214213,0.594898) (0.528596,0.604295) (0.898816,0.315840) (0.978680,0.092710)};
\node[myyellow, opacity=0.8, anchor=north east] at (axis cs:0.214213,0.594898) [yshift=2pt] {MpSAE};

\addplot[gray!70!black, thick, densely dashed, opacity=0.8, no marks] coordinates {
     (0.010829,0.971718) (0.324873,0.643255) (0.553976,0.412684)
    (0.747208,0.215064) (0.946531,0.091218)
};
\addplot[only marks, mark=*, mark options={fill=black,draw=black}, black, opacity=0.8]
coordinates { (0.010829,0.971718) (0.324873,0.643255) (0.553976,0.412684) (0.747208,0.215064) (0.946531,0.091218)};
\node[black, opacity=0.8, anchor=north] at (axis cs:0.553976,0.412684) [yshift=-3pt, xshift=3pt] {Original};

\addplot[mypink, very thick, opacity=0.8, no marks] coordinates {
     (0.009137,0.970962) (0.213875,0.808994) (0.541117,0.669933) (0.691371,0.562533)
    (0.954653,0.084009)
};
\addplot[only marks, mark=*, mypink, opacity=0.8]
coordinates { (0.009137,0.970962) (0.213875,0.808994) (0.541117,0.669933) (0.691371,0.562533) (0.954653,0.084009)};
\node[mypink, opacity=0.8, anchor=south] at (axis cs:0.541117,0.669933) [yshift=3pt] {\ours \textit{(ours)}};

\addplot[only marks, mark=square*, mark options={fill=white,draw=black,thick}, black]
coordinates {(0.0000,1.0000)};
\node[black, opacity=0.8, anchor=west] at (axis cs:0.0000,1.0000) [xshift=3pt] {no intervention};

\addplot[only marks, mark=square*, mark options={fill=white,draw=black,thick}, black]
coordinates {(1.0000,0.0000)};
\node[black, opacity=0.8, anchor=east] at (axis cs:1.0000,0.0000) [xshift=-10pt, yshift=4pt] {collapsed model};

\end{axis}

\pgfresetboundingbox
\path ([xshift=-20pt,yshift=-20pt]current axis.south west) rectangle (current axis.north east);

\end{tikzpicture}
\end{sansmath}
  \vspace{-0.5em}  
\caption{\textbf{Target removal \vsiclr non-target retention on COCO.} We vary intervention strength and plot target removal against non-target retention. Points mark the setting that best balances target removal and non-target retention across multiple target-removal ranges per method. %
\ours\~(\colorcircle{mypink}) achieves the best Pareto front relative to the tested baselines. See \cref{sec:appendix:steering} for full results.}
  \label{fig:steering_quantitative_50}
  \vspace{-3.5em}
\end{wrapfigure}

We compare \ours with interventions on the original representation and the four best-performing SAE baselines from \cref{tab:interpretability_vs_faithfulness}. We measure \emph{target removal}, whether the edited caption omits the target concept, and \emph{non-target retention}, whether the other concepts in the original caption are preserved. \Cref{fig:steering_quantitative_50} shows the trade-off. At multiple target-removal rates, \ours-based steering preserves more non-target concepts than steering on the original neurons and the SAE baselines. This indicates that its subunits enable more targeted interventions, removing the target concept while causing fewer changes to the remaining caption. Full details and results are provided in \cref{sec:appendix:steering_definition,sec:appendix:steering}.

\section{Discussion}\label{sec:discussion}
\paragraph{Limitations.} In this work, we focus on linear layers and $1 \times 1$ convolutions, which cover a wide range of relevant layers; extending \ours to attention heads and other architectural components remains future work. Since the disentanglement is run independently per neuron, subunits across different neurons may encode the same concept, creating redundancy at the layer level. This redundancy could be reduced through correlation-based pruning if required by the application, which however would sacrifice faithfulness. Relatedly, unlike SAEs, \ours doesn't produce sparse or non-negative codes. Instead, its lossless linear disentanglement yields dense, partially negative activations, which may limit some aspects of interpretability. Nevertheless, our strong steering results show that concept-specific subunits can still be effectively identified and manipulated despite these properties. Moreover, monosemanticity is inherently hard to measure; like prior work, we rely on complementary proxy metrics that may not fully capture whether a unit represents a single human-interpretable concept~\citep{Dreyer:2025:MUA, Hesse:2025:DPC, Pach:2025:SAL}. %

\paragraph{Conclusion.} We introduce \ours for disentangling polysemantic units into more monosemantic subunits. \ours uses concept-specific structure in upstream circuits and restructures weights accordingly, ensuring that each subunit primarily connects to units in the preceding layer that respond to the corresponding concept. The disentanglement is constructed such that the sum of all subunits faithfully reconstructs the original neuron, enabling application without any loss in downstream predictive performance.
Through a comprehensive empirical evaluation, we show that \ours achieves near-perfect faithfulness (up to numerical precision) while often outperforming existing methods in terms of monosemanticity. We further demonstrate improved stability compared to SAEs and transcoders, and provide quantitative evidence that \ours can be used for steering.
Overall, we consider \ours a promising addition to the interpretability toolkit: to the best of our knowledge, it is the first explicit disentanglement method to achieve state-of-the-art monosemanticity at scale without sacrificing any predictive performance, making it a strong foundation for future research.

\subsection*{AI use statement}
We used generative AI tools to assist with drafting and editing manuscript text and with writing and editing research code. We did not use them to develop the research idea, propose hypotheses, design the methodology or experiments, or analyze or interpret the results. The authors reviewed AI-assisted work and take responsibility for the final manuscript, code, and reported results.

\subsection*{Ethics statement}

Interpretability research of the kind presented here contributes to safer and more transparent AI: more monosemantic representations make it easier to audit models for spurious correlations, detect failure modes, and satisfy emerging regulatory requirements around model explainability. As with any interpretability method, however, a more precise understanding of feature structure could in principle assist adversarial attack design, and the concept-steering capability we demonstrate could be applied to manipulate model outputs in unintended ways. These risks are not unique to \ours and are shared by mechanistic interpretability research broadly; they do not require mitigation beyond standard responsible-use practices for ML research artifacts.

\subsection*{Reproducibility statement}

We describe \ours and the experimental comparisons in the main text. The appendix provides the metric definitions and evaluation procedures, including the judge prompts and steering protocol, as well as baseline training settings, datasets, model configurations, hardware, and software details. We report the data and initialization seeds used for repeated experiments so that the comparisons can be reproduced.

\section*{Acknowledgments}\label{sec:Acknowledgments}

We acknowledge support from the Max Planck Computing and Data Facility. 
SSM has been funded by the Deutsche Forschungsgemeinschaft (DFG, German Research Foundation) --~project No.~529680848. Additionally, SSM has received funding from the DFG under Germany's Excellence Strategy (EXC-3057/1 ``Reasonable Artificial Intelligence'', Project No.~533677015).

\bibliography{iclr2027_conference}
\bibliographystyle{iclr2027_conference}

\appendix

\newpage

{\huge \textbf{Appendix}}
{
  \startcontents[appendix]
  \printcontents[appendix]{}{1}{}
}
\newpage

\section{Metric definitions}\label{sec:appendix:definitions}

This section gives the detailed definition of every metric used in \cref{sec:experiments,sec:appendix:results}. %

\subsection{LLM Rank}\label{sec:appendix:coherence}
The LLM Rank metric reported in \cref{tab:interpretability_vs_faithfulness}, \cref{tab:interpretability_vs_faithfulness2} and \cref{tab:generalization} ranks methods by how often three vision--language model (VLM) judges, used as proxies for a human evaluator~\citep{Zheng:2023:JLA}, prefer one method's disentangled units over another's in a pairwise coherence test. Each comparison shows two grids of top-activating images side by side and asks which is more coherent; methods are ranked by their consensus win rate across all such pairs.

\paragraph{Grid construction.} For each unit, we build a $3{\times}3$ grid of its top-$9$ most-activating images from the ImageNet-1k evaluation split, at $336{\times}336$ pixels per cell. For backbones that produce token sequences, each image's activation score is the maximum unit activation across all tokens (matching the C-Score convention; \cf~\cref{sec:appendix:msscore}). \Cref{fig:appendix:example_grid} shows one such grid.

\paragraph{Judges.} The three judges are Qwen3-VL-30B-A3B-Instruct~\citep{Qwen:2025:QVL} (Apache-2.0), Gemma-3-27B~\citep{Gemma:2025:GTR} (Gemma terms of use), and InternVL3-14B~\citep{Zhu:2025:IEA} (MIT). Each is served through \texttt{vLLM} with \texttt{tensor\_parallel\_size}~$=$~$1$, \texttt{max\_model\_len}~$=$~\num{32768}, \texttt{limit\_mm\_per\_prompt}~$=$~\texttt{\{image: 2\}}, and a generation batch size of $64$. Decoding is greedy: \texttt{temperature}~$=$~$0$, \texttt{max\_tokens}~$=$~$256$, with a fixed seed of $42$ passed to vLLM's \texttt{SamplingParams}.

\paragraph{Pairwise presentation.} For each seeded token dataset, we sample $50$ units from each method. For every pair of methods, we present the sampled units as $50$ side-by-side comparisons. The paired units are not assumed to represent the same concept. We show each comparison again with the positions swapped to reduce position bias. With $12$ methods, this yields $19{,}800$ presentations per VLM judge; the six-method generalization comparison yields $4{,}500$ per setting.

\paragraph{Aggregation and confidence intervals.} Each pair of grids has a \emph{consensus winner}: the option chosen by the majority of the three judges. Each method has a \emph{consensus win rate}: the fraction of pairs it wins, pooled across the three seeded token datasets. Methods are ranked by this rate, and we attach $95\%$ confidence intervals by bootstrapping the pairwise comparisons $1{,}000$ times. The ordinal LLM Rank in \cref{tab:interpretability_vs_faithfulness} is derived from these rates; the rates themselves and their intervals are listed in \cref{tab:appendix:agreement}.

\paragraph{Prompt.} All three VLM judges receive the identical model-agnostic prompt:
\begin{quote}\small\ttfamily
Each grid shows the 9 images that most strongly activate a neural network unit. \\[0.4em]
A COHERENT unit captures a single, nameable concept: most images share a common object category, object part, color, texture, material, shape, or other visual or semantic attribute. \\[0.4em]
An INCOHERENT unit shows images with no discernible shared property. \\[0.4em]
Grid A = first image. Grid B = second image. \\[0.4em]
Describe what concept each grid captures (or why it seems incoherent), then state which is more coherent. \\[0.4em]
Respond as JSON only: \\
\{"reasoning\_a": "...", "reasoning\_b": "...", "answer": "A" or "B"\}
\end{quote}

\begin{figure}[t]
  \centering
  \resizebox{\linewidth}{!}{\input{figures/example_vlm_grid}}
  \caption{\textbf{Example pairwise grid shown to a VLM judge.} Each panel is a $3{\times}3$ collage of the top-$9$ most-activating images for one disentangled unit; two such panels from competing methods (left: \ours, right: Archetypal SAE) are placed side by side and the three judges select the more coherent grid. Beneath the grids each judge's per-grid reasoning and the resulting consensus winner is shown. The example is from DINOv2-S/14.}
  \label{fig:appendix:example_grid}
\end{figure}

\begin{table}[t]
  \caption{
    \textbf{Per-method consensus win rate.} Fraction of pairwise comparisons
    in which a method was declared the winner under the majority-of-three VLM
    consensus (ties dropped), pooled across the three seeded token datasets.
    These are the continuous values backing the ordinal LLM Rank in
    \cref{tab:interpretability_vs_faithfulness, tab:interpretability_vs_faithfulness2}. Brackets give the $95\%$
    bootstrap confidence interval from $1{,}000$ resamples of unit-pair
    comparisons. Same backbones, method order, and shading as
    \cref{tab:interpretability_vs_faithfulness} and \cref{tab:interpretability_vs_faithfulness2}.
  }
  \label{tab:appendix:agreement}
  \centering
  \setlength{\tabcolsep}{5.0pt}
  \scriptsize
  \begin{tabularx}{\linewidth}{@{}rXS[table-format=2.2]c@{}l@{}}
    \toprule
    & & {\bfseries Consensus win rate (\%)} & {\bfseries 95\% CI} \\
    \midrule
    \parbox[t]{2mm}{\multirow{16}{*}{\rotatebox[origin=c]{90}{ImageNet-21k ViT-B/16}}}
    & \textit{Backbone only} & \num{39.82} & {[38.1,\,41.4]} \\
    \cmidrule(l){2-4}
    & {\textcolor{black!55}{\textit{Linear baselines}}} & \\
    & PCA              & \num{42.27} & {[40.6,\,44.4]} \\
    & Semi-NMF         & \num{46.55} & {[44.9,\,48.1]} \\ 
    \cmidrule(l){2-4}
    & {\textcolor{black!55}{\textit{Nonlinear baselines}}} & \\
    & Vanilla SAE      & \num{49.18} & {[47.4,\,50.8]} \\
    & TopK-SAE         & \num{55.42} & {[53.7,\,57.0]} \\
    & JumpReLU         & \num{21.52} & {[20.2,\,22.9]} \\ 
    & BatchTopK-SAE    & \num{52.79} & {[51.3,\,54.6]} \\ 
    & Matryoshka SAE   & \num{53.42} & {[51.6,\,55.2]} \\ 
    & Archetypal SAE   & \uline{\num{62.79}} & {[61.1,\,64.4]} \\ 
    & Matching Pursuit & \num{58.64} & {[56.9,\,60.3]} \\ 
    & Transcoder       & \num{38.48} & {[36.8,\,40.1]} \\
    \cmidrule(l){2-4}
    & \tikz[remember picture,baseline] \node[inner sep=0pt,outer sep=0pt,minimum size=0pt] (agrowAbeg) {};\ours \textit{(ours)} & \bfseries \num{79.12} & {[77.8,\,80.4]} & {\rlap{\tikz[remember picture,baseline] \node[inner sep=0pt,outer sep=0pt,minimum size=0pt] (agrowAend) {};}}\\
    \midrule
    \parbox[t]{2mm}{\multirow{15}{*}{\rotatebox[origin=c]{90}{DINOv2 ViT-B/14}}}
    & \emph{Backbone only} & \num{42.79} & {[41.0,\,44.45]} \\
    \cmidrule(l){2-4}
    & {\textcolor{black!55}{\textit{Linear baselines}}} & \\
    & PCA              & \num{53.91} & {[52.1,\,55.6]} \\
    & Semi-NMF         & \bfseries{\num{78.12}} & {[76.7,\,79.5]} \\
    \cmidrule(l){2-4}
    & {\textcolor{black!55}{\textit{Nonlinear baselines}}} & \\
    & Vanilla SAE      & \num{35.27} & {[33.6,\,36.9]} \\
    & TopK-SAE         & \num{47.39} & {[45.7,\,49.1]} \\
    & JumpReLU         & \num{27.27} & {[25.8,\,28.8]} \\
    & BatchTopK-SAE    & \num{52.70} & {[51.0,\,54.4]} \\
    & Matryoshka SAE   & \num{49.18} & {[47.5,\,50.8]} \\
    & Archetypal SAE   & \num{55.82} & {[54.1,\,57.4]} \\
    & Matching Pursuit & \num{57.00} & {[55.3,\,58.8]} \\
    & Transcoder       & \num{33.67} & {[32.0,\,35.3]} \\
    \cmidrule(l){2-4}
    & \tikz[remember picture,baseline] \node[inner sep=0pt,outer sep=0pt,minimum size=0pt] (agrowBbeg) {};\ours \textit{(ours)} &  \uline{\num{66.88}} & {[65.3,\,68.5]} & {\rlap{\tikz[remember picture,baseline] \node[inner sep=0pt,outer sep=0pt,minimum size=0pt] (agrowBend) {};}}\\
  \bottomrule
  \end{tabularx}
  \begin{tikzpicture}[remember picture,overlay]
    \begin{scope}[on background layer]
      \fill[black,opacity=0.10]
        ([xshift=0,yshift=2.7ex]agrowAbeg.north west)
        rectangle ([xshift=0,yshift=-1.4ex]agrowAend.south east);
    \end{scope}
    \begin{scope}[on background layer]
      \fill[black,opacity=0.10]
        ([xshift=0,yshift=2.7ex]agrowBbeg.north west)
        rectangle ([xshift=0,yshift=-1.4ex]agrowBend.south east);
    \end{scope}
  \end{tikzpicture}
\end{table}

\subsection{Coherence score}\label{sec:appendix:msscore}
To complement the LLM-based coherence metric, we also measure whether the images that most strongly activate a unit have similar representations. Like the monosemanticity score of~\citet{Pach:2025:SAL}, C-Score compares images according to their representation similarity and gives more weight to images that activate the unit strongly. Our calculation differs in two ways: we evaluate only the most-activating images, and we measure their similarity in the backbone's own representation space before disentanglement, rather than with a separate image encoder.

\paragraph{Top-$K$ selection.} For each unit, we select its $100$ highest-activating images from the ImageNet-1k evaluation split, using the unit's maximum activation across image positions. Evaluating a fixed set of $100$ highly activating examples follows~\citet{Dreyer:2024:PTP,OMahony:2023:DNR}; unlike the score of~\citet{Pach:2025:SAL} over a broader image set, our C-Score focuses on the examples that most strongly activate each unit.

\paragraph{Per-image embedding.} For each selected image, we localize the spatial position (or token) at which the unit attains its maximum activation, and extract the backbone's representation at that position. We use the \emph{original (pre-disentanglement) backbone representation} -- the same activation space in which the units live -- rather than a fixed external vision model. This ensures the metric only measures separability of concepts already encoded in the model, aligning the metric with the capabilities of any disentanglement method that operates on this backbone. Each embedding is $\ell_2$-normalized before scoring.

\paragraph{Weight construction.} Each image $i$ is weighted by the unit's activation at its chosen position, min--max normalized to $[0, 1]$ using the activation range observed across the entire evaluation split. Higher-activating images contribute more.

\paragraph{Score.} Let $\{(e_i, w_i)\}_{i=1}^{K}$ be the $\ell_2$-normalized embeddings and weights for the top-$K$ images. The C-Score is the weighted mean off-diagonal pairwise cosine similarity:
\begin{equation}
    \mathrm{C}(u) \;=\; \frac{\bigl\lVert \sum_i w_i e_i \bigr\rVert^2 \;-\; \sum_i w_i^2}{\bigl(\sum_i w_i\bigr)^2 \;-\; \sum_i w_i^2}\,.
\end{equation}
The score lies in $[0, 1]$: $1$ corresponds to all top-$K$ embeddings being identical (perfectly monosemantic), $0$ to them being orthogonal in expectation. We report the metric as a percentage. Units that never activate on the evaluation split are excluded from the per-method aggregation.

\subsection{Simulation metric (SIM@1)}\label{sec:appendix:sim}
To test whether a concept explains a unit's behavior, we ask whether the concept can predict that unit's activation on images not used to choose the explanation. Following~\citet{Oikarinen:2024:LEF}, we select one concept per unit using training and validation images. The ``@1'' in SIM@1 denotes this one-concept explanation. We use SigLIP2-Base (\texttt{google/siglip2-base-patch16-224}) to estimate concept presence while selecting the explanation. On held-out images, a separate model, SigLIP2-SO400M (\texttt{google/siglip2-so400m-patch14-384}), estimates how strongly that concept is present. We use these estimates and a fitted weight to predict the unit's activation, then measure the correlation between predicted and actual activations across the held-out images. We average this correlation across units and report it as a percentage. Higher scores mean that a single concept better predicts when and how strongly a unit responds---not only for its most activating images, but across the test images. This complements the top-activating-image metrics used for the activation-to-concept perspective in \cref{sec:experiments:interp_faith}.

\subsection{Reconstruction $R^2$}\label{sec:appendix:reconstruction_r2}
We compare each reconstructed activation with the activation produced by the original layer. Let $a_i$ be an original activation value, $\hat a_i$ its reconstruction, and $\bar a$ the mean of the original activation values. We compute $R^2=1-\frac{\sum_i(a_i-\hat a_i)^2}{\sum_i(a_i-\bar a)^2}$ over all evaluated images, tokens, and channels, and report $100R^2$ in the table. An exact reconstruction scores $100\%$; always predicting the mean scores $0\%$; a worse reconstruction can score below $0\%$.

\subsection{Quantitative steering benchmark}\label{sec:appendix:steering_definition}

We report target removal and non-target retention. For target removal, we ask Gemma-3-12B-IT~\citep{Gemma:2025:GTR} whether the edited caption mentions the target concept or a close synonym. Target removal is the fraction of captions for which the answer is ``No.'' We evaluate only image--concept pairs for which the judge confirms that the original caption mentions the target concept.

Non-target retention is the fraction of other COCO concepts named in the original caption that remain in the edited caption. We use whole-word matching for this metric. We also report removal--retention utility, \(\mathcal{U}=100r_{\mathrm{rem}}r_{\mathrm{ret}}\), where $r_{rem}$ is target removal rate and $r_{ret}$ is non-target retention rate.

The tested intervention settings and the procedure used to select points for the main-paper scatter plot are described alongside the full results in \cref{sec:appendix:steering}.

\Cref{tab:steering_prompts} lists the prompts used for attribution and target removal.

\begin{table}[t]
  \caption{\textbf{Prompts for the quantitative steering benchmark.}}
  \label{tab:steering_prompts}
  \centering
  \scriptsize
  \begin{tabularx}{\linewidth}{@{}lX@{}}
    \toprule
    \textbf{Purpose} & \textbf{Prompt} \\
    \midrule
    Caption generation &
    ``Describe this image in detail, mentioning every distinct object you can see.'' \\
    Attribution VQA &
    ``Is there a \texttt{\{concept\}} in the image? Answer Yes or No.'' \\
    Target judge &
    ``Does this caption mention \texttt{\{concept\}} (or a close synonym)? Caption: \texttt{\{caption\}}. Answer YES or NO.'' \\
    Coherence judge &
    ``Is this caption grammatically correct and coherent? Caption: \texttt{\{caption\}}. Answer YES or NO.'' \\
    Scene judge &
    ``Caption A: \texttt{\{caption A\}}. Caption B: \texttt{\{caption B\}}. Do these describe the same overall scene? Answer YES or NO.'' \\
    \bottomrule
  \end{tabularx}
\end{table}

\subsection{Stability}\label{sec:appendix:dict_structure}
This subsection details the cross-init stability protocol used in \cref{sec:experiments:applications} and reported in \cref{tab:stability}, following~\citet{Fel:2025:ASA}.

\paragraph{Setup.} We evaluate stability on DINOv2-B/14 using the configurations from the main comparison: expansion factor $8$ for the learned baselines and $0.25$ for PCA. All three runs of a method use the same token dataset, selected with data seed $42$. The learned baselines are fitted with initialization seeds $\{42, 7, 1024\}$ while keeping their data, hyperparameters, and training schedule fixed. We fit PCA three times on the same data; because it is deterministic, the resulting dictionaries should agree. \ours has no random initialization, so we likewise repeat its construction three times on identical inputs to measure any variation in clustering. We exclude dead concept vectors before comparing dictionaries.

\paragraph{Measuring stability.} We measure whether separate runs recover the same concept directions. A dictionary is a collection of vectors, but their order has no meaning: the first vector in one run need not represent the same concept as the first vector in another. Following~\citet{Fel:2025:ASA}, we therefore find the one-to-one pairing that makes the vectors from two runs as similar as possible, then average the cosine similarity of the paired vectors. For PCA, Semi-NMF, the SAEs, and the Transcoder, vectors can be paired across the whole dictionary. See~\citet{Fel:2025:ASA} for further details of this matching metric.

\ours's subunits belong to particular original neurons, so we pair subunits only when they come from the same neuron. The number of subunits for a neuron can differ between runs. When this happens, the implementation takes the first set of vectors up to the size of the smaller group, finds their best pairing, and lowers that neuron's similarity score to account for subunits without a counterpart. For example, if two runs produce five and six subunits and their paired vectors score $0.90$, the additional penalty of one-sixth lowers the score to approximately $0.73$. We average the scores across neurons and then across the three pairs of runs. The \ours value in \cref{tab:stability} includes this penalty; higher values indicate greater stability.

\subsection{Computational efficiency}\label{sec:appendix:throughput}

\paragraph{Protocol.} We measure how long each method takes to process a batch of $256$ images on one H100 GPU, and how much GPU memory it uses while doing so. Each method and seed runs in a fresh process. After $200$ practice batches that prepare the model, we measure $500$ further batches and report their average time. For memory, we report the largest amount occupied by PyTorch tensors during those $500$ measured batches. This includes the model and inputs already in memory, but excludes temporary memory used earlier while loading and preparing the model. Thus, the memory column describes repeated inference, not the maximum memory needed to start the program.

All methods process the images through the backbone model. PCA and the SAEs take the activation at the target layer, reconstruct it, and pass that reconstruction through the remainder of the backbone. The Transcoder replaces the target layer's final projection, so the original projection is not run as well. \ours likewise replaces that projection. It produces sub-unit activations and sums those belonging to each original neuron, restoring the layer's original output size before the rest of the backbone runs.

For methods that support it, we prepare the computation for repeated execution using \texttt{torch.compile} with CUDA graphs. This preparation happens before the measured batches, so the table reports the cost of repeatedly running the prepared model. Semi-NMF requires an additional calculation for every new image. It repeatedly searches for the values that best reconstruct that image's activation. Our Semi-NMF row does not time this optimization step. It measures only the final reconstruction step using placeholder values of the correct size. Its time and memory therefore understate the cost of using Semi-NMF on new images and should be interpreted cautiously.

\section{Results}\label{sec:appendix:results}

\subsection{More polysemanticity metrics}\label{sec:appendix:polysem}

To complement C-Score and SIM@1, we adapt the Clarity and Polysemanticity measures of \citet{Dreyer:2025:MUA}. For each unit, we collect its 30 most activating examples and represent them using CLIP ViT-B/16~\citep{Radford:2021:LTV} CLS embeddings. Clarity is the mean pairwise cosine similarity between these embeddings. A high Clarity score indicates that the examples share a common semantic theme. For Polysemanticity, we partition the same embeddings into two $k$-means clusters and compute one minus the similarity between their centroids. A high Polysemanticity score therefore indicates that a unit responds to two semantically distinct groups, whereas a low score indicates more coherent behavior. As shown in \cref{tab:add_mono_exp}, \ours achieves the highest Clarity and the second-lowest Polysemanticity score.

\begin{table}[t]
  \caption{
    \textbf{Additional monosemanticity metrics on DINOv2 ViT-B/14.} Following~\citet{Dreyer:2025:MUA}, Clarity measures the similarity among each unit's $30$ most-activating examples (higher is better); Polysemanticity measures the separation between two groups of those examples (lower is better). Both use CLIP ViT-B/16 image representations. Scores are percentages, reported as mean $\pm$ standard deviation across three seeded token datasets. \ours's row is shaded; the best and second-best values in each column are bold and underlined.
  }
  \label{tab:add_mono_exp}
  \centering
  \setlength{\tabcolsep}{5.0pt}
  \scriptsize
  \begin{tabularx}{\linewidth}{@{}rXS[table-format=1(1.2)]S[table-format=2.2(1.4)]S[table-format=1.2(1.4)]@{}l@{}}
    \toprule
    & & \multicolumn{2}{c}{\bfseries Monosemanticity} \\
    \cmidrule(lr){3-3}\cmidrule(l){4-5}
    {\bfseries Model} & {\bfseries Method}
      & {\bfseries Clarity\,$(\uparrow)$}
      & {\bfseries Polysemanticity\,$(\downarrow)$} \\
    \midrule
    \parbox[t]{2mm}{\multirow{15}{*}{\rotatebox[origin=c]{90}{DINOv2 ViT-B/14}}}
    & \emph{Backbone only} & \num{59.91} & \num{23.45}\\
    \cmidrule(l){2-5}
    & {\textcolor{black!55}{\textit{Linear baselines}}} & \\
    & PCA & \num{62.84}\tablestd{0.11} & \num{23.45}\tablestd{0.65}\\
    & Semi-NMF & \num{55.90}\tablestd{0.02} & \num{21.69}\tablestd{0.05}\\
    \cmidrule(l){2-5}
    & {\textcolor{black!55}{\textit{Nonlinear baselines}}} & \\
    & Vanilla SAE & \num{54.80}\tablestd{1.20} & \num{22.88}\tablestd{0.37}\\
    & TopK-SAE & \num{61.42}\tablestd{0.19} & \num{16.97}\tablestd{0.13}\\
    & JumpReLU  & \num{53.46}\tablestd{0.40} & \num{19.95}\tablestd{0.32}\\
    & BatchTopK-SAE & \num{62.89}\tablestd{0.10} & \num{16.53}\tablestd{0.06}\\
    & Matryoshka SAE  & \num{63.53}\tablestd{0.10} & \num{15.24}\tablestd{0.04}\\
    & Archetypal SAE  & \num{64.82}\tablestd{0.11} & \num{15.52}\tablestd{0.17}\\
    & Matching Pursuit  & \uline{\num{65.53}}\tablestd{0.13} & \textbf{\num{15.04}}\tablestd{0.04}\\
    & Transcoder  & \num{57.73}\tablestd{0.37} & \num{20.62}\tablestd{0.62}\\    \cmidrule(l){2-5}
    & \tikz[remember picture,baseline] \node[inner sep=0pt,outer sep=0pt,minimum size=0pt] (stabrowbeg) {};\ours \textit{(ours)} & \textbf{\num{67.98}}\tablestd{0.43} & \uline{\num{15.07}}\tablestd{0.49} & {\rlap{\tikz[remember picture,baseline] \node[inner sep=0pt,outer sep=0pt,minimum size=0pt] (stabrowend) {};}}\\
    \bottomrule
  \end{tabularx}
  \begin{tikzpicture}[remember picture,overlay]
    \begin{scope}[on background layer]
      \fill[black,opacity=0.10]
        ([xshift=0,yshift=2.7ex]stabrowbeg.north west)
        rectangle
        ([xshift=0,yshift=-1.4ex]stabrowend.south east);
    \end{scope}
  \end{tikzpicture}
\end{table}

\subsection{Stability}\label{sec:appendix:stability_results}

\Cref{tab:stability} reports the per-method cross-init stability on DINOv2-B/14. The analysis is in \cref{sec:experiments:applications}; the measurement protocol is in \cref{sec:appendix:dict_structure}.

\begin{table}[t]
  \caption{
    \textbf{Stability.} Cross-init stability for all methods on DINOv2-B/14 at the same
    configuration as \cref{sec:experiments:interp_faith}. Protocol details are in \cref{sec:appendix:dict_structure}. Highest similarity in \textbf{bold}, second-highest \underline{underlined}.
  }
  \label{tab:stability}
  \centering
  \setlength{\tabcolsep}{5.0pt}
  \scriptsize
  \begin{tabularx}{\linewidth}{@{}rXS[table-format=1(1.2)]@{}l@{}}
    \toprule
    & & \multicolumn{1}{c}{\bfseries Stability}\\
    \cmidrule(lr){3-3}
    {\bfseries Model} & {\bfseries Method}
      & {\bfseries Cross-Init Sim.\,$(\uparrow)$}\\
    \midrule
    \parbox[t]{2mm}{\multirow{14}{*}{\rotatebox[origin=c]{90}{DINOv2 ViT-B/14}}}
    & {\textcolor{black!55}{\textit{Linear baselines}}} & \\
    & PCA              & \bfseries \num{100} \\
    & Semi-NMF         & \num{77.31}\tablestd{0.1} \\
    \cmidrule(l){2-3}
    & {\textcolor{black!55}{\textit{Nonlinear baselines}}} & \\
    & Vanilla SAE      & \num{13.32}\tablestd{0.01} \\
    & TopK-SAE         & \num{81.24}\tablestd{0.44} \\
    & JumpReLU         & \num{13.38}\tablestd{0.01} \\
    & BatchTopK-SAE    & \num{58.93}\tablestd{0.29} \\
    & Matryoshka SAE   & \num{50.77}\tablestd{0.12} \\
    & Archetypal SAE   & \uline{\num{86.85}}\tablestd{0.58}  \\
    & Matching Pursuit & \num{78.08}\tablestd{0.81} \\
    & Transcoder       & \num{65.07}\tablestd{0.88}  \\
    \cmidrule(l){2-3}
    & \tikz[remember picture,baseline] \node[inner sep=0pt,outer sep=0pt,minimum size=0pt] (stabrowbeg) {};\ours \textit{(ours)} & \uline{\num{91.66}}\tablestd{3.85} & {\rlap{\tikz[remember picture,baseline] \node[inner sep=0pt,outer sep=0pt,minimum size=0pt] (stabrowend) {};}}\\
    \bottomrule
  \end{tabularx}
  \begin{tikzpicture}[remember picture,overlay]
    \begin{scope}[on background layer]
      \fill[black,opacity=0.10]
        ([xshift=0,yshift=2.7ex]stabrowbeg.north west)
        rectangle
        ([xshift=0,yshift=-1.4ex]stabrowend.south east);
    \end{scope}
  \end{tikzpicture}
\end{table}

\subsection{Computational efficiency}\label{sec:appendix:throughput_results}

\Cref{tab:throughput} reports per-method throughput and peak GPU memory on DINOv2-B/14 layer~$12$ at batch size $256$ on a single H100, with the original backbone as reference. The analysis is in \cref{sec:experiments:applications}; the measurement protocol is in \cref{sec:appendix:throughput}.

\begin{table}[t]
    \caption{
      \textbf{Computational Efficiency.}
      We measure time per batch (in millisecond per batch) and peak GPU memory during measured inference for methods on DINOv2-B/14 layer~11, batch size $256$, single H100. Input dim is the size of the activation each method receives at the target layer. Means $\pm$ std.\ over three seeds; the original backbone is a reference. Semi-NMF measures reconstruction from placeholder values only, without the search required for new images, and is not directly comparable to the other methods. Full protocol: \cref{sec:appendix:throughput}.
    }
  \label{tab:throughput}
  \centering
  \setlength{\tabcolsep}{5.0pt}
  \scriptsize
  \begin{tabularx}{\linewidth}{@{}rXS[table-format=4]S[table-format=4.2(1.2)]S[table-format=5.2(4.2)]@{}l@{}}
    \toprule
    & & & \multicolumn{2}{c}{\bfseries Comp. Efficiency} \\
    \cmidrule(l){4-5}
    {\bfseries Model} & {\bfseries Method}
      & {\bfseries Input dim}
      & {\bfseries ms/batch $(\downarrow)$}
      & {\bfseries Peak mem\,MB $(\downarrow)$} \\
    \midrule
    \parbox[t]{2mm}{\multirow{15}{*}{\rotatebox[origin=c]{90}{DINOv2 ViT-B/14}}}
    & \textit{Backbone only} & N/A & \num{374.06}\tablestd{0.72} & \num{501.80}\tablestd{0.00} \\
    \cmidrule(l){2-5}
    & {\textcolor{black!55}{\textit{Linear baselines}}} & \\
    & PCA (EF $1$)     & \num{768}  & \num{377.48}\tablestd{0.21} & \num{506.52}\tablestd{0.00} \\
    & Semi-NMF (EF $8$, decode only) & \num{768} & \num{387.44}\tablestd{0.45} & \num{539.55}\tablestd{0.00} \\
    \cmidrule(l){2-5}
    & {\textcolor{black!55}{\textit{Nonlinear baselines}}} & \\
    & Vanilla SAE      & \num{768}  & \num{407.60}\tablestd{0.76} & \num{558.45}\tablestd{0.00} \\
    & TopK-SAE         & \num{768}  & \num{415.72}\tablestd{0.31} & \num{558.45}\tablestd{0.00} \\
    & JumpReLU         & \num{768}  & \num{409.37}\tablestd{0.48} & \num{558.48}\tablestd{0.00} \\
    & BatchTopK-SAE    & \num{768}  & \num{409.13}\tablestd{0.23} & \num{558.45}\tablestd{0.00} \\
    & Matryoshka SAE   & \num{768}  & \num{409.42}\tablestd{0.07} & \num{558.45}\tablestd{0.00} \\
    & Archetypal SAE   & \num{768}  & \num{418.22}\tablestd{0.26} & \num{1461.28}\tablestd{0.00} \\
    & Matching Pursuit & \num{768}  & \num{1083.01}\tablestd{1.24} & \num{539.55}\tablestd{0.00} \\
    & Transcoder       & \num{3072} & \num{705.42}\tablestd{1.16} & \num{953.84}\tablestd{0.00} \\
    \cmidrule(l){2-5}
    & \tikz[remember picture,baseline] \node[inner sep=0pt,outer sep=0pt,minimum size=0pt] (tprowBbeg) {};\ours \textit{(ours)} & \num{3072} & \num{436.44}\tablestd{3.62} & \num{577.49}\tablestd{4.33} & {\rlap{\tikz[remember picture,baseline] \node[inner sep=0pt,outer sep=0pt,minimum size=0pt] (tprowBend) {};}}\\
    \bottomrule
  \end{tabularx}
  \begin{tikzpicture}[remember picture,overlay]
    \begin{scope}[on background layer]
      \fill[black,opacity=0.10]
        ([xshift=0,yshift=2.7ex]tprowBbeg.north west)
        rectangle
        ([xshift=0,yshift=-1.4ex]tprowBend.south east);
    \end{scope}
  \end{tikzpicture}
\end{table}

\subsection{Quantitative steering benchmark}\label{sec:appendix:steering}

We evaluate target-concept suppression in LLaVA-OneVision~\citep{Li:2025:LOE} on COCO validation images~\citep{Lin:2014:MCC}. We form an image--concept pair when a COCO category is present in an image and mentioned in its original caption. We sample $80$ pairs per retained category, following \citet{Wittenmayer:2026:CLA}. This gives $2{,}960$ pairs across $37$ categories. A sanity check excludes five pairs for which the judge does not identify the target concept in the original caption, since a concept cannot be removed if it is not present initially.

For each image--concept pair, we use attribution patching~\citep{Syed:2023:APO,Kramar:2024:ATP} to rank features by their gradient$\times$activation attribution for the difference between the ``Yes'' and ``No'' logits. We select the smallest set of positively attributing features that accounts for a fraction $m$ of the total positive attribution. We evaluate $m\in\{20\%,40\%,60\%,80\%,100\%\}$.
For steering the selected neurons, we set their activations to\[a_{\mathrm{new}}= -s\,a_{\max}, \] where $s$ is the steering factor and $a_{\max}$ is the feature's maximum activation on the COCO validation set. We evaluate $s\in\{0,\pm0.5,\pm1,\pm2.5,\pm5\}$. For the no-intervention condition, we use the reconstructed activation, $a_{\mathrm{new}}=a_{\mathrm{recon}}$. 

\Cref{tab:steering_utility_grid,tab:steering_non_target_retention_grid,tab:steering_judge_removal_grid} report removal--retention utility, non-target retention, and target removal, respectively. Setting $s=0$ with \ours achieves the highest removal--retention utility: \num{36.25} at $m=80\%$ and \num{38.89} at $m=100\%$.

\paragraph{Scatter-plot construction.} For \cref{fig:steering_quantitative_50}, we divide each method’s evaluated settings into target-removal bins of width \(0.33\) and show the setting with the highest removal–retention utility in each bin, where \(\mathcal{U}=100r_{\mathrm{rem}}r_{\mathrm{ret}}\). Here, \(r_{\mathrm{rem}}\) is the fraction of target concepts removed, and \(r_{\mathrm{ret}}\) is the fraction of non-target concepts retained. We also show the setting with the highest removal–retention utility in the \(0.9\)–\(1.0\) interval and the no-intervention condition defined above.

\begin{table*}[t]
\caption{\textbf{Quantitative concept suppression: removal--retention utility.} Results on $2{,}955$ COCO image--concept pairs. Columns show the selected fraction of positive attribution mass. Utility is target removal $\times$ non-target retention. No intervention uses the reconstructed representation without modifying features; all other rows use steering strength $s$. \ours rows are shaded. The best and second-best utility in each column are shown in \textbf{bold} and \underline{underlined}, respectively.}
  \label{tab:steering_utility_grid}
  \centering
  \scriptsize
  \setlength{\tabcolsep}{3.3pt}
  \begin{tabularx}{\textwidth}{@{}lX*{5}{S[table-format=3.2]}l@{}}
    \toprule
    \textbf{Intervention} & \textbf{Method}
    & \multicolumn{1}{c}{$m=20\%$}
    & \multicolumn{1}{c}{$m=40\%$}
    & \multicolumn{1}{c}{$m=60\%$}
    & \multicolumn{1}{c}{$m=80\%$}
    & \multicolumn{1}{c}{$m=100\%$} & \\
    \midrule
    \multirow{6}{*}{No intervention} & Original & \num{1.05} & \num{1.05} & \num{1.05} & \num{1.05} & \num{1.05} & \\
     & BatchTopK & \num{2.51} & \num{2.51} & \num{2.51} & \num{2.51} & \num{2.51} & \\
     & MSAE & \num{3.07} & \num{3.07} & \num{3.07} & \num{3.07} & \num{3.07} & \\
     & MpSAE & \num{2.03} & \num{2.03} & \num{2.03} & \num{2.03} & \num{2.03} & \\
     & ASAE & \num{7.26} & \num{7.26} & \num{7.26} & \num{7.26} & \num{7.26} & \\
    \cmidrule(lr){2-7}
    & \tikz[remember picture,baseline]\node[inner sep=0pt,outer sep=0pt,minimum size=0pt] (tabsteering_utility_grid_nointervention_oursbeg) {}; \ours\ \textit{(ours)} & \num{0.89} & \num{0.89} & \num{0.89} & \num{0.89} & \num{0.89} & {\rlap{\tikz[remember picture,baseline]\node[inner sep=0pt,outer sep=0pt,minimum size=0pt] (tabsteering_utility_grid_nointervention_oursend) {};}} \\
    \midrule
    \multirow{6}{*}{$0$}  & Original & \num{6.75} & \num{8.03} & \num{10.67} & \num{12.21} & \num{14.66} & \\
     & BatchTopK & \num{5.80} & \num{6.97} & \num{7.93} & \num{8.20} & \num{8.51} & \\
     & MSAE & \num{6.23} & \num{7.53} & \num{7.96} & \num{8.26} & \num{8.61} & \\
     & MpSAE & \num{6.48} & \num{7.70} & \num{8.58} & \num{9.05} & \num{9.47} & \\
     & ASAE & \num{7.46} & \num{7.13} & \num{7.33} & \num{7.37} & \num{7.99} & \\
    \cmidrule(lr){2-7}
    & \tikz[remember picture,baseline]\node[inner sep=0pt,outer sep=0pt,minimum size=0pt] (tabsteering_utility_grid_zeroablation_oursbeg) {}; \ours\ \textit{(ours)} & \num{9.02} & \num{17.30} & \uline{\num{26.88}} & \bfseries \num{36.25} & \bfseries \num{38.89} & {\rlap{\tikz[remember picture,baseline]\node[inner sep=0pt,outer sep=0pt,minimum size=0pt] (tabsteering_utility_grid_zeroablation_oursend) {};}} \\
    \midrule
    \multirow{6}{*}{$-0.5$} & Original & \num{6.44} & \num{7.96} & \num{10.38} & \num{11.87} & \num{17.44} & \\
     & BatchTopK & \num{3.26} & \num{2.90} & \num{4.29} & \num{8.36} & \num{2.63} & \\
     & MSAE & \num{3.28} & \num{2.74} & \num{2.99} & \num{7.28} & \num{7.98} & \\
     & MpSAE & \num{12.74} & \num{7.36} & \num{9.07} & \num{7.19} & \num{1.25} & \\
     & ASAE & \num{11.36} & \num{17.40} & \num{20.22} & \num{4.24} & \num{8.88} & \\
    \cmidrule(lr){2-7}
    & \tikz[remember picture,baseline]\node[inner sep=0pt,outer sep=0pt,minimum size=0pt] (tabsteering_utility_grid_neg0p5_oursbeg) {}; \ours\ \textit{(ours)} & \num{8.16} & \num{11.52} & \num{18.52} & \num{11.23} & \num{5.26} & {\rlap{\tikz[remember picture,baseline]\node[inner sep=0pt,outer sep=0pt,minimum size=0pt] (tabsteering_utility_grid_neg0p5_oursend) {};}} \\
    \midrule
    \multirow{6}{*}{$+0.5$} & Original & \num{7.67} & \num{11.15} & \num{15.33} & \uline{\num{20.90}} & \uline{\num{22.86}} & \\
     & BatchTopK & \num{14.14} & \uline{\num{28.85}} & \num{26.21} & \num{3.22} & \num{0.84} & \\
     & MSAE & \num{17.02} & \num{28.31} & \bfseries \num{27.21} & \num{5.20} & \num{0.35} & \\
     & MpSAE & \bfseries \num{31.94} & \num{6.00} & \num{0.10} & \num{0.08} & \num{0.21} & \\
     & ASAE & \num{12.08} & \num{20.70} & \num{13.65} & \num{3.41} & \num{0.03} & \\
    \cmidrule(lr){2-7}
    & \tikz[remember picture,baseline]\node[inner sep=0pt,outer sep=0pt,minimum size=0pt] (tabsteering_utility_grid_pos0p5_oursbeg) {}; \ours\ \textit{(ours)} & \num{16.94} & \bfseries \num{29.61} & \num{8.02} & \num{3.16} & \num{3.59} & {\rlap{\tikz[remember picture,baseline]\node[inner sep=0pt,outer sep=0pt,minimum size=0pt] (tabsteering_utility_grid_pos0p5_oursend) {};}} \\
    \midrule
    \multirow{6}{*}{$-1$} & Original & \num{7.17} & \num{10.21} & \num{14.50} & \num{19.50} & \num{16.07} & \\
     & BatchTopK & \num{2.50} & \num{3.48} & \num{5.00} & \num{5.58} & \num{1.54} & \\
     & MSAE & \num{5.35} & \num{5.35} & \num{5.14} & \num{6.15} & \num{7.46} & \\
     & MpSAE & \num{12.58} & \num{7.94} & \num{8.00} & \num{6.25} & \num{1.49} & \\
     & ASAE & \num{17.55} & \num{19.13} & \num{6.18} & \num{4.48} & \num{5.34} & \\
    \cmidrule(lr){2-7}
    & \tikz[remember picture,baseline]\node[inner sep=0pt,outer sep=0pt,minimum size=0pt] (tabsteering_utility_grid_neg1_oursbeg) {}; \ours\ \textit{(ours)} & \num{12.50} & \num{11.13} & \num{6.92} & \num{5.00} & \num{4.42} & {\rlap{\tikz[remember picture,baseline]\node[inner sep=0pt,outer sep=0pt,minimum size=0pt] (tabsteering_utility_grid_neg1_oursend) {};}} \\
    \midrule
    \multirow{6}{*}{$+1$} & Original & \num{9.32} & \num{18.06} & \num{20.75} & \num{13.59} & \num{5.59} & \\
     & BatchTopK & \uline{\num{30.40}} & \num{24.95} & \num{5.71} & \num{2.36} & \num{0.73} & \\
     & MSAE & \num{28.58} & \num{27.25} & \num{9.76} & \num{1.78} & \num{0.33} & \\
     & MpSAE & \num{28.39} & \num{1.51} & \num{0.00} & \num{0.00} & \num{0.27} & \\
     & ASAE & \num{19.60} & \num{15.57} & \num{4.80} & \num{0.72} & \num{0.00} & \\
    \cmidrule(lr){2-7}
    & \tikz[remember picture,baseline]\node[inner sep=0pt,outer sep=0pt,minimum size=0pt] (tabsteering_utility_grid_pos1_oursbeg) {}; \ours\ \textit{(ours)} & \num{26.09} & \num{5.65} & \num{2.86} & \num{3.21} & \num{2.82} & {\rlap{\tikz[remember picture,baseline]\node[inner sep=0pt,outer sep=0pt,minimum size=0pt] (tabsteering_utility_grid_pos1_oursend) {};}} \\
    \midrule
    \multirow{6}{*}{$-2.5$} & Original & \num{17.47} & \num{10.54} & \num{7.37} & \num{6.42} & \num{6.26} & \\
     & BatchTopK & \num{4.56} & \num{2.85} & \num{3.96} & \num{4.91} & \num{1.42} & \\
     & MSAE & \num{6.96} & \num{3.07} & \num{3.64} & \num{4.85} & \num{7.15} & \\
     & MpSAE & \num{3.95} & \num{8.64} & \num{7.12} & \num{5.24} & \num{1.53} & \\
     & ASAE & \num{5.63} & \num{4.46} & \num{5.85} & \num{6.72} & \num{2.60} & \\
    \cmidrule(lr){2-7}
    & \tikz[remember picture,baseline]\node[inner sep=0pt,outer sep=0pt,minimum size=0pt] (tabsteering_utility_grid_neg2p5_oursbeg) {}; \ours\ \textit{(ours)} & \num{6.54} & \num{5.21} & \num{5.06} & \num{4.81} & \num{2.66} & {\rlap{\tikz[remember picture,baseline]\node[inner sep=0pt,outer sep=0pt,minimum size=0pt] (tabsteering_utility_grid_neg2p5_oursend) {};}} \\
    \midrule
    \multirow{6}{*}{$+2.5$} & Original & \num{17.78} & \num{5.38} & \num{2.96} & \num{2.53} & \num{2.58} & \\
     & BatchTopK & \num{15.48} & \num{4.68} & \num{1.97} & \num{3.03} & \num{0.62} & \\
     & MSAE & \num{15.60} & \num{3.98} & \num{2.04} & \num{0.98} & \num{0.27} & \\
     & MpSAE & \num{4.91} & \num{0.09} & \num{0.00} & \num{0.05} & \num{0.06} & \\
     & ASAE & \num{6.56} & \num{3.49} & \num{0.81} & \num{0.12} & \num{0.00} & \\
    \cmidrule(lr){2-7}
    & \tikz[remember picture,baseline]\node[inner sep=0pt,outer sep=0pt,minimum size=0pt] (tabsteering_utility_grid_pos2p5_oursbeg) {}; \ours\ \textit{(ours)} & \num{3.88} & \num{2.78} & \num{2.36} & \num{1.92} & \num{1.57} & {\rlap{\tikz[remember picture,baseline]\node[inner sep=0pt,outer sep=0pt,minimum size=0pt] (tabsteering_utility_grid_pos2p5_oursend) {};}} \\
    \midrule
    \multirow{6}{*}{$-5$} & Original & \num{8.63} & \num{6.07} & \num{5.59} & \num{4.63} & \num{4.78} & \\
     & BatchTopK & \num{2.29} & \num{2.74} & \num{4.44} & \num{4.47} & \num{1.20} & \\
     & MSAE & \num{2.58} & \num{3.10} & \num{3.87} & \num{4.68} & \num{6.89} & \\
     & MpSAE & \num{5.38} & \num{8.57} & \num{7.03} & \num{4.99} & \num{1.59} & \\
     & ASAE & \num{1.19} & \num{3.10} & \num{5.18} & \num{5.75} & \num{1.79} & \\
    \cmidrule(lr){2-7}
    & \tikz[remember picture,baseline]\node[inner sep=0pt,outer sep=0pt,minimum size=0pt] (tabsteering_utility_grid_neg5_oursbeg) {}; \ours\ \textit{(ours)} & \num{5.23} & \num{4.65} & \num{4.50} & \num{3.78} & \num{0.00} & {\rlap{\tikz[remember picture,baseline]\node[inner sep=0pt,outer sep=0pt,minimum size=0pt] (tabsteering_utility_grid_neg5_oursend) {};}} \\
    \midrule
    \multirow{6}{*}{$+5$} & Original & \num{4.18} & \num{2.86} & \num{2.35} & \num{2.60} & \num{2.17} & \\
     & BatchTopK & \num{2.98} & \num{2.38} & \num{1.27} & \num{2.32} & \num{0.41} & \\
     & MSAE & \num{4.03} & \num{1.67} & \num{0.97} & \num{0.77} & \num{0.33} & \\
     & MpSAE & \num{1.49} & \num{0.00} & \num{0.00} & \num{0.05} & \num{0.21} & \\
     & ASAE & \num{2.86} & \num{1.06} & \num{0.06} & \num{0.06} & \num{0.00} & \\
    \cmidrule(lr){2-7}
    & \tikz[remember picture,baseline]\node[inner sep=0pt,outer sep=0pt,minimum size=0pt] (tabsteering_utility_grid_pos5_oursbeg) {}; \ours\ \textit{(ours)} & \num{3.11} & \num{2.51} & \num{2.62} & \num{1.86} & \num{0.00} & {\rlap{\tikz[remember picture,baseline]\node[inner sep=0pt,outer sep=0pt,minimum size=0pt] (tabsteering_utility_grid_pos5_oursend) {};}} \\
    \bottomrule
  \end{tabularx}
  \begin{tikzpicture}[remember picture,overlay]
    \begin{scope}[on background layer]
      \fill[black,opacity=0.10] ([xshift=0,yshift=2.7ex]tabsteering_utility_grid_nointervention_oursbeg.north west) rectangle ([xshift=0,yshift=-1.4ex]tabsteering_utility_grid_nointervention_oursend.south east);
      \fill[black,opacity=0.10] ([xshift=0,yshift=2.7ex]tabsteering_utility_grid_zeroablation_oursbeg.north west) rectangle ([xshift=0,yshift=-1.4ex]tabsteering_utility_grid_zeroablation_oursend.south east);
      \fill[black,opacity=0.10] ([xshift=0,yshift=2.7ex]tabsteering_utility_grid_neg0p5_oursbeg.north west) rectangle ([xshift=0,yshift=-1.4ex]tabsteering_utility_grid_neg0p5_oursend.south east);
      \fill[black,opacity=0.10] ([xshift=0,yshift=2.7ex]tabsteering_utility_grid_pos0p5_oursbeg.north west) rectangle ([xshift=0,yshift=-1.4ex]tabsteering_utility_grid_pos0p5_oursend.south east);
      \fill[black,opacity=0.10] ([xshift=0,yshift=2.7ex]tabsteering_utility_grid_neg1_oursbeg.north west) rectangle ([xshift=0,yshift=-1.4ex]tabsteering_utility_grid_neg1_oursend.south east);
      \fill[black,opacity=0.10]
        ([xshift=0,yshift=2.7ex]tabsteering_utility_grid_pos1_oursbeg.north west)
        rectangle ([xshift=0,yshift=-1.4ex]tabsteering_utility_grid_pos1_oursend.south east);
      \fill[black,opacity=0.10]
        ([xshift=0,yshift=2.7ex]tabsteering_utility_grid_neg2p5_oursbeg.north west)
        rectangle ([xshift=0,yshift=-1.4ex]tabsteering_utility_grid_neg2p5_oursend.south east);
      \fill[black,opacity=0.10]
        ([xshift=0,yshift=2.7ex]tabsteering_utility_grid_pos2p5_oursbeg.north west)
        rectangle ([xshift=0,yshift=-1.4ex]tabsteering_utility_grid_pos2p5_oursend.south east);
      \fill[black,opacity=0.10]
        ([xshift=0,yshift=2.7ex]tabsteering_utility_grid_neg5_oursbeg.north west)
        rectangle ([xshift=0,yshift=-1.4ex]tabsteering_utility_grid_neg5_oursend.south east);
      \fill[black,opacity=0.10]
        ([xshift=0,yshift=2.7ex]tabsteering_utility_grid_pos5_oursbeg.north west)
        rectangle ([xshift=0,yshift=-1.4ex]tabsteering_utility_grid_pos5_oursend.south east);
    \end{scope}
  \end{tikzpicture}
\end{table*}
\begin{table*}[t]
\caption{\textbf{Quantitative concept suppression: non-target retention.}
Results on $2{,}955$ COCO image--concept pairs. Values are the percentage of non-target COCO concepts from the original caption that remain after intervention. Columns show the selected fraction of positive attribution mass. No intervention uses the reconstructed representation without modifying features; all other rows use steering strength $s$. \ours rows are shaded.}
  \label{tab:steering_non_target_retention_grid}
  \centering
  \scriptsize
  \setlength{\tabcolsep}{3.3pt}
  \begin{tabularx}{\textwidth}{@{}lX*{5}{S[table-format=3.2]}l@{}}
    \toprule
    \textbf{Intervention} & \textbf{Method}
    & \multicolumn{1}{c}{$m=20\%$}
    & \multicolumn{1}{c}{$m=40\%$}
    & \multicolumn{1}{c}{$m=60\%$}
    & \multicolumn{1}{c}{$m=80\%$}
    & \multicolumn{1}{c}{$m=100\%$} & \\
    \midrule
    \multirow{6}{*}{No intervention} & Original & \num{97.17} & \num{97.17} & \num{97.17} & \num{97.17} & \num{97.17} & \\
     & BatchTopK & \num{92.58} & \num{92.58} & \num{92.58} & \num{92.58} & \num{92.58} & \\
     & MSAE & \num{91.69} & \num{91.69} & \num{91.69} & \num{91.69} & \num{91.69} & \\
     & MpSAE & \num{95.11} & \num{95.11} & \num{95.11} & \num{95.11} & \num{95.11} & \\
     & ASAE & \num{79.75} & \num{79.75} & \num{79.75} & \num{79.75} & \num{79.75} & \\
    \cmidrule(lr){2-7}
    & \tikz[remember picture,baseline]\node[inner sep=0pt,outer sep=0pt,minimum size=0pt] (tabsteering_non_target_retention_grid_nointervention_oursbeg) {}; \ours\ \textit{(ours)} & \num{97.10} & \num{97.10} & \num{97.10} & \num{97.10} & \num{97.10} & {\rlap{\tikz[remember picture,baseline]\node[inner sep=0pt,outer sep=0pt,minimum size=0pt] (tabsteering_non_target_retention_grid_nointervention_oursend) {};}} \\
    \midrule
    \multirow{6}{*}{$0$}  & Original & \num{84.57} & \num{83.00} & \num{81.03} & \num{78.95} & \num{77.65} & \\
     & BatchTopK & \num{87.07} & \num{83.74} & \num{81.09} & \num{80.74} & \num{79.37} & \\
     & MSAE & \num{84.78} & \num{81.48} & \num{80.32} & \num{79.25} & \num{79.52} & \\
     & MpSAE & \num{85.47} & \num{83.30} & \num{82.30} & \num{81.57} & \num{79.47} & \\
     & ASAE & \num{79.61} & \num{79.48} & \num{78.45} & \num{78.94} & \num{78.14} & \\
    \cmidrule(lr){2-7}
    & \tikz[remember picture,baseline]\node[inner sep=0pt,outer sep=0pt,minimum size=0pt] (tabsteering_non_target_retention_grid_zeroablation_oursbeg) {}; \ours\ \textit{(ours)} & \num{83.59} & \num{80.90} & \num{75.09} & \num{66.99} & \num{56.25} & {\rlap{\tikz[remember picture,baseline]\node[inner sep=0pt,outer sep=0pt,minimum size=0pt] (tabsteering_non_target_retention_grid_zeroablation_oursend) {};}} \\
    \midrule
    \multirow{6}{*}{$-0.5$} & Original & \num{82.69} & \num{79.21} & \num{78.28} & \num{75.26} & \num{70.02} & \\
     & BatchTopK & \num{81.70} & \num{77.33} & \num{63.43} & \num{27.10} & \num{2.67} & \\
     & MSAE & \num{81.33} & \num{77.95} & \num{67.93} & \num{36.64} & \num{8.26} & \\
     & MpSAE & \num{59.49} & \num{8.14} & \num{9.27} & \num{7.25} & \num{1.26} & \\
     & ASAE & \num{72.79} & \num{60.35} & \num{32.38} & \num{4.41} & \num{9.14} & \\
    \cmidrule(lr){2-7}
    & \tikz[remember picture,baseline]\node[inner sep=0pt,outer sep=0pt,minimum size=0pt] (tabsteering_non_target_retention_grid_neg0p5_oursbeg) {}; \ours\ \textit{(ours)} & \num{79.34} & \num{69.22} & \num{45.11} & \num{13.91} & \num{5.50} & {\rlap{\tikz[remember picture,baseline]\node[inner sep=0pt,outer sep=0pt,minimum size=0pt] (tabsteering_non_target_retention_grid_neg0p5_oursend) {};}} \\
    \midrule
    \multirow{6}{*}{$+0.5$} & Original & \num{81.22} & \num{78.28} & \num{73.43} & \num{64.33} & \num{41.27} & \\
     & BatchTopK & \num{78.39} & \num{65.33} & \num{31.24} & \num{3.28} & \num{0.85} & \\
     & MSAE & \num{72.59} & \num{59.12} & \num{34.63} & \num{5.33} & \num{0.35} & \\
     & MpSAE & \num{60.43} & \num{6.13} & \num{0.10} & \num{0.08} & \num{0.21} & \\
     & ASAE & \num{68.37} & \num{45.95} & \num{16.78} & \num{3.51} & \num{0.03} & \\
    \cmidrule(lr){2-7}
    & \tikz[remember picture,baseline]\node[inner sep=0pt,outer sep=0pt,minimum size=0pt] (tabsteering_non_target_retention_grid_pos0p5_oursbeg) {}; \ours\ \textit{(ours)} & \num{75.72} & \num{40.33} & \num{8.40} & \num{3.21} & \num{3.68} & {\rlap{\tikz[remember picture,baseline]\node[inner sep=0pt,outer sep=0pt,minimum size=0pt] (tabsteering_non_target_retention_grid_pos0p5_oursend) {};}} \\
    \midrule
    \multirow{6}{*}{$-1$} & Original & \num{77.60} & \num{72.91} & \num{60.42} & \num{39.59} & \num{21.51} & \\
     & BatchTopK & \num{68.27} & \num{49.98} & \num{20.28} & \num{8.40} & \num{1.56} & \\
     & MSAE & \num{65.05} & \num{44.63} & \num{22.32} & \num{9.76} & \num{7.76} & \\
     & MpSAE & \num{25.82} & \num{8.35} & \num{8.15} & \num{6.31} & \num{1.50} & \\
     & ASAE & \num{59.21} & \num{31.34} & \num{6.69} & \num{4.55} & \num{5.45} & \\
    \cmidrule(lr){2-7}
    & \tikz[remember picture,baseline]\node[inner sep=0pt,outer sep=0pt,minimum size=0pt] (tabsteering_non_target_retention_grid_neg1_oursbeg) {}; \ours\ \textit{(ours)} & \num{59.21} & \num{14.14} & \num{7.33} & \num{5.20} & \num{4.57} & {\rlap{\tikz[remember picture,baseline]\node[inner sep=0pt,outer sep=0pt,minimum size=0pt] (tabsteering_non_target_retention_grid_neg1_oursend) {};}} \\
    \midrule
    \multirow{6}{*}{$+1$} & Original & \num{77.54} & \num{66.07} & \num{41.28} & \num{17.97} & \num{5.88} & \\
     & BatchTopK & \num{64.03} & \num{29.00} & \num{5.84} & \num{2.40} & \num{0.74} & \\
     & MSAE & \num{59.94} & \num{34.44} & \num{10.19} & \num{1.81} & \num{0.33} & \\
     & MpSAE & \num{31.58} & \num{1.52} & \num{0.00} & \num{0.00} & \num{0.27} & \\
     & ASAE & \num{47.72} & \num{21.30} & \num{5.03} & \num{0.73} & \num{0.00} & \\
    \cmidrule(lr){2-7}
    & \tikz[remember picture,baseline]\node[inner sep=0pt,outer sep=0pt,minimum size=0pt] (tabsteering_non_target_retention_grid_pos1_oursbeg) {}; \ours\ \textit{(ours)} & \num{43.78} & \num{5.87} & \num{2.91} & \num{3.26} & \num{2.87} & {\rlap{\tikz[remember picture,baseline]\node[inner sep=0pt,outer sep=0pt,minimum size=0pt] (tabsteering_non_target_retention_grid_pos1_oursend) {};}} \\
    \midrule
    \multirow{6}{*}{$-2.5$} & Original & \num{45.57} & \num{11.80} & \num{7.67} & \num{6.59} & \num{6.44} & \\
     & BatchTopK & \num{12.27} & \num{4.78} & \num{5.18} & \num{5.97} & \num{1.44} & \\
     & MSAE & \num{21.80} & \num{5.67} & \num{5.13} & \num{6.06} & \num{7.41} & \\
     & MpSAE & \num{5.11} & \num{8.91} & \num{7.23} & \num{5.27} & \num{1.54} & \\
     & ASAE & \num{6.27} & \num{4.69} & \num{5.97} & \num{6.84} & \num{2.63} & \\
    \cmidrule(lr){2-7}
    & \tikz[remember picture,baseline]\node[inner sep=0pt,outer sep=0pt,minimum size=0pt] (tabsteering_non_target_retention_grid_neg2p5_oursbeg) {}; \ours\ \textit{(ours)} & \num{6.94} & \num{5.44} & \num{5.29} & \num{5.01} & \num{2.73} & {\rlap{\tikz[remember picture,baseline]\node[inner sep=0pt,outer sep=0pt,minimum size=0pt] (tabsteering_non_target_retention_grid_neg2p5_oursend) {};}} \\
    \midrule
    \multirow{6}{*}{$+2.5$} & Original & \num{32.86} & \num{5.95} & \num{3.06} & \num{2.60} & \num{2.64} & \\
     & BatchTopK & \num{17.17} & \num{4.76} & \num{2.00} & \num{3.09} & \num{0.63} & \\
     & MSAE & \num{18.14} & \num{4.05} & \num{2.05} & \num{0.99} & \num{0.27} & \\
     & MpSAE & \num{4.95} & \num{0.09} & \num{0.00} & \num{0.05} & \num{0.06} & \\
     & ASAE & \num{7.14} & \num{3.63} & \num{0.82} & \num{0.12} & \num{0.00} & \\
    \cmidrule(lr){2-7}
    & \tikz[remember picture,baseline]\node[inner sep=0pt,outer sep=0pt,minimum size=0pt] (tabsteering_non_target_retention_grid_pos2p5_oursbeg) {}; \ours\ \textit{(ours)} & \num{3.98} & \num{2.83} & \num{2.39} & \num{1.95} & \num{1.59} & {\rlap{\tikz[remember picture,baseline]\node[inner sep=0pt,outer sep=0pt,minimum size=0pt] (tabsteering_non_target_retention_grid_pos2p5_oursend) {};}} \\
    \midrule
    \multirow{6}{*}{$-5$} & Original & \num{9.12} & \num{6.26} & \num{5.76} & \num{4.75} & \num{4.91} & \\
     & BatchTopK & \num{3.88} & \num{3.81} & \num{5.39} & \num{5.38} & \num{1.22} & \\
     & MSAE & \num{3.83} & \num{4.56} & \num{5.01} & \num{5.71} & \num{7.13} & \\
     & MpSAE & \num{6.29} & \num{8.82} & \num{7.14} & \num{5.03} & \num{1.60} & \\
     & ASAE & \num{1.23} & \num{3.15} & \num{5.24} & \num{5.83} & \num{1.80} & \\
    \cmidrule(lr){2-7}
    & \tikz[remember picture,baseline]\node[inner sep=0pt,outer sep=0pt,minimum size=0pt] (tabsteering_non_target_retention_grid_neg5_oursbeg) {}; \ours\ \textit{(ours)} & \num{5.40} & \num{4.84} & \num{4.73} & \num{3.95} & \num{0.00} & {\rlap{\tikz[remember picture,baseline]\node[inner sep=0pt,outer sep=0pt,minimum size=0pt] (tabsteering_non_target_retention_grid_neg5_oursend) {};}} \\
    \midrule
    \multirow{6}{*}{$+5$} & Original & \num{4.33} & \num{2.91} & \num{2.40} & \num{2.66} & \num{2.21} & \\
     & BatchTopK & \num{3.03} & \num{2.41} & \num{1.28} & \num{2.36} & \num{0.41} & \\
     & MSAE & \num{4.13} & \num{1.69} & \num{0.97} & \num{0.78} & \num{0.33} & \\
     & MpSAE & \num{1.49} & \num{0.00} & \num{0.00} & \num{0.05} & \num{0.21} & \\
     & ASAE & \num{2.93} & \num{1.08} & \num{0.06} & \num{0.06} & \num{0.00} & \\
    \cmidrule(lr){2-7}
    & \tikz[remember picture,baseline]\node[inner sep=0pt,outer sep=0pt,minimum size=0pt] (tabsteering_non_target_retention_grid_pos5_oursbeg) {}; \ours\ \textit{(ours)} & \num{3.17} & \num{2.55} & \num{2.66} & \num{1.89} & \num{0.00} & {\rlap{\tikz[remember picture,baseline]\node[inner sep=0pt,outer sep=0pt,minimum size=0pt] (tabsteering_non_target_retention_grid_pos5_oursend) {};}} \\
    \bottomrule
  \end{tabularx}
  \begin{tikzpicture}[remember picture,overlay]
    \begin{scope}[on background layer]
      \fill[black,opacity=0.10] ([xshift=0,yshift=2.7ex]tabsteering_non_target_retention_grid_nointervention_oursbeg.north west) rectangle ([xshift=0,yshift=-1.4ex]tabsteering_non_target_retention_grid_nointervention_oursend.south east);
      \fill[black,opacity=0.10] ([xshift=0,yshift=2.7ex]tabsteering_non_target_retention_grid_zeroablation_oursbeg.north west) rectangle ([xshift=0,yshift=-1.4ex]tabsteering_non_target_retention_grid_zeroablation_oursend.south east);
      \fill[black,opacity=0.10] ([xshift=0,yshift=2.7ex]tabsteering_non_target_retention_grid_neg0p5_oursbeg.north west) rectangle ([xshift=0,yshift=-1.4ex]tabsteering_non_target_retention_grid_neg0p5_oursend.south east);
      \fill[black,opacity=0.10] ([xshift=0,yshift=2.7ex]tabsteering_non_target_retention_grid_pos0p5_oursbeg.north west) rectangle ([xshift=0,yshift=-1.4ex]tabsteering_non_target_retention_grid_pos0p5_oursend.south east);
      \fill[black,opacity=0.10] ([xshift=0,yshift=2.7ex]tabsteering_non_target_retention_grid_neg1_oursbeg.north west) rectangle ([xshift=0,yshift=-1.4ex]tabsteering_non_target_retention_grid_neg1_oursend.south east);
      \fill[black,opacity=0.10]
        ([xshift=0,yshift=2.7ex]tabsteering_non_target_retention_grid_pos1_oursbeg.north west)
        rectangle ([xshift=0,yshift=-1.4ex]tabsteering_non_target_retention_grid_pos1_oursend.south east);
      \fill[black,opacity=0.10]
        ([xshift=0,yshift=2.7ex]tabsteering_non_target_retention_grid_neg2p5_oursbeg.north west)
        rectangle ([xshift=0,yshift=-1.4ex]tabsteering_non_target_retention_grid_neg2p5_oursend.south east);
      \fill[black,opacity=0.10]
        ([xshift=0,yshift=2.7ex]tabsteering_non_target_retention_grid_pos2p5_oursbeg.north west)
        rectangle ([xshift=0,yshift=-1.4ex]tabsteering_non_target_retention_grid_pos2p5_oursend.south east);
      \fill[black,opacity=0.10]
        ([xshift=0,yshift=2.7ex]tabsteering_non_target_retention_grid_neg5_oursbeg.north west)
        rectangle ([xshift=0,yshift=-1.4ex]tabsteering_non_target_retention_grid_neg5_oursend.south east);
      \fill[black,opacity=0.10]
        ([xshift=0,yshift=2.7ex]tabsteering_non_target_retention_grid_pos5_oursbeg.north west)
        rectangle ([xshift=0,yshift=-1.4ex]tabsteering_non_target_retention_grid_pos5_oursend.south east);
    \end{scope}
  \end{tikzpicture}
\end{table*}
\begin{table*}[t]
    \caption{\textbf{Quantitative concept suppression: target removal.}
    Results on $2{,}955$ COCO image--concept pairs. Values are the percentage of edited captions for which the VLM judge finds the target concept or a close synonym absent. Columns show the selected fraction of positive attribution mass. No intervention uses the reconstructed representation without modifying features; all other rows use steering strength $s$. \ours rows are shaded.}
  \label{tab:steering_judge_removal_grid}
  \centering
  \scriptsize
  \setlength{\tabcolsep}{3.3pt}
  \begin{tabularx}{\textwidth}{@{}lX*{5}{S[table-format=3.2]}l@{}}
    \toprule
    \textbf{Intervention} & \textbf{Method}
    & \multicolumn{1}{c}{$m=20\%$}
    & \multicolumn{1}{c}{$m=40\%$}
    & \multicolumn{1}{c}{$m=60\%$}
    & \multicolumn{1}{c}{$m=80\%$}
    & \multicolumn{1}{c}{$m=100\%$} & \\
    \midrule
    \multirow{6}{*}{No intervention} & Original & \num{1.08} & \num{1.08} & \num{1.08} & \num{1.08} & \num{1.08} & \\
     & BatchTopK & \num{2.71} & \num{2.71} & \num{2.71} & \num{2.71} & \num{2.71} & \\
     & MSAE & \num{3.35} & \num{3.35} & \num{3.35} & \num{3.35} & \num{3.35} & \\
     & MpSAE & \num{2.13} & \num{2.13} & \num{2.13} & \num{2.13} & \num{2.13} & \\
     & ASAE & \num{9.10} & \num{9.10} & \num{9.10} & \num{9.10} & \num{9.10} & \\
    \cmidrule(lr){2-7}
    & \tikz[remember picture,baseline]\node[inner sep=0pt,outer sep=0pt,minimum size=0pt] (tabsteering_judge_removal_grid_nointervention_oursbeg) {}; \ours\ \textit{(ours)} & \num{0.91} & \num{0.91} & \num{0.91} & \num{0.91} & \num{0.91} & {\rlap{\tikz[remember picture,baseline]\node[inner sep=0pt,outer sep=0pt,minimum size=0pt] (tabsteering_judge_removal_grid_nointervention_oursend) {};}} \\
    \midrule
    \multirow{6}{*}{$0$} & Original & \num{7.99} & \num{9.68} & \num{13.16} & \num{15.47} & \num{18.88} & \\
     & BatchTopK & \num{6.67} & \num{8.32} & \num{9.78} & \num{10.15} & \num{10.73} & \\
     & MSAE & \num{7.34} & \num{9.24} & \num{9.92} & \num{10.42} & \num{10.83} & \\
     & MpSAE & \num{7.58} & \num{9.24} & \num{10.42} & \num{11.10} & \num{11.91} & \\
     & ASAE & \num{9.37} & \num{8.97} & \num{9.34} & \num{9.34} & \num{10.22} & \\
    \cmidrule(lr){2-7}
    & \tikz[remember picture,baseline]\node[inner sep=0pt,outer sep=0pt,minimum size=0pt] (tabsteering_judge_removal_grid_zeroablation_oursbeg) {}; \ours\ \textit{(ours)} & \num{10.80} & \num{21.39} & \num{35.80} & \num{54.11} & \num{69.14} & {\rlap{\tikz[remember picture,baseline]\node[inner sep=0pt,outer sep=0pt,minimum size=0pt] (tabsteering_judge_removal_grid_zeroablation_oursend) {};}} \\
    \midrule
    \multirow{6}{*}{$-0.5$} & Original & \num{7.78} & \num{10.05} & \num{13.27} & \num{15.77} & \num{24.91} & \\
     & BatchTopK & \num{3.99} & \num{3.76} & \num{6.77} & \num{30.83} & \num{98.48} & \\
     & MSAE & \num{4.03} & \num{3.52} & \num{4.40} & \num{19.86} & \num{96.68} & \\
     & MpSAE & \num{21.42} & \num{90.52} & \num{97.87} & \num{99.15} & \num{99.09} & \\
     & ASAE & \num{15.60} & \num{28.83} & \num{62.44} & \num{96.21} & \num{97.12} & \\
    \cmidrule(lr){2-7}
    & \tikz[remember picture,baseline]\node[inner sep=0pt,outer sep=0pt,minimum size=0pt] (tabsteering_judge_removal_grid_neg0p5_oursbeg) {}; \ours\ \textit{(ours)} & \num{10.29} & \num{16.65} & \num{41.05} & \num{80.78} & \num{95.63} & {\rlap{\tikz[remember picture,baseline]\node[inner sep=0pt,outer sep=0pt,minimum size=0pt] (tabsteering_judge_removal_grid_neg0p5_oursend) {};}} \\
    \midrule
    \multirow{6}{*}{$+0.5$} & Original & \num{9.44} & \num{14.25} & \num{20.88} & \num{32.49} & \num{55.40} & \\
     & BatchTopK & \num{18.04} & \num{44.16} & \num{83.93} & \num{98.31} & \num{99.36} & \\
     & MSAE & \num{23.45} & \num{47.88} & \num{78.58} & \num{97.73} & \num{99.29} & \\
     & MpSAE & \num{52.86} & \num{97.97} & \num{99.90} & \num{99.90} & \num{98.65} & \\
     & ASAE & \num{17.66} & \num{45.04} & \num{81.35} & \num{97.33} & \num{99.56} & \\
    \cmidrule(lr){2-7}
    & \tikz[remember picture,baseline]\node[inner sep=0pt,outer sep=0pt,minimum size=0pt] (tabsteering_judge_removal_grid_pos0p5_oursbeg) {}; \ours\ \textit{(ours)} & \num{22.37} & \num{73.43} & \num{95.47} & \num{98.27} & \num{97.73} & {\rlap{\tikz[remember picture,baseline]\node[inner sep=0pt,outer sep=0pt,minimum size=0pt] (tabsteering_judge_removal_grid_pos0p5_oursend) {};}} \\
    \midrule
    \multirow{6}{*}{$-1$} & Original & \num{9.24} & \num{14.01} & \num{23.99} & \num{49.24} & \num{74.72} & \\
     & BatchTopK & \num{3.65} & \num{6.97} & \num{24.64} & \num{66.43} & \num{98.61} & \\
     & MSAE & \num{8.22} & \num{11.98} & \num{23.05} & \num{63.01} & \num{96.21} & \\
     & MpSAE & \num{48.73} & \num{95.03} & \num{98.17} & \num{99.05} & \num{98.92} & \\
     & ASAE & \num{29.64} & \num{61.05} & \num{92.32} & \num{98.38} & \num{98.04} & \\
    \cmidrule(lr){2-7}
    & \tikz[remember picture,baseline]\node[inner sep=0pt,outer sep=0pt,minimum size=0pt] (tabsteering_judge_removal_grid_neg1_oursbeg) {}; \ours\ \textit{(ours)} & \num{21.12} & \num{78.75} & \num{94.35} & \num{96.11} & \num{96.72} & {\rlap{\tikz[remember picture,baseline]\node[inner sep=0pt,outer sep=0pt,minimum size=0pt] (tabsteering_judge_removal_grid_neg1_oursend) {};}} \\
    \midrule
    \multirow{6}{*}{$+1$} & Original & \num{12.01} & \num{27.34} & \num{50.25} & \num{75.63} & \num{95.03} & \\
     & BatchTopK & \num{47.48} & \num{86.06} & \num{97.77} & \num{98.38} & \num{99.19} & \\
     & MSAE & \num{47.68} & \num{79.12} & \num{95.77} & \num{98.58} & \num{99.29} & \\
     & MpSAE & \num{89.88} & \num{99.22} & \num{99.76} & \num{99.83} & \num{98.88} & \\
     & ASAE & \num{41.08} & \num{73.10} & \num{95.43} & \num{98.68} & \num{99.83} & \\
    \cmidrule(lr){2-7}
    & \tikz[remember picture,baseline]\node[inner sep=0pt,outer sep=0pt,minimum size=0pt] (tabsteering_judge_removal_grid_pos1_oursbeg) {}; \ours\ \textit{(ours)} & \num{59.59} & \num{96.28} & \num{98.31} & \num{98.44} & \num{98.31} & {\rlap{\tikz[remember picture,baseline]\node[inner sep=0pt,outer sep=0pt,minimum size=0pt] (tabsteering_judge_removal_grid_pos1_oursend) {};}} \\
    \midrule
    \multirow{6}{*}{$-2.5$} & Original & \num{38.34} & \num{89.34} & \num{96.04} & \num{97.33} & \num{97.12} & \\
     & BatchTopK & \num{37.16} & \num{59.63} & \num{76.55} & \num{82.23} & \num{98.65} & \\
     & MSAE & \num{31.91} & \num{54.25} & \num{71.00} & \num{79.97} & \num{96.55} & \\
     & MpSAE & \num{77.36} & \num{97.02} & \num{98.48} & \num{99.46} & \num{98.95} & \\
     & ASAE & \num{89.85} & \num{94.96} & \num{97.94} & \num{98.34} & \num{98.92} & \\
    \cmidrule(lr){2-7}
    & \tikz[remember picture,baseline]\node[inner sep=0pt,outer sep=0pt,minimum size=0pt] (tabsteering_judge_removal_grid_neg2p5_oursbeg) {}; \ours\ \textit{(ours)} & \num{94.31} & \num{95.60} & \num{95.50} & \num{96.04} & \num{97.26} & {\rlap{\tikz[remember picture,baseline]\node[inner sep=0pt,outer sep=0pt,minimum size=0pt] (tabsteering_judge_removal_grid_neg2p5_oursend) {};}} \\
    \midrule
    \multirow{6}{*}{$+2.5$} & Original & \num{54.11} & \num{90.36} & \num{96.58} & \num{97.39} & \num{97.87} & \\
     & BatchTopK & \num{90.19} & \num{98.44} & \num{98.58} & \num{98.14} & \num{99.19} & \\
     & MSAE & \num{85.96} & \num{98.38} & \num{99.42} & \num{98.95} & \num{99.42} & \\
     & MpSAE & \num{99.22} & \num{99.53} & \num{99.63} & \num{99.56} & \num{98.51} & \\
     & ASAE & \num{91.95} & \num{96.07} & \num{98.88} & \num{99.49} & \num{99.80} & \\
    \cmidrule(lr){2-7}
    & \tikz[remember picture,baseline]\node[inner sep=0pt,outer sep=0pt,minimum size=0pt] (tabsteering_judge_removal_grid_pos2p5_oursbeg) {}; \ours\ \textit{(ours)} & \num{97.63} & \num{98.38} & \num{98.71} & \num{98.21} & \num{99.15} & {\rlap{\tikz[remember picture,baseline]\node[inner sep=0pt,outer sep=0pt,minimum size=0pt] (tabsteering_judge_removal_grid_pos2p5_oursend) {};}} \\
    \midrule
    \multirow{6}{*}{$-5$} & Original & \num{94.65} & \num{96.95} & \num{97.06} & \num{97.53} & \num{97.39} & \\
     & BatchTopK & \num{59.02} & \num{71.95} & \num{82.40} & \num{83.01} & \num{98.68} & \\
     & MSAE & \num{67.34} & \num{68.12} & \num{77.33} & \num{82.03} & \num{96.68} & \\
     & MpSAE & \num{85.41} & \num{97.19} & \num{98.48} & \num{99.36} & \num{98.98} & \\
     & ASAE & \num{97.06} & \num{98.48} & \num{98.82} & \num{98.75} & \num{98.95} & \\
    \cmidrule(lr){2-7}
    & \tikz[remember picture,baseline]\node[inner sep=0pt,outer sep=0pt,minimum size=0pt] (tabsteering_judge_removal_grid_neg5_oursbeg) {}; \ours\ \textit{(ours)} & \num{96.92} & \num{96.04} & \num{95.16} & \num{95.70} & \num{100.00} & {\rlap{\tikz[remember picture,baseline]\node[inner sep=0pt,outer sep=0pt,minimum size=0pt] (tabsteering_judge_removal_grid_neg5_oursend) {};}} \\
    \midrule
    \multirow{6}{*}{$+5$} & Original & \num{96.68} & \num{98.21} & \num{97.80} & \num{97.73} & \num{98.34} & \\
     & BatchTopK & \num{98.48} & \num{98.95} & \num{99.05} & \num{98.41} & \num{99.19} & \\
     & MSAE & \num{97.56} & \num{99.05} & \num{99.39} & \num{98.95} & \num{99.59} & \\
     & MpSAE & \num{99.70} & \num{99.32} & \num{99.63} & \num{99.66} & \num{98.27} & \\
     & ASAE & \num{97.43} & \num{98.27} & \num{99.29} & \num{99.56} & \num{99.83} & \\
    \cmidrule(lr){2-7}
    & \tikz[remember picture,baseline]\node[inner sep=0pt,outer sep=0pt,minimum size=0pt] (tabsteering_judge_removal_grid_pos5_oursbeg) {}; \ours\ \textit{(ours)} & \num{98.17} & \num{98.41} & \num{98.54} & \num{98.31} & \num{100.00} & {\rlap{\tikz[remember picture,baseline]\node[inner sep=0pt,outer sep=0pt,minimum size=0pt] (tabsteering_judge_removal_grid_pos5_oursend) {};}} \\
    \bottomrule
  \end{tabularx}
  \begin{tikzpicture}[remember picture,overlay]
    \begin{scope}[on background layer]
      \fill[black,opacity=0.10] ([xshift=0,yshift=2.7ex]tabsteering_judge_removal_grid_nointervention_oursbeg.north west) rectangle ([xshift=0,yshift=-1.4ex]tabsteering_judge_removal_grid_nointervention_oursend.south east);
      \fill[black,opacity=0.10] ([xshift=0,yshift=2.7ex]tabsteering_judge_removal_grid_zeroablation_oursbeg.north west) rectangle ([xshift=0,yshift=-1.4ex]tabsteering_judge_removal_grid_zeroablation_oursend.south east);
      \fill[black,opacity=0.10] ([xshift=0,yshift=2.7ex]tabsteering_judge_removal_grid_neg0p5_oursbeg.north west) rectangle ([xshift=0,yshift=-1.4ex]tabsteering_judge_removal_grid_neg0p5_oursend.south east);
      \fill[black,opacity=0.10] ([xshift=0,yshift=2.7ex]tabsteering_judge_removal_grid_pos0p5_oursbeg.north west) rectangle ([xshift=0,yshift=-1.4ex]tabsteering_judge_removal_grid_pos0p5_oursend.south east);
      \fill[black,opacity=0.10] ([xshift=0,yshift=2.7ex]tabsteering_judge_removal_grid_neg1_oursbeg.north west) rectangle ([xshift=0,yshift=-1.4ex]tabsteering_judge_removal_grid_neg1_oursend.south east);
      \fill[black,opacity=0.10]
        ([xshift=0,yshift=2.7ex]tabsteering_judge_removal_grid_pos1_oursbeg.north west)
        rectangle ([xshift=0,yshift=-1.4ex]tabsteering_judge_removal_grid_pos1_oursend.south east);
      \fill[black,opacity=0.10]
        ([xshift=0,yshift=2.7ex]tabsteering_judge_removal_grid_neg2p5_oursbeg.north west)
        rectangle ([xshift=0,yshift=-1.4ex]tabsteering_judge_removal_grid_neg2p5_oursend.south east);
      \fill[black,opacity=0.10]
        ([xshift=0,yshift=2.7ex]tabsteering_judge_removal_grid_pos2p5_oursbeg.north west)
        rectangle ([xshift=0,yshift=-1.4ex]tabsteering_judge_removal_grid_pos2p5_oursend.south east);
      \fill[black,opacity=0.10]
        ([xshift=0,yshift=2.7ex]tabsteering_judge_removal_grid_neg5_oursbeg.north west)
        rectangle ([xshift=0,yshift=-1.4ex]tabsteering_judge_removal_grid_neg5_oursend.south east);
      \fill[black,opacity=0.10]
        ([xshift=0,yshift=2.7ex]tabsteering_judge_removal_grid_pos5_oursbeg.north west)
        rectangle ([xshift=0,yshift=-1.4ex]tabsteering_judge_removal_grid_pos5_oursend.south east);
    \end{scope}
  \end{tikzpicture}
\end{table*}

\begin{figure*}[t]
  \centering
  \resizebox{0.96\textwidth}{!}{%
    \input{figures/steering_attribution_dog}%
  }
  \caption{\textbf{Additional dog-suppression example with \ours.}
For a dog scene, we use the query ``Is there a
\textcolor{myyellow!80}{dog}?'' to identify dog-supporting \ours subunits.
Suppressing these subunits removes the dog, while keeping a description of the scene and replacing the dog with a bird.}
  \label{fig:steering_attribution_cake}
\end{figure*}

\begin{figure*}[t]
  \centering
  \resizebox{0.96\textwidth}{!}{%
    \input{figures/steering_attribution_car}%
  }
  \caption{\textbf{Additional car-suppression example with \ours.}
For a rainy street scene, we use the query ``Is there a
\textcolor{myyellow!80}{car}?'' to identify car-supporting \ours subunits.
Suppressing these subunits removes the car, windshield, and intersection
description from the generated caption, while retaining a description of the
rainy scene and the Washington street sign.}
  \label{fig:steering_attribution_car}
\end{figure*}

\begin{figure*}[t]
  \centering
  \resizebox{0.96\textwidth}{!}{%
    \input{figures/steering_attribution_pizza}%
  }
    \caption{\textbf{Additional pizza-suppression example with \ours.}
    The original caption identifies the food as a \textcolor{myyellow!80}{pizza} with arugula and prosciutto. Removing the pizza-attributing \ours subunits instead yields a generic flatbread description, while preserving details of the wooden board and its green and red toppings.}
  \label{fig:steering_attribution_pizza}
\end{figure*}

\begin{figure*}[t]
  \centering
  \resizebox{0.96\textwidth}{!}{%
    \input{figures/steering_attribution_laptop}%
  }
  \caption{\textbf{Additional laptop-suppression example with \ours.}
For an image of a man seated on a couch, we use the query ``Is there a
\textcolor{myyellow!80}{laptop}?'' to identify laptop-supporting \ours
subunits. Suppressing these subunits removes the laptop and related
work-or-browsing description from the caption, while retaining the man, couch,
clothing, and living-room context.}
  \label{fig:steering_attribution_laptop}
\end{figure*}

\begin{figure*}[t]
  \centering
  \resizebox{0.96\textwidth}{!}{%
    \input{figures/steering_attribution_train}%
  }
\caption{\textbf{Additional train-suppression example with \ours.}
The original caption interprets the maroon-and-yellow vehicle as a
\textcolor{myyellow!80}{train} and describes its locomotive, passenger car,
and tracks. Once train-supporting \ours subunits are suppressed, the same
visual features are instead described as a tractor on a
gravel path, while preserving its color, orientation, and outdoor setting.}
  \label{fig:steering_attribution_train}
\end{figure*}

\subsection{PCA expansion factors}\label{sec:appendix:pca_exp_fac}

\Cref{tab:pca_table} reports PCA at expansion factors $0.25$, $0.5$, and $1$. Retaining more components improves reconstruction and downstream accuracy, reaching exact reconstruction at expansion factor $1$, but reduces C-Score and SIM@1. Thus, PCA is lossless only at full rank; lower-rank PCA can improve SIM@1 only by discarding information.

\begin{table}[t]
  \caption{
    \textbf{Monosemanticity \vsiclr~faithfulness.}
    We compare multiple PCA expansion factors across two monosemanticity metrics---SIM@1 (higher is better)
    and C-Score (higher is better)---and two faithfulness metrics---downstream
    classification accuracy and $R^2$ reconstruction (for both higher is
    better)---on DINOv2-B/14. We report mean and
    standard deviation across three seeded token probing datasets. The best
    method is shown in \textbf{bold} and the second best is \underline{underlined}.
  }
  \label{tab:pca_table}
  \centering
  \setlength{\tabcolsep}{4.0pt}
  \scriptsize
  \begin{tabularx}{\linewidth}{@{}XS[table-format=2.2(.2)]S[table-format=2.2(.1)]S[table-format=2.2(.1)]S[table-format=3.2(.2)]@{}}
    \toprule
    & \multicolumn{2}{c}{\bfseries Monosemanticity} & \multicolumn{2}{c}{\bfseries Faithfulness} \\ 
    \cmidrule(lr){2-3}\cmidrule(l){4-5}
    {\bfseries Method} & {\bfseries C-Score $(\uparrow)$} & {\bfseries SIM@1 $(\uparrow)$} & {\bfseries Accuracy $(\uparrow)$} & {\bfseries R$^2$ $(\uparrow)$} \\
    \midrule
    \emph{Backbone only} & \num{36.2} & \num{10.49}\tablestd{0.06} & \num{84.3} & \num{100.00} \\
    \cmidrule(l){1-5}
    PCA \textcolor{black!55}{\textit{(EF=0.25)}} & \textbf{\num{32.95}}\tablestd{0.60} & \num{14.49}\tablestd{0.33} & \num{81.83}\tablestd{0.04} & \num{84.17}\tablestd{0.00} \\
    PCA \textcolor{black!55}{\textit{(EF=0.5)}} & \uline{\num{27.50}}\tablestd{0.36} & \num{11.34}\tablestd{0.02} & \num{83.36}\tablestd{0.04} & \uline{\num{92.88}}\tablestd{0.00} \\
    PCA \textcolor{black!55}{\textit{(EF=1)}} & \num{23.38}\tablestd{0.21} & \num{9.10}\tablestd{0.12} & \num{84.29}\tablestd{0.00} & \textbf{\num{100.00}}\tablestd{0.00} \\
    \bottomrule
  \end{tabularx}
\end{table}

\subsection{Generalization}\label{sec:appendix:generalization_results}

To assess whether our findings extend beyond the DINOv2-B/14 setting, we consider two complementary evaluations. \Cref{tab:interpretability_vs_faithfulness2} repeats the complete comparison on the supervised ImageNet-21k ViT-B/16, including PCA and Semi-NMF, alongside the SAE and transcoder baselines. \Cref{tab:generalization} then evaluates \ours across layers, backbone sizes, model families, and expansion factors. Given the breadth of this evaluation, we compare \ours with the four strongest SAE baselines from the primary comparison across all settings. Both tables report LLM Rank, C-Score, SIM@1, downstream accuracy, and $R^2$. The analysis is in \cref{sec:experiments:interp_faith}; metric definitions are in \cref{sec:appendix:definitions}.

\begin{table}[t]
  \caption{\textbf{Complete comparison on supervised ImageNet-21k ViT-B/16.} We compare \ours (shaded) with PCA, Semi-NMF, SAE, and transcoder baselines. Metrics and formatting conventions match \cref{tab:interpretability_vs_faithfulness}; the best method per metric is shown in \textbf{bold} and the second best in \underline{underlined}.}
  \label{tab:interpretability_vs_faithfulness2}
  \centering
  \setlength{\tabcolsep}{5.0pt}
  \scriptsize
  \begin{tabularx}{\linewidth}{@{}rXS[table-format=1]S[table-format=2.2(.2)]S[table-format=2.2(.1)]S[table-format=2.2(.1)]S[table-format=3.2(.2)]@{}l@{}}
    \toprule
    &  & \multicolumn{3}{c}{\bfseries Interpretability} & \multicolumn{2}{c}{\bfseries Faithfulness}\\
    \cmidrule(lr){3-5}\cmidrule(l){6-7}
    {\bfseries Model} & {\bfseries Method} & {\bfseries LLM Rank $(\downarrow)$} & {\bfseries C-Score $(\uparrow)$} & {\bfseries SIM@1 $(\uparrow)$} & {\bfseries Accuracy $(\uparrow)$} & {\bfseries R$^2$ $(\uparrow)$}\\
    \midrule
    \parbox[t]{2mm}{\multirow{15}{*}{\rotatebox[origin=c]{90}{ImageNet-21k ViT-B/16}}} & %
    \textit{Backbone only} & \num{10} & \num{36.42} & \num{11.31}\tablestd{0.06} & \num{85.1} & \num{100.00}\\
    \cmidrule(l){2-7}
    & {\textcolor{black!55}{\textit{Linear baselines}}} & \\
    & PCA\textsuperscript{\textdagger} & 9 & \num{35.61}\tablestd{0.44} & \num{12.23}\tablestd{0.07} & \num{85.00}\tablestd{0.01} & \num{83.86}\tablestd{0.00}\\
    & Semi-NMF & \num{8} & \num{41.91}\tablestd{0.03} & \num{14.59}\tablestd{0.06} & \num{84.90}\tablestd{0.01} & \num{66.93}\tablestd{0.05}\\
    \cmidrule(l){2-7}
    & {\textcolor{black!55}{\textit{Nonlinear baselines}}} & \\
   & Vanilla SAE & \num{7} & \num{54.55}\tablestd{0.06} & \num{34.59}\tablestd{0.19} & \num{85.01}\tablestd{0.01} & \num{91.65}\tablestd{0.02}\\
   & TopK-SAE & \num{4} & \num{54.50}\tablestd{0.17} & \num{34.08}\tablestd{0.49} & \num{84.94}\tablestd{0.01} & \num{87.00}\tablestd{0.01}\\
   & JumpReLU & \num{12} & \num{31.82}\tablestd{0.08} & \num{6.05}\tablestd{0.16} & \uline{\num{85.10}}\tablestd{0.01} & \uline{\num{99.99}}\tablestd{0.00}\\
   & BatchTopK-SAE & \num{6} & \num{53.56}\tablestd{0.55} & \num{34.40}\tablestd{0.26} & \num{84.98}\tablestd{0.04} & \num{87.76}\tablestd{0.08}\\
   & Matryoshka SAE & \num{5} & \num{51.24}\tablestd{0.21} & \num{30.29}\tablestd{0.30} & \num{84.97}\tablestd{0.04} & \num{86.85}\tablestd{0.01}\\
   & Archetypal SAE & \uline{\num{2}} & \uline{\num{59.18}}\tablestd{4.03} & \bfseries \num{45.11}\tablestd{7.89} & \num{84.88}\tablestd{0.03} & \num{79.76}\tablestd{0.57}\\
   & Matching Pursuit & \num{3} & \num{56.91}\tablestd{0.12} & \num{32.81}\tablestd{0.29} & \num{85.03}\tablestd{0.05} & \num{89.66}\tablestd{0.07}\\
   & Transcoder & \num{11} & \num{46.31}\tablestd{0.03} & \num{23.00}\tablestd{0.12} & \num{85.05}\tablestd{0.03} & \num{93.54}\tablestd{0.01}\\
    \cmidrule(l){2-7}
    & \tikz[remember picture,baseline] \node[inner sep=0pt,outer sep=0pt,minimum size=0pt] (row1beg) {};\ours \textit{(ours)} & \textbf{\num{1}} & \bfseries \num{71.05}\tablestd{0.06} & \uline{\num{42.90}}\tablestd{0.15} & \bfseries \num{85.10}\tablestd{0.00} & \bfseries \num{100.00}\tablestd{0.00} & {\rlap{\tikz[remember picture,baseline] \node[inner sep=0pt,outer sep=0pt,minimum size=0pt] (row1end) {};}}\\
  \bottomrule
  \end{tabularx}
  \begin{tikzpicture}[remember picture,overlay]
    \begin{scope}[on background layer]
      \fill[black,opacity=0.10]
        ([xshift=0,yshift=2.7ex]row1beg.north west) %
        rectangle
        ([xshift=0,yshift=-1.4ex]row1end.south east);
    \end{scope}
  \end{tikzpicture}
\end{table}

\begin{table}[t]
  \caption{
    \textbf{Generalization.} We compare \ours (shaded) against the four strongest baselines from \cref{tab:interpretability_vs_faithfulness} -- Matching Pursuit, Archetypal SAE, Matryoshka SAE, and BatchTopK-SAE -- across four axes of generalization: layer, backbone size, model family, and expansion factor. Metrics and formatting conventions match \cref{tab:interpretability_vs_faithfulness}; the best method per axis and metric is shown in \textbf{bold} and the second best is \underline{underlined}.
  }
  \label{tab:generalization}
  \centering
  \setlength{\tabcolsep}{5.0pt}
  \scriptsize
  \begin{tabularx}{\linewidth}{@{}rXS[table-format=1]S[table-format=2.1]S[table-format=2.1]S[table-format=2.1]S[table-format=2.1]@{}l@{}}
    \toprule
    &  & \multicolumn{3}{c}{\bfseries Monosemanticity} & \multicolumn{2}{c}{\bfseries Faithfulness}\\
    \cmidrule(lr){3-5}\cmidrule(l){6-7}
    {\bfseries \shortstack{Axis}} & {\bfseries Method} & {\bfseries LLM Rank $(\downarrow)$} & {\bfseries C-Score $(\uparrow)$} & {\bfseries SIM@1 $(\uparrow)$} & {\bfseries Accuracy $(\uparrow)$} & {\bfseries R$^2$ $(\uparrow)$}\\
    \midrule
    \parbox[t]{2mm}{\multirow{17}{*}{\rotatebox[origin=c]{90}{Layer}}} &
    \textit{DINOv2 ViT-B/14 layer 10 (backbone only)} & \num{6} & \num{37.02} & \num{8.95} & \num{84.3} & \num{100}\\
    \cmidrule(l){2-7}
   & BatchTopK-SAE & \uline{\num{2}} & \num{51.58}\tablestd{0.49} & \num{17.98}\tablestd{0.54} & \num{83.61}\tablestd{0.04} & \num{81.66}\tablestd{0.03}\\
   & Matryoshka SAE & \num{3} & \num{50.40}\tablestd{0.27} & \num{16.40}\tablestd{0.27} & \num{83.61}\tablestd{0.06} & \num{81.12}\tablestd{0.07}\\
   & Archetypal SAE & \num{5} & \num{46.73}\tablestd{0.06} & \num{15.37}\tablestd{0.07} & \num{81.22}\tablestd{0.24} & \num{69.01}\tablestd{0.00}\\
   & Matching Pursuit & \num{4} & \uline{\num{55.77}}\tablestd{0.07} & \uline{\num{18.09}}\tablestd{0.02} & \uline{\num{83.91}}\tablestd{0.03} & \uline{\num{83.57}}\tablestd{0.07}\\
    \cmidrule(l){2-7}
    & \tikz[remember picture,baseline] \node[inner sep=0pt,outer sep=0pt,minimum size=0pt] (row1beg) {};\ours \textit{(ours)} & \bfseries \num{1} & \bfseries \num{70.00}\tablestd{0.70} & \bfseries \num{28.42}\tablestd{0.96} & \bfseries \num{84.29}\tablestd{0.00} & \bfseries \num{100.00}\tablestd{0.00} & {\rlap{\tikz[remember picture,baseline] \node[inner sep=0pt,outer sep=0pt,minimum size=0pt] (row1end) {};}}\\
    \cmidrule(l){2-7}
    & \textit{DINOv2 ViT-B/14 layer 11 (backbone only)} & \num{6} & \num{30.7} & \num{9.12} & \num{84.3} & \num{100}\\
    \cmidrule(l){2-7}
   & BatchTopK-SAE & \uline{\num{2}} & \num{43.92}\tablestd{0.30} & \num{16.13}\tablestd{0.31} & \num{83.54}\tablestd{0.07} & \num{83.22}\tablestd{0.06}\\
   & Matryoshka SAE & \num{3} & \num{46.55}\tablestd{0.43} & \num{17.42}\tablestd{0.27} & \num{83.61}\tablestd{0.06} & \num{83.84}\tablestd{0.03}\\
   & Archetypal SAE & \num{4} & \num{41.51}\tablestd{0.35} & \num{15.77}\tablestd{0.15} & \num{81.77}\tablestd{0.05} & \num{71.21}\tablestd{0.13}\\
   & Matching Pursuit & \num{5} & \uline{\num{53.91}}\tablestd{0.16} & \uline{\num{19.14}}\tablestd{0.05} & \uline{\num{83.89}}\tablestd{0.08} & \uline{\num{86.46}}\tablestd{0.06}\\
    \cmidrule(l){2-7}
    & \tikz[remember picture,baseline] \node[inner sep=0pt,outer sep=0pt,minimum size=0pt] (row2beg) {};\ours \textit{(ours)} & \bfseries \num{1} & \bfseries \num{63.53}\tablestd{0.22} & \bfseries \num{24.85}\tablestd{0.21} & \bfseries \num{84.29}\tablestd{0.00} & \bfseries \num{100.00}\tablestd{0.00} & {\rlap{\tikz[remember picture,baseline] \node[inner sep=0pt,outer sep=0pt,minimum size=0pt] (row2end) {};}}\\

    \midrule

    \parbox[t]{2mm}{\multirow{15}{*}{\rotatebox[origin=c]{90}{Size}}} &
    \textit{DINOv2 ViT-S/14 (backbone only)} & \num{6} & \num{48.33} & \num{24.51} & \num{81.16} & \num{100} \\
    \cmidrule(l){2-7}
   & BatchTopK-SAE & \num{3} & \num{68.08}\tablestd{0.20} & \num{30.21}\tablestd{0.37} & \num{78.69}\tablestd{0.14} & \num{87.24}\tablestd{0.04}\\
   & Matryoshka SAE & \uline{\num{2}} & \num{70.22}\tablestd{0.06} & \uline{\num{32.76}}\tablestd{0.18} & \num{78.80}\tablestd{0.19} & \num{87.64}\tablestd{0.03}\\
   & Archetypal SAE & \num{5} & \num{58.62}\tablestd{0.47} & \num{23.13}\tablestd{1.32} & \num{77.38}\tablestd{0.85} & \num{81.43}\tablestd{0.07}\\
   & Matching Pursuit & \num{4} & \uline{\num{70.65}}\tablestd{0.02} & \num{26.32}\tablestd{0.54} & \uline{\num{79.34}}\tablestd{0.24} & \uline{\num{90.28}}\tablestd{0.10}\\
    \cmidrule(l){2-7}
    & \tikz[remember picture,baseline] \node[inner sep=0pt,outer sep=0pt,minimum size=0pt] (row3beg) {};\ours \textit{(ours)} & \bfseries \num{1} & \bfseries \num{73.31}\tablestd{0.06} & \bfseries \num{39.70}\tablestd{0.35} & \bfseries \num{81.15}\tablestd{0.00} & \bfseries \num{100.00}\tablestd{0.00} & {\rlap{\tikz[remember picture,baseline] \node[inner sep=0pt,outer sep=0pt,minimum size=0pt] (row3end) {};}}\\
    \cmidrule(l){2-7}
    
    & \textit{DINOv2 ViT-L/14 (backbone only)} & \num{6} & \num{30.06} & \num{10.10} & \num{86.12} & \num{100}\\
    \cmidrule(l){2-7}
   & BatchTopK-SAE & \num{4} & \num{56.97}\tablestd{0.24} & \num{31.90}\tablestd{0.19} & \num{84.48}\tablestd{0.08} & \num{81.28}\tablestd{0.15}\\
   & Matryoshka SAE & \bfseries{\num{1}} & \num{59.03}\tablestd{0.06} & \uline{\num{33.47}}\tablestd{0.07} & \num{84.48}\tablestd{0.09} & \num{81.15}\tablestd{0.07}\\
   & Archetypal SAE & \uline{\num{2}} & \num{60.02}\tablestd{1.32} & \bfseries \num{37.23}\tablestd{1.08} & \num{83.81}\tablestd{0.10} & \num{74.35}\tablestd{0.38}\\
   & Matching Pursuit & \num{5} & \uline{\num{61.41}}\tablestd{0.18} & \num{32.15}\tablestd{0.24} & \uline{\num{84.95}}\tablestd{0.06} & \uline{\num{83.10}}\tablestd{0.11}\\
    \cmidrule(l){2-7}
    & \tikz[remember picture,baseline] \node[inner sep=0pt,outer sep=0pt,minimum size=0pt] (row4beg) {};\ours \textit{(ours)} & \num{3} & \bfseries \num{65.86}\tablestd{0.79} & \num{30.24}\tablestd{1.09} & \bfseries \num{86.12}\tablestd{0.00} & \bfseries \num{100.00}\tablestd{0.00} & {\rlap{\tikz[remember picture,baseline] \node[inner sep=0pt,outer sep=0pt,minimum size=0pt] (row4end) {};}}\\

    \midrule

    \parbox[t]{2mm}{\multirow{15}{*}{\rotatebox[origin=c]{90}{Family}}} &
    \textit{SigLIP2-B/16 (backbone only)} & \num{6} & \num{34.97} & \num{12.43} & \num{78.08} & \num{100.00} \\
    \cmidrule(l){2-7}
      & BatchTopK-SAE & \num{3} & \num{44.83}\tablestd{0.31} & \num{14.64}\tablestd{0.17} & \num{77.69}\tablestd{0.01} & \num{93.69}\tablestd{0.12}\\
      & Matryoshka SAE & \num{4} &  \num{44.22}\tablestd{0.32} & \num{13.44}\tablestd{0.34} & \num{77.73}\tablestd{0.04} & \num{93.82}\tablestd{0.05}\\
      & Archetypal SAE & \num{5} & \num{36.84}\tablestd{0.51} & \num{9.24}\tablestd{0.40} & \num{74.48}\tablestd{0.18} & \num{77.54}\tablestd{0.17}\\
      & Matching Pursuit & \bfseries \num{1} & \uline{\num{55.70}}\tablestd{0.22} & \uline{\num{22.81}}\tablestd{0.42} & \uline{\num{77.94}}\tablestd{0.05} & \uline{\num{95.62}}\tablestd{0.02}\\
    \cmidrule(l){2-7}
    & \tikz[remember picture,baseline] \node[inner sep=0pt,outer sep=0pt,minimum size=0pt] (row5beg) {};\ours \textit{(ours)} & \uline{\num{2}} & \bfseries \num{67.98}\tablestd{3.20} & \bfseries \num{24.04}\tablestd{1.07} & \bfseries \num{78.08}\tablestd{0.00} & \bfseries \num{100.00}\tablestd{0.00} & {\rlap{\tikz[remember picture,baseline] \node[inner sep=0pt,outer sep=0pt,minimum size=0pt] (row5end) {};}}\\
    \cmidrule(l){2-7}
    & \textit{ConvNeXtV2-Base (backbone only)} & \num{6} & \num{9.94} & \num{18.2} & \num{86.74} & \num{100}\\
    \cmidrule(l){2-7}
    & BatchTopK-SAE & \bfseries \num{1} & \num{25.65}\tablestd{0.12} & \uline{\num{29.81}}\tablestd{0.19} & \uline{\num{86.70}}\tablestd{0.01} & \uline{\num{84.77}}\tablestd{0.06}\\
   & Matryoshka SAE & \num{3} & \num{23.44}\tablestd{0.07} & \num{27.46}\tablestd{0.08} & \num{86.67}\tablestd{0.01} & \num{84.42}\tablestd{0.01}\\
   & Archetypal SAE & \num{5} & \num{17.35}\tablestd{0.32} & \num{19.13}\tablestd{0.59} & \num{86.66}\tablestd{0.02} & \num{73.85}\tablestd{0.06}\\
   & Matching Pursuit & \uline{\num{2}} & \uline{\num{30.43}}\tablestd{0.08} & \bfseries \num{32.97}\tablestd{0.08} & \num{86.69}\tablestd{0.01} & \num{84.03}\tablestd{0.01}\\
   \cmidrule(l){2-7}
    & \tikz[remember picture,baseline] \node[inner sep=0pt,outer sep=0pt,minimum size=0pt] (row6beg) {};\ours \textit{(ours)} & \num{4} & \bfseries \num{32.66}\tablestd{0.09} & \num{18.53}\tablestd{0.62} & \bfseries \num{86.74}\tablestd{0.00} & \bfseries \num{100.00}\tablestd{0.00} & {\rlap{\tikz[remember picture,baseline] \node[inner sep=0pt,outer sep=0pt,minimum size=0pt] (row6end) {};}}\\

    \midrule

    \parbox[t]{2mm}{\multirow{18}{*}{\rotatebox[origin=c]{90}{Expansion Factor}}} &
    \textit{DINOv2 ViT-B/14 layer 12 (backbone only) (EF4)} & \num{6} & \num{36.2} & \num{10.49} & \num{84.3} & \num{100.00}\\
    \cmidrule(l){2-7}
   & BatchTopK-SAE & \num{5} & \num{53.67}\tablestd{0.43} & \num{32.42}\tablestd{0.57} & \num{81.93}\tablestd{0.02} & \num{87.96}\tablestd{0.07}\\
   & Matryoshka SAE & \bfseries \num{1} & \num{56.36}\tablestd{0.60} & \uline{\num{33.65}}\tablestd{0.68} & \num{81.97}\tablestd{0.03} & \num{87.58}\tablestd{0.04}\\
   & Archetypal SAE & \num{3} & \uline{\num{61.98}}\tablestd{1.47} & \bfseries \num{40.40}\tablestd{1.20} & \num{81.03}\tablestd{0.08} & \num{83.65}\tablestd{0.10}\\
   & Matching Pursuit & \uline{\num{2}} & \num{58.40}\tablestd{0.12} & \num{32.35}\tablestd{0.13} & \uline{\num{82.51}}\tablestd{0.05} & \uline{\num{88.77}}\tablestd{0.01}\\
    \cmidrule(l){2-7}
    & \tikz[remember picture,baseline] \node[inner sep=0pt,outer sep=0pt,minimum size=0pt] (row7beg) {};\ours \textit{(ours)} & \num{4} & \bfseries \num{62.72}\tablestd{1.58} & \num{28.64}\tablestd{0.52} & \bfseries \num{84.29}\tablestd{0.00} & \bfseries \num{100.00}\tablestd{0.00} & {\rlap{\tikz[remember picture,baseline] \node[inner sep=0pt,outer sep=0pt,minimum size=0pt] (row7end) {};}}\\
    \cmidrule(l){2-7}
    & \textit{DINOv2 ViT-B/14 layer 12 (backbone only) (EF16)} & \num{6} & \num{36.2} & \num{10.49} & \num{84.3} & \num{100.00}\\
    \cmidrule(l){2-7}
   & BatchTopK-SAE & \num{5} & \num{44.13}\tablestd{0.91} & \num{23.60}\tablestd{0.90} & \num{82.16}\tablestd{0.05} & \num{89.65}\tablestd{0.06}\\
   & Matryoshka SAE & \num{4} & \num{47.48}\tablestd{0.03} & \num{25.86}\tablestd{0.13} & \num{82.15}\tablestd{0.05} & \num{89.38}\tablestd{0.01}\\
   & Archetypal SAE & \uline{\num{2}} & \num{43.03}\tablestd{1.33} & \num{26.09}\tablestd{1.51} & \num{81.32}\tablestd{0.11} & \num{85.26}\tablestd{0.07}\\
   & Matching Pursuit & \num{3} & \uline{\num{55.35}}\tablestd{0.36} & \uline{\num{30.19}}\tablestd{0.41} & \uline{\num{83.04}}\tablestd{0.02} & \uline{\num{91.37}}\tablestd{0.02}\\
    \cmidrule(l){2-7}
    & \tikz[remember picture,baseline] \node[inner sep=0pt,outer sep=0pt,minimum size=0pt] (row8beg) {};\ours \textit{(ours)} & \bfseries \num{1} & \bfseries \num{66.77}\tablestd{0.95} & \bfseries \num{36.21}\tablestd{0.65} & \bfseries \num{84.29}\tablestd{0.00} & \bfseries \num{100.00}\tablestd{0.00} & {\rlap{\tikz[remember picture,baseline] \node[inner sep=0pt,outer sep=0pt,minimum size=0pt] (row8end) {};}}\\

  \bottomrule
  \end{tabularx}
  \begin{tikzpicture}[remember picture,overlay]
    \begin{scope}[on background layer]
      \fill[black,opacity=0.10]
        ([xshift=0,yshift=2.7ex]row1beg.north west)
        rectangle ([xshift=0,yshift=-1.4ex]row1end.south east);
    \end{scope}
    \begin{scope}[on background layer]
      \fill[black,opacity=0.10]
        ([xshift=0,yshift=2.7ex]row2beg.north west)
        rectangle ([xshift=0,yshift=-1.4ex]row2end.south east);
    \end{scope}
    \begin{scope}[on background layer]
      \fill[black,opacity=0.10]
        ([xshift=0,yshift=2.7ex]row3beg.north west)
        rectangle ([xshift=0,yshift=-1.4ex]row3end.south east);
    \end{scope}
    \begin{scope}[on background layer]
      \fill[black,opacity=0.10]
        ([xshift=0,yshift=2.7ex]row4beg.north west)
        rectangle ([xshift=0,yshift=-1.4ex]row4end.south east);
    \end{scope}
    \begin{scope}[on background layer]
      \fill[black,opacity=0.10]
        ([xshift=0,yshift=2.7ex]row5beg.north west)
        rectangle ([xshift=0,yshift=-1.4ex]row5end.south east);
    \end{scope}
    \begin{scope}[on background layer]
      \fill[black,opacity=0.10]
        ([xshift=0,yshift=2.7ex]row6beg.north west)
        rectangle ([xshift=0,yshift=-1.4ex]row6end.south east);
    \end{scope}
    \begin{scope}[on background layer]
      \fill[black,opacity=0.10]
        ([xshift=0,yshift=2.7ex]row7beg.north west)
        rectangle ([xshift=0,yshift=-1.4ex]row7end.south east);
    \end{scope}
    \begin{scope}[on background layer]
      \fill[black,opacity=0.10]
        ([xshift=0,yshift=2.7ex]row8beg.north west)
        rectangle ([xshift=0,yshift=-1.4ex]row8end.south east);
    \end{scope}
  \end{tikzpicture}
\end{table}

\subsection{Ablations}\label{sec:appendix:ablations_results}

We evaluate two components of \ours's construction on DINOv2-B/14, layer~$12$. \ours obtains subunits by clustering each neuron's contribution patterns and restructuring its weights. The following ablations isolate the roles of feature construction and the clustering algorithm. The fixed-width alternatives use expansion factor $8$; HDBSCAN determines the number of subunits from the data. We use expansion factor $8$ to align them with the configuration in \cref{tab:interpretability_vs_faithfulness}.

\paragraph{Feature construction.} We compare \ours with three alternatives. The \textit{Random} control partitions each neuron's weights into random groups without using its contribution patterns. This tests whether an increase in the number of units alone accounts for the improvement. The other two alternatives cluster raw activations with either HDBSCAN or $k$-means. They use the resulting centroids as feature directions, compute feature activations by dot product, and reconstruct the original representation with a pseudoinverse decoder. Thus, these alternatives change how features are constructed, rather than only changing the clustering algorithm.

\paragraph{Clustering algorithm.} To isolate this choice, we retain \ours's contribution patterns and weight-restructuring procedure and replace HDBSCAN with $k$-means. The \ours row represents the HDBSCAN configuration from \cref{tab:interpretability_vs_faithfulness}.

\cref{tab:cluster_ablation} shows our ablation results. Random splitting performs substantially worse across all three metrics, indicating that gains do not solely result from expansion.  The dot-product variants achieve higher C-Scores, which is expected because their directions are constructed from centroids in the same backbone representation space used by C-Score. However, this advantage does not transfer to the complementary metrics. \ours ranks first under VLM judgments and substantially outperforms both dot-product variants on SIM@1. Moreover, unlike the dot-product variants with pseudoinverse decoding, \ours preserves the original representation exactly by construction. Overall, \ours provides the strongest balance between semantic coherence, concept-to-activation predictability, and faithfulness. Within the attribution-based construction, HDBSCAN improves LLM Rank and C-Score over $k$-means, while $k$-means performs slightly better on SIM@1.

\begin{table}[t]
  \caption{\textbf{Clustering and feature construction ablations at expansion factor $8$.}
  C-Score on DINOv2-B/14, layer~$12$. The first block compares random splitting with dot-product representations obtained by clustering raw activations. The second block varies the clustering algorithm for
  attribution patterns; \ours uses HDBSCAN.}
  \label{tab:cluster_ablation}
  \centering
  \setlength{\tabcolsep}{5.0pt}
  \scriptsize
  \begin{tabularx}{\linewidth}{
      @{}rXS[table-format=2.2]S[table-format=2.2(1.4)]S[table-format=2.2]
      @{}l@{}
  }
    \toprule
    & & \multicolumn{3}{c}{\bfseries Monosemanticity} \\
    \cmidrule(lr){3-5}
    {\bfseries Model} & {\bfseries Ablation axis}
      &  {\bfseries LLM Rank\,$(\downarrow)$} & {\bfseries C-Score\,$(\uparrow)$} & {\bfseries SIM@1\,$(\uparrow)$} \\
    \midrule
    \parbox[t]{2mm}{\multirow{8}{*}{\rotatebox[origin=c]{90}{DINOv2 ViT-B/14}}}
    & {\textcolor{black!55}{\textit{Feature construction}}} & \\
    & Random
     & \num{4} & \num{35.71}\tablestd{0.08} & \num{13.42}\tablestd{0.03} \\
    & HDBSCAN + dot product
      & \num{3} & \bfseries{\num{80.08}\tablestd{0.28}} & \num{25.68}\tablestd{0.13} \\
    & $k$-Means + dot product
      & \num{5} & \uline{\num{78.57}}\tablestd{0.13} & \num{27.30}\tablestd{0.11} \\
    \cmidrule(l){2-5}
    & {\textcolor{black!55}{\textit{Clustering algorithm}}} & \\
    & $k$-Means
      & \uline{\num{2}} & \num{52.67}\tablestd{0.07} & \bfseries{\num{35.56}\tablestd{0.36}}\\
    \cmidrule(l){2-5}
    & \tikz[remember picture,baseline]
      \node[inner sep=0pt,outer sep=0pt,minimum size=0pt]
      (crablrowbeg) {};\ours \textit{(HDBSCAN)}
      &  \bfseries{\num{1} }& \num{64.85}\tablestd{0.52} & \uline{\num{34.09}}\tablestd{0.85}
      & {\rlap{\tikz[remember picture,baseline]
        \node[inner sep=0pt,outer sep=0pt,minimum size=0pt]
        (crablrowend) {};}}\\
    \bottomrule
  \end{tabularx}
  \begin{tikzpicture}[remember picture,overlay]
    \begin{scope}[on background layer]
      \fill[black,opacity=0.10]
        ([xshift=0,yshift=2.7ex]crablrowbeg.north west)
        rectangle ([xshift=0,yshift=-1.4ex]crablrowend.south east);
    \end{scope}
  \end{tikzpicture}
\end{table}

\paragraph{HDBSCAN hyperparameter ablation} We ablate \ours's clustering choices, one hyperparameter at a time, around the default configuration (HDBSCAN \texttt{cluster\_selection\_method}, \texttt{min\_cluster\_size}, \texttt{min\_samples}, and UMAP pre-reduction), evaluated by C-Score on DINOv2-B/14 layer~$12$. Because $R^2$ and downstream accuracy are exact by construction regardless of clustering choices. We omit the LLM Rank due to limited computational budget. A random neuron disentanglement baseline confirms that the gains are
driven by the structure in the attribution vectors, not the expansion factor alone. Results in \cref{tab:appendix:ablations} show \ours is robust across all axes.
\Cref{tab:appendix:ablations} reports C-Score on DINOv2-B/14 for the ablation axes swept around the \ours baseline (HDBSCAN cluster-selection method, \texttt{min\_cluster\_size}, \texttt{min\_samples}, UMAP dimensionality reduction) at expansion factor $4$. We run the broader hyperparameter sweep at an effective expansion factor of approximately 4 to keep the cost of evaluating multiple clustering settings manageable. The baseline clustering hyperparameters are listed in \cref{sec:appendix:clustering_hp}.

\begin{table}[t]
  \caption{
    \textbf{Ablations at expansion factor 4.} C-Score on DINOv2-B/14 as we vary one \ours clustering hyperparameter at a time around the baseline (\texttt{cluster\_selection\_method}~$=$~\texttt{leaf}, \texttt{min\_cluster\_size}~$=$~$1000$, \texttt{min\_samples} at the library default, ARV unit-normalization enabled, no UMAP). Random Baseline partitions the original channels into random groups as a lower-bound control. Best per axis in \textbf{bold}.
  }
  \label{tab:appendix:ablations}
  \centering
  \setlength{\tabcolsep}{5.0pt}
  \scriptsize
  \begin{tabularx}{\linewidth}{@{}rXS[table-format=2.2(1.4)]S[table-format=2.2(1.4)]@{}l@{}}
    \toprule
    & & \multicolumn{2}{c}{\bfseries Monosemanticity} \\
    \cmidrule(lr){3-4}
    {\bfseries Model} & {\bfseries Ablation axis}
      & {\bfseries C-Score\,$(\uparrow)$} & {\bfseries SIM@1\,$(\uparrow)$} \\
    \midrule
    \parbox[t]{2mm}{\multirow{12}{*}{\rotatebox[origin=c]{90}{DINOv2 ViT-B/14}}}
    & Random Baseline                   & \num{34.52}\tablestd{0.11} & \num{12.01}\tablestd{0.08} \\
    \cmidrule(l){2-4}
    & EOM                               & \num{60.13}\tablestd{1.48} & \num{29.70}\tablestd{0.32}\\
    \cmidrule(l){2-4}
    & Min Cluster Size = 100            & \num{58.01}\tablestd{0.40} & \num{35.62}\tablestd{0.17}\\
    & Min Cluster Size = 250            & \num{58.09}\tablestd{0.26} & \num{34.49}\tablestd{1.00}\\
    & Min Cluster Size = 500            & \num{61.09}\tablestd{3.57} & \num{30.04}\tablestd{0.44}\\
    \cmidrule(l){2-4}
    & Min Samples = 1                   & \num{55.10}\tablestd{0.48}  &\num{35.26}\tablestd{0.36}\\
    & Min Samples = 100                 & \num{58.38}\tablestd{2.72} & \num{31.64}\tablestd{0.47}\\
    & Min Samples = 500                 & \num{61.90}\tablestd{2.62} & \num{28.86}\tablestd{0.23}\\
    \cmidrule(l){2-4}
    & UMAP                              & \num{54.24}\tablestd{0.57} & \bfseries{\num{39.18}\tablestd{0.60}}\\
    \cmidrule(l){2-4}
    & \tikz[remember picture,baseline] \node[inner sep=0pt,outer sep=0pt,minimum size=0pt] (ablrowbeg) {};\ours \textit{(ours)} & \bfseries \num{62.72}\tablestd{1.58} & \num{28.64}\tablestd{0.52} &{\rlap{\tikz[remember picture,baseline] \node[inner sep=0pt,outer sep=0pt,minimum size=0pt] (ablrowend) {};}}\\
    \bottomrule
  \end{tabularx}
  \begin{tikzpicture}[remember picture,overlay]
    \begin{scope}[on background layer]
      \fill[black,opacity=0.10]
        ([xshift=0,yshift=2.7ex]ablrowbeg.north west)
        rectangle ([xshift=0,yshift=-1.4ex]ablrowend.south east);
    \end{scope}
  \end{tikzpicture}
\end{table}

\paragraph{$\rho$ selection and ablation.}\label{sec:appendix:rho}

For each split neuron, we select $\rho$ independently from $\{0.05, 0.10, \dots, 1.00\}$. For every candidate value, we construct the corresponding subunits and measure, for each cluster, how much more strongly its assigned subunit activates than the other subunits on examples from that cluster. We select the value that maximizes this average separation across the neuron's clusters (see \cref{eq:winner_margin}). The resulting selection distribution is reported in \cref{tab:rho_distribution}.

To assess the effect of this selection, we repeat the DINOv2-B/14, configuration from \cref{tab:appendix:ablations} while fixing one value of $\rho$ for all neurons. In this configuration, clustering divides 673 original neurons into multiple subunits; neurons for which it finds no split remain unchanged. Because, clustering is independent of $\rho$, all runs split the same $673$ neurons; only the assignment of the original weights to subunits changes. Faithfulness remains exact in every setting, since the subunit weights always sum to the original weight.

The selected margins span the full candidate grid, confirming that no single value is preferred across all neurons. When $\rho$ is fixed globally, C-Score remains relatively stable from $\rho=0.05$ to $\rho=0.95$, ranging from $61.86$ to $64.51$, but drops sharply to $50.07$ at $\rho=1.00$. Although $\rho=1.00$ is selected for some individual neurons, applying it globally makes every split maximally exclusive and is detrimental overall. The best fixed value, $\rho=0.20$, can only be identified in hindsight. More generally, selecting an optimal per-neuron assignment through validation would require considering $20 \times {673}$ possible combinations in this setting, while a global value would need to be retuned for each model and configuration. The automatic rule instead selects $\rho$ locally for each neuron without model-level tuning.

\begin{table}[t]
  \caption{\textbf{Distribution of automatically selected margins $\rho$.}
  Number of the $673$ split neurons selecting each candidate value on
  DINOv2-B/14, layer~12, with top-$k=60{,}000$.
  Margins are selected independently per neuron as defined in
  \cref{eq:winner_margin}.}
  \label{tab:rho_distribution}
  \centering
  \scriptsize
  \setlength{\tabcolsep}{2.5pt}
  \begin{tabular}{@{}l*{10}{c}@{}}
    \toprule
    $\rho$ & 0.05 & 0.10 & 0.15 & 0.20 & 0.25
            & 0.30 & 0.35 & 0.40 & 0.45 & 0.50 \\
    Neurons & 11 & 12 & 30 & 41 & 34 & 20 & 26 & 20 & 27 & 26 \\
    \midrule
    $\rho$ & 0.55 & 0.60 & 0.65 & 0.70 & 0.75
            & 0.80 & 0.85 & 0.90 & 0.95 & 1.00 \\
    Neurons & 30 & 48 & 64 & 48 & 57 & 31 & 25 & 16 & 19 & 88 \\
    \bottomrule
  \end{tabular}
\end{table}

\begin{table}[t]
  \caption{\textbf{Ablation of the margin $\rho$.}
  C-Score on DINOv2-B/14, layer~12, when one global $\rho$ is fixed
  for all neurons. Fixed-margin results use the seed-$42$ token
  dataset and top-$k=60{,}000$; faithfulness is exact by construction
  in every setting ($R^2=100\%$). The final row gives the automatic
  per-neuron result across the three seeded token datasets for reference.}
  \label{tab:rho_ablation}
  \centering
  \scriptsize
  \setlength{\tabcolsep}{5.0pt}
  \begin{tabular}{@{}lS[table-format=3.2(2.2)]@{}l@{}}
    \toprule
    {\bfseries $\rho$} & {\bfseries C-Score $(\uparrow)$} & {\bfseries SIM@1 $(\uparrow)$} \\
    \midrule
    0.05 & \num{64.03} & \num{27.86} \\
    0.10 & \num{64.28} & \num{27.84} \\
    0.15 & \num{64.28} & \num{27.91} \\
    \bfseries{\num{0.20}} & \bfseries{\num{64.51}} & \num{27.77} \\
    0.25 & \num{64.06} & \num{27.53} \\
    0.30 & \num{63.83} & \num{27.40}  \\
    0.35 & \num{63.61} & \num{27.28} \\
    0.40 & \num{63.54} & \num{27.04} \\
    0.45 & \num{63.32} & \num{27.04} \\
    0.50 & \num{62.92} & \num{27.05} \\
    0.55 & \num{63.07} & \num{26.82} \\
    0.60 & \num{62.85} & \num{26.70} \\
    0.65 & \num{62.53} & \num{26.74} \\
    0.70 & \num{62.31} & \num{26.65} \\
    0.75 & \num{62.42} & \num{26.50} \\
    0.80 & \num{62.31} & \num{26.40} \\
    0.85 & \num{62.16} & \num{26.19} \\
    0.90 & \num{61.99} & \num{26.18} \\
    0.95 & \num{61.86} & \num{26.14} \\
    1.00 & \num{50.07} & \num{26.71} \\
    \midrule
    \tikz[remember picture,baseline]
      \node[inner sep=0pt,outer sep=0pt,minimum size=0pt] (rhoautobeg) {};
      \textit{Automatic per-neuron selection}
      &\num{62.72}\tablestd{1.58} & \num{28.64}\tablestd{0.52}
       {\rlap{\tikz[remember picture,baseline]
          \node[inner sep=0pt,outer sep=0pt,minimum size=0pt] (rhoautoend) {};}}\\
    \bottomrule
  \end{tabular}
  \begin{tikzpicture}[remember picture,overlay]
    \begin{scope}[on background layer]
      \fill[black,opacity=0.10]
        ([xshift=0,yshift=2.7ex]rhoautobeg.north west)
        rectangle ([xshift=0,yshift=-1.4ex]rhoautoend.south east);
    \end{scope}
  \end{tikzpicture}
\end{table}

\subsection{Additional qualitative examples}\label{sec:appendix:qualitative}

\Cref{fig:qualitative_examples} shows representative \ours disentanglements on DINOv2-B/14 (a $2$-way and a $4$-way split). The number of subunits is not a fixed hyperparameter -- HDBSCAN discovers it automatically per unit (\cf~\cref{sec:method:2}) and varies neuron-by-neuron depending on how many distinct concepts drive the unit's activation. Some units produce a single cluster (effectively monosemantic already and not split further); others split into many. To illustrate the range we observe in practice, we show one representative example for each of $K \in \{1, 3, 5, 6, 7, 8\}$ subunits in \cref{fig:appendix:collage_nosplit,fig:appendix:collage_3split,fig:appendix:collage_5split,fig:appendix:collage_6split,fig:appendix:collage_7split,fig:appendix:collage_8split}. The format mirrors that of \cref{fig:qualitative_examples}: the left panel shows the top-$16$ activating images of the parent unit, and each pink panel on the right shows the top-$4$ activating images for one \ours subunit. Concept labels above each subunit panel are short human-assigned descriptions of the visible theme. All examples use the same configuration as \cref{sec:experiments:interp_faith}: DINOv2-B/14, layer~$12$, TopK~$=400$k, seed~$42$.

\begin{figure}[t]
  \centering
  \resizebox{\linewidth}{!}{\input{figures/neuron_collage_nosplit.tex}}
  \caption{\textbf{$K=1$: an already-monosemantic unit.} HDBSCAN finds only a single cluster in the unit's contribution-vector set, so \ours leaves the unit unchanged. }
  \label{fig:appendix:collage_nosplit}
\end{figure}

\begin{figure}[t]
  \centering
  \resizebox{\linewidth}{!}{\input{figures/neuron_collage_3split.tex}}
  \caption{\textbf{3-way \ours disentanglements.} A polysemantic DINOv2-B/14 unit (yellow frame, left; top-$16$) decomposed by \ours into $3$ subunits (pink frames, right; top-$4$ each). }
  \label{fig:appendix:collage_3split}
\end{figure}

\begin{figure}[t]
  \centering
  \resizebox{\linewidth}{!}{\input{figures/neuron_collage_5split.tex}}
  \caption{\textbf{5-way \ours disentanglements.} A polysemantic DINOv2-B/14 unit (yellow frame, left; top-$16$) decomposed by \ours into $5$ subunits (pink frames, right; top-$4$ each). }
  \label{fig:appendix:collage_5split}
\end{figure}

\begin{figure}[t]
  \centering
  \resizebox{\linewidth}{!}{\input{figures/neuron_collage_6split.tex}}
  \caption{\textbf{6-way \ours disentanglements.} A polysemantic DINOv2-B/14 unit (yellow frame, left; top-$16$) decomposed by \ours into $6$ subunits (pink frames, right; top-$4$ each). }
  \label{fig:appendix:collage_6split}
\end{figure}

\begin{figure}[t]
  \centering
  \resizebox{\linewidth}{!}{\input{figures/neuron_collage_7split.tex}}
  \caption{\textbf{7-way \ours disentanglements.} A polysemantic DINOv2-B/14 unit (yellow frame, left; top-$16$) decomposed by \ours into $7$ subunits (pink frames, right; top-$4$ each). }
  \label{fig:appendix:collage_7split}
\end{figure}

\begin{figure}[t]
  \centering
  \resizebox{\linewidth}{!}{\input{figures/neuron_collage.tex}}
  \caption{\textbf{8-way \ours disentanglements.} A polysemantic DINOv2-B/14 unit (yellow frame, left; top-$16$) decomposed by \ours into $8$ subunits (pink frames, right; top-$4$ each). }
  \label{fig:appendix:collage_8split}
\end{figure}

\section{Applying \ours to Language models}\label{sec:appendix:language_models}

We assess whether \ours transfers from vision to language models without changing its construction. We apply it to MLP down-projection output neurons in two instruction-tuned models, Gemma 3 270M~\citep{Gemma:2025:GTR} and Qwen3 0.6B~\citep{Qwen:2025:QTR}, similar to \citet{Lieberum:2024:GSO, McDougall:2025:GST}. For each neuron, we collect activating text examples, cluster the corresponding contribution patterns, and split its incoming weights into subunits whose sum preserves the original neuron. We limit \ours runs to approximately two billion tokens per model to keep its computational cost manageable.

\subsection{Experimental setup.}
We study layer~16 of Gemma 3 270M-IT~\citep{Gemma:2025:GTR}, then repeat the evaluation on Gemma layer~13 and Qwen3 0.6B layer~21~\citep{Qwen:2025:QTR}. We compare \ours with the backbone-only reference and four self-trained methods: Archetypal SAE (ASAE)~\citep{Fel:2025:ASA}, BatchTopK-SAE~\citep{Bussmann:2024:BTK}, Matching Pursuit SAE (MpSAE)~\citep{Costa:2025:FFH}, and Matryoshka SAE (MSAE)~\citep{Bussmann:2025:LML}. We chose these four because they were the strongest SAE baselines in our main vision experiments. For Gemma, we also evaluate published GemmaScope2 SAEs and transcoders~\citep{McDougall:2025:GST}. To compare methods with similar feature capacity, we use GemmaScope2 width-16k checkpoints at three sparsity levels (expansion factor 25.6), self-trained SAE checkpoints at expansion factor 25, and the \ours checkpoint closest to that factor for each layer.

We assess monosemanticity using four complementary evaluations. AutoInterp~\citep{Paulo:2025:AIM} generates an explanation of a feature from activating examples and tests whether that explanation predicts activity on new examples. Simulation~\citep{Bills:2023:LMC} tests whether an explanation predicts \emph{how strongly} the feature activates across a new passage. Intruder~\citep{Paulo:2026:ESI} asks a language-model judge to identify a non-matching passage among four activating passages. Embedding~\citep{Paulo:2026:ESI} tests whether activating and comparison passages can be distinguished using an independent text encoder. Intruder and embedding complement explanation-based scores because they do not depend on a generated explanation or on the choice of the language model that generates it. We separately report reconstruction $R^2$ and next-token KL divergence to assess how faithfully each representation preserves the original model's computation. The KL divergence compares the model's predicted distribution over the next token with that of the unchanged backbone on the same text; lower values mean that the predictions are more similar, and zero means they are identical.

We use Gemma 3 12B Instruct~\citep{Gemma:2025:GTR} (\texttt{google/gemma-3-12b-it}) to generate explanations and make judge decisions for AutoInterp and Intruder. Simulation uses Llama 3.1 8B~\citep{LLama:2024:TLH} (\texttt{meta-llama/Llama-3.1-8B}) to predict activation magnitudes. Embedding uses \texttt{all-MiniLM-L6-v2}~\citep{Reimers:2019:SBS} and does not use a language-model judge.

The monosemanticity evaluations use the same fixed two-million-token Pile-validation corpus. Each metric samples up to 1,000 eligible features. Reconstruction follows the separate OpenWebText\citep{Gokaslan:2019:OWT} setup used by SAEBench~\citep{Karvonene:2025:SAC}.

We report the original AutoInterp, Intruder, and embedding protocols alongside variants with activation thresholds of 25\%, 50\%, and 75\% of each feature's observed maximum in \Cref{fig:appendix:llm_threshold_ai,fig:appendix:llm_threshold_int,fig:appendix:llm_threshold_emb} alongsinde the reconstruction $R^2$ on the x-axis. The original AutoInterp protocol uses a 1\% threshold. For denser representations such as \ours, this low threshold can label weak responses as active even when they differ substantially from the feature's strongest responses. The threshold variants let us examine how this choice affects the scores. For Intruder and embedding, an activating 128-token block must have a peak activation above the specified cutoff. In the original evaluation, a comparison passage must have zero activation of the tested feature; in the threshold variants, its activation may be nonzero but must remain at or below the cutoff. Each metric and threshold selects its eligible features separately.

We construct only one \ours checkpoint and one checkpoint per self-trained SAE configuration because repeating checkpoint construction would exceed our computational budget. We repeat evaluation with seeds 42, 43, and 44, which resample the evaluated features and examples, and report the mean and standard deviation across these runs. These deviations reflect evaluation sampling, not variation across independently constructed checkpoints.

\subsection{Quantitative results.}
\Cref{tab:appendix:language_models} compares methods with similar numbers of features on Gemma layer~16 under the original evaluation protocols. It reports AutoInterp, Intruder, embedding, Simulation, reconstruction $R^2$, and next-token KL divergence. \Cref{tab:appendix:gemma12,tab:appendix:qwen21} repeat the comparison on Gemma layer~13 and Qwen layer~21 as generalization experiments.

On Gemma layer~16, \ours achieves $100.0\%$ reconstruction $R^2$, and its next-token KL divergence rounds to $0.00$ at the table's displayed precision. It also obtains the highest Simulation score ($8.3$). Some SAEs and transcoders outperform \ours on AutoInterp, Intruder, and Embedding under the evaluation protocols of SAEBench~\citep{Karvonene:2025:SAC} and \citet{Paulo:2026:ESI}. Together, these results show that \ours produces interpretable subunits while preserving the backbone's computation.

The scatter plots in \Cref{fig:appendix:llm_threshold_ai,fig:appendix:llm_threshold_int,fig:appendix:llm_threshold_emb} show that the definition of an active feature matters. At the $25\%$ activation threshold, \ours reaches an AutoInterp score of $85.5$, exceeding every compared method at that threshold while retaining exact reconstruction. At the $75\%$ threshold, its Intruder score reaches $38.3$, however, it still lacks behind BatchTopK SAE's $42.63$ which, however, has a lower reconstruction $R^2$ of $94.35\%$. %
\ours also improves the Embedding score compared to the original model, from $69.5$ to $77.3$, but remains below the strongest baselines. This sensitivity to activation thresholds appears to be more pronounced in the language-model evaluation than in our vision experiments, where \ours scored well on SIM@1 despite that metric considering lower-activation examples (\cref{sec:experiments:interp_faith}). Furthermore, \ours achieves the highest language-model Simulation score, which evaluates predictions of activation strength across new text \cref{tab:appendix:language_models}. Weak activations are therefore not inherently a problem for \ours; rather, the language-model scores depend more strongly on where the evaluation draws the line between active and inactive features. %

On Gemma layer~13 and Qwen layer~21, \ours again retains $100.0\%$ reconstruction $R^2$. Its monosemanticity scores vary across the two settings and metrics, suggesting that the lossless construction transfers more consistently than any particular ranking against the SAE baselines.

\begin{table*}[t]
  \caption{\textbf{Monosemanticity \vsiclr~faithfulness on Gemma 3 270M-IT, layer~16.} We compare \ours (shaded) with self-trained SAEs at expansion factor 25 and published GemmaScope2 SAEs and transcoders at width 16k (expansion factor 25.6). AI denotes AutoInterp, INT Intruder, EMB Embedding, and SIM Simulation. R$^2$ and next-token KL measure faithfulness. We report means and standard deviations across evaluation seeds 42, 43, and 44. KL is multiplied by 100 for easier reading.}
  \label{tab:appendix:language_models}
  \centering
  \setlength{\tabcolsep}{5.0pt}
  \scriptsize
  \begin{tabularx}{\textwidth}{@{}rXS[table-format=1.3]S[table-format=1.3]S[table-format=1.3]S[table-format=1.3]S[table-format=1.3]S[table-format=1.3]@{}l@{}}
    \toprule
    & & \multicolumn{4}{c}{\bfseries Monosemanticity} & \multicolumn{2}{c}{\bfseries Faithfulness}\\
    \cmidrule(lr){3-6}\cmidrule(l){7-8}
    {\bfseries Model} & {\bfseries Method} &
    {\bfseries AI $(\uparrow)$} &
    {\bfseries INT $(\uparrow)$} &
    {\bfseries EMB $(\uparrow)$} &
    {\bfseries SIM $(\uparrow)$} &
    {\bfseries R$^2$ $(\uparrow)$} &
    {\bfseries KL $(\downarrow)$}\\
    \midrule

    \parbox[t]{2mm}{\multirow{15}{*}{\rotatebox[origin=c]{90}{Gemma 3 270M IT}}}
    & \textit{Backbone only} & \num{62.1}\tablestd{0.63} & \num{20.3}\tablestd{0.05} & \num{66.9}\tablestd{0.17} & \num{1.5}\tablestd{0.18} & \num{100.0}\tablestd{0.00} & \num{0.00}\tablestd{0.00} \\
    \cmidrule(l){2-8}
    & {\textcolor{black!55}{\textit{GemmaScope2}}} & \\
    & JumpReLU~\textcolor{black!55}{(Small)} & \num{77.3}\tablestd{0.61} & \num{29.1}\tablestd{0.57} & \num{82.8}\tablestd{0.25} & \num{4.5}\tablestd{0.16} & \num{78.1}\tablestd{0.09} & \num{10.15}\tablestd{0.12} \\
    & JumpReLU~\textcolor{black!55}{(Medium)} & \num{74.4}\tablestd{0.23} & \num{25.9}\tablestd{0.27} & \num{80.2}\tablestd{0.27} & \num{3.6}\tablestd{0.41} & \num{83.4}\tablestd{0.07} & \num{10.94}\tablestd{0.06} \\
    & JumpReLU~\textcolor{black!55}{(Large)} & \num{71.7}\tablestd{0.58} & \num{23.7}\tablestd{0.15} & \num{75.6}\tablestd{0.09} & \num{3.8}\tablestd{0.37} & \num{87.7}\tablestd{0.05} & \num{8.25}\tablestd{0.04} \\
    & Transcoder~\textcolor{black!55}{(Small)} & \num{80.7}\tablestd{0.40} & \num{32.3}\tablestd{0.29} & \num{84.7}\tablestd{0.21} & \num{4.7}\tablestd{0.24} & \num{54.0}\tablestd{0.21} & \num{10.97}\tablestd{0.15} \\
    & Transcoder~\textcolor{black!55}{(Medium)} & \num{78.0}\tablestd{0.74} & \num{29.7}\tablestd{0.12} & \num{80.7}\tablestd{0.10} & \num{4.9}\tablestd{0.20} & \num{64.1}\tablestd{0.17} & \num{7.92}\tablestd{0.11} \\
    & Transcoder~\textcolor{black!55}{(Large)} & \num{75.9}\tablestd{1.07} & \num{27.0}\tablestd{0.21} & \num{76.1}\tablestd{0.02} & \num{4.3}\tablestd{0.04} & \num{67.5}\tablestd{0.14} & \num{6.04}\tablestd{0.08} \\
    \cmidrule(l){2-8}
    & {\textcolor{black!55}{\textit{Self-Trained SAEs at EF25}}} & \\
    & Matching Pursuit SAE & \num{76.4}\tablestd{0.44} & \num{23.7}\tablestd{0.40} & \num{78.7}\tablestd{0.13} & \num{6.3}\tablestd{0.27} & \num{95.9}\tablestd{0.04} & \num{7.25}\tablestd{0.11} \\
    & Archetypal SAE & \num{71.4}\tablestd{0.60} & \num{24.4}\tablestd{0.53} & \num{71.3}\tablestd{0.47} & \num{-1.1}\tablestd{0.06} & \num{94.7}\tablestd{0.04} & \num{11.28}\tablestd{0.04} \\
    & Matryoshka SAE & \num{80.2}\tablestd{0.26} & \num{28.4}\tablestd{0.09} & \num{83.0}\tablestd{0.18} & \num{6.4}\tablestd{0.22} & \num{94.4}\tablestd{0.07} & \num{9.59}\tablestd{0.09} \\
    & BatchTopK-SAE & \num{80.7}\tablestd{0.27} & \num{27.8}\tablestd{0.38} & \num{83.0}\tablestd{0.04} & \num{7.0}\tablestd{0.41} & \num{94.3}\tablestd{0.07} & \num{9.62}\tablestd{0.09} \\
    \cmidrule(l){2-8}
    & \tikz[remember picture,baseline] \node[inner sep=0pt,outer sep=0pt,minimum size=0pt] (llmgemma16beg) {};\ours \textit{(ours)} & \num{75.9}\tablestd{0.42} & \num{20.8}\tablestd{0.05} & \num{69.5}\tablestd{0.42} & \num{8.3}\tablestd{0.29} & \num{100.0}\tablestd{0.00} & \num{0.00}\tablestd{0.00} & {\rlap{\tikz[remember picture,baseline] \node[inner sep=0pt,outer sep=0pt,minimum size=0pt] (llmgemma16end) {};}} \\
    \bottomrule
  \end{tabularx}
  \begin{tikzpicture}[remember picture,overlay]
    \begin{scope}[on background layer]
      \fill[black,opacity=0.10]
        ([xshift=0,yshift=2.7ex]llmgemma16beg.north west)
        rectangle
        ([xshift=0,yshift=-1.4ex]llmgemma16end.south east);
    \end{scope}
  \end{tikzpicture}
\end{table*}

\begin{figure*}[t]
  \centering
  \begin{minipage}{0.70\textwidth}
    \begin{sansmath}
\begin{tikzpicture}
\scriptsize
\sffamily
\begin{axis}[
    width=1\linewidth,
    height=0.68\linewidth,
    xmin=0.50, xmax=1.04,
    ymin=0.50, ymax=0.95,
    xlabel={Faithfulness ($\uparrow$)},
    ylabel={AutoInterp ($\uparrow$)},
    xtick={0.5,0.6,0.7,0.8,0.9,1.0},
    ytick={0.5,0.6,0.7,0.8,0.9},
    label style={font=\scriptsize},
    axis line style={thick},
    enlargelimits=false,
    clip=false,
    axis background/.style={fill=gray!5,rounded corners=1pt},
    grid=both,
    major grid style={gray!30},
    minor grid style={gray!15},
    axis lines=left,
    trim axis left,
    trim axis right,
]
\addplot[
    only marks, scatter, scatter src=explicit symbolic,
    scatter/classes={raw={black},gs={myblue},gt={mypurple},sae={myyellow},dune={mypink}},
    mark=diamond*, mark size=2.2pt, opacity=0.85,
] table[meta=family,x=x,y=y] {
x y family
1.000000 0.620650 raw
0.780804 0.773317 gs
0.833652 0.744398 gs
0.876731 0.716952 gs
0.540204 0.806890 gt
0.640777 0.779677 gt
0.675103 0.758799 gt
0.959318 0.763714 sae
0.946808 0.713617 sae
0.943825 0.802452 sae
0.943467 0.806698 sae
0.999998 0.759500 dune
};

\addplot[
    only marks, scatter, scatter src=explicit symbolic,
    scatter/classes={raw={black},gs={myblue},gt={mypurple},sae={myyellow},dune={mypink}},
    mark=triangle*, mark size=2.2pt, opacity=0.85,
] table[meta=family,x=x,y=y] {
x y family
1.000000 0.651837 raw
0.780804 0.780176 gs
0.833652 0.748235 gs
0.876731 0.732429 gs
0.540204 0.807465 gt
0.640777 0.784734 gt
0.675103 0.767482 gt
0.959318 0.801548 sae
0.946808 0.751901 sae
0.943825 0.823262 sae
0.943467 0.826274 sae
0.999998 0.854571 dune
};

\addplot[
    only marks, scatter, scatter src=explicit symbolic,
    scatter/classes={raw={black},gs={myblue},gt={mypurple},sae={myyellow},dune={mypink}},
    mark=*, mark size=2.2pt, opacity=0.85,
] table[meta=family,x=x,y=y] {
x y family
1.000000 0.672926 raw
0.780804 0.779288 gs
0.833652 0.753623 gs
0.876731 0.739762 gs
0.540204 0.795660 gt
0.640777 0.777023 gt
0.675103 0.767755 gt
0.959318 0.805452 sae
0.946808 0.765228 sae
0.943825 0.816833 sae
0.943467 0.816200 sae
0.999998 0.854786 dune
};

\addplot[
    only marks, scatter, scatter src=explicit symbolic,
    scatter/classes={raw={black},gs={myblue},gt={mypurple},sae={myyellow},dune={mypink}},
    mark=square*, mark size=2.2pt, opacity=0.85,
] table[meta=family,x=x,y=y] {
x y family
1.000000 0.666071 raw
0.780804 0.757380 gs
0.833652 0.746351 gs
0.876731 0.737143 gs
0.540204 0.762811 gt
0.640777 0.755718 gt
0.675103 0.755003 gt
0.959318 0.784238 sae
0.946808 0.761729 sae
0.943825 0.785286 sae
0.943467 0.784427 sae
0.999998 0.834571 dune
};
\end{axis}
\end{tikzpicture}
\end{sansmath}
  \end{minipage}
    \caption{\textbf{AutoInterp versus reconstruction on Gemma 3 270M-IT, layer~16.} Diamonds, triangles, circles, and squares denote the original evaluation (\colordiamond{black}) and activation thresholds of 25\% (\colortriangle{black}), 50\%(\colorcircle{black}), and 75\%(\colorsquare{black}), respectively. Colors identify the backbone-only reference (\colorcircle{black}), GemmaScope2 SAEs (\colorcircle{myblue}), transcoders (\colorcircle{mypurple}), self-trained SAEs (\colorcircle{myyellow}), and \ours (\colorcircle{mypink}). Each metric and threshold independently samples eligible features, so plotted points need not represent the same features.}
  \label{fig:appendix:llm_threshold_ai}
\end{figure*}

\begin{figure*}[t]
  \centering
  \begin{minipage}{0.70\textwidth}
    \begin{sansmath}
\begin{tikzpicture}
\scriptsize
\sffamily
\begin{axis}[
    width=1\linewidth,
    height=0.68\linewidth,
    xmin=0.50, xmax=1.04,
    ymin=0.10, ymax=0.50,
    xlabel={Faithfulness ($\uparrow$)},
    ylabel={Intruder ($\uparrow$)},
    xtick={0.5,0.6,0.7,0.8,0.9,1.0},
    ytick={0.1,0.2,0.3,0.4,0.5},
    label style={font=\scriptsize},
    axis line style={thick},
    enlargelimits=false,
    clip=false,
    axis background/.style={fill=gray!5,rounded corners=1pt},
    grid=both,
    major grid style={gray!30},
    minor grid style={gray!15},
    axis lines=left,
    trim axis left,
    trim axis right,
]
\addplot[
    only marks, scatter, scatter src=explicit symbolic,
    scatter/classes={raw={black},gs={myblue},gt={mypurple},sae={myyellow},dune={mypink}},
    mark=diamond*, mark size=2.2pt, opacity=0.85,
] table[meta=family,x=x,y=y] {
x y family
1.000000 0.203183 raw
0.780804 0.290545 gs
0.833652 0.258573 gs
0.876731 0.237132 gs
0.540204 0.323329 gt
0.640777 0.297010 gt
0.675103 0.269982 gt
0.959318 0.236784 sae
0.946808 0.244069 sae
0.943825 0.284121 sae
0.943467 0.278312 sae
0.999998 0.207991 dune
};

\addplot[
    only marks, scatter, scatter src=explicit symbolic,
    scatter/classes={raw={black},gs={myblue},gt={mypurple},sae={myyellow},dune={mypink}},
    mark=triangle*, mark size=2.2pt, opacity=0.85,
] table[meta=family,x=x,y=y] {
x y family
1.000000 0.201122 raw
0.780804 0.310810 gs
0.833652 0.273599 gs
0.876731 0.259792 gs
0.540204 0.344129 gt
0.640777 0.311623 gt
0.675103 0.283354 gt
0.959318 0.299576 sae
0.946808 0.244982 sae
0.943825 0.344310 sae
0.943467 0.346690 sae
0.999998 0.288628 dune
};

\addplot[
    only marks, scatter, scatter src=explicit symbolic,
    scatter/classes={raw={black},gs={myblue},gt={mypurple},sae={myyellow},dune={mypink}},
    mark=*, mark size=2.2pt, opacity=0.85,
] table[meta=family,x=x,y=y] {
x y family
1.000000 0.204850 raw
0.780804 0.344101 gs
0.833652 0.313262 gs
0.876731 0.299283 gs
0.540204 0.364793 gt
0.640777 0.337198 gt
0.675103 0.311120 gt
0.959318 0.353145 sae
0.946808 0.274413 sae
0.943825 0.390668 sae
0.943467 0.392910 sae
0.999998 0.347285 dune
};

\addplot[
    only marks, scatter, scatter src=explicit symbolic,
    scatter/classes={raw={black},gs={myblue},gt={mypurple},sae={myyellow},dune={mypink}},
    mark=square*, mark size=2.2pt, opacity=0.85,
] table[meta=family,x=x,y=y] {
x y family
1.000000 0.224855 raw
0.780804 0.376663 gs
0.833652 0.370082 gs
0.876731 0.377939 gs
0.540204 0.403296 gt
0.640777 0.386302 gt
0.675103 0.364041 gt
0.959318 0.385210 sae
0.946808 0.318641 sae
0.943825 0.424725 sae
0.943467 0.426250 sae
0.999998 0.383064 dune
};
\end{axis}
\end{tikzpicture}
\end{sansmath}
  \end{minipage}
  \caption{\textbf{Intruder versus reconstruction on Gemma 3 270M-IT, layer~16.} Colors and markers follow \cref{fig:appendix:llm_threshold_ai}.}
  \label{fig:appendix:llm_threshold_int}
\end{figure*}

\begin{figure*}[t]
  \centering
  \begin{minipage}{0.70\textwidth}
    \begin{sansmath}
\begin{tikzpicture}
\scriptsize
\sffamily
\begin{axis}[
    width=1\linewidth,
    height=0.68\linewidth,
    xmin=0.50, xmax=1.04,
    ymin=0.45, ymax=0.95,
    xlabel={Faithfulness ($\uparrow$)},
    ylabel={Embedding ($\uparrow$)},
    xtick={0.5,0.6,0.7,0.8,0.9,1.0},
    ytick={0.5,0.6,0.7,0.8,0.9},
    label style={font=\scriptsize},
    axis line style={thick},
    enlargelimits=false,
    clip=false,
    axis background/.style={fill=gray!5,rounded corners=1pt},
    grid=both,
    major grid style={gray!30},
    minor grid style={gray!15},
    axis lines=left,
    trim axis left,
    trim axis right,
]
\addplot[
    only marks, scatter, scatter src=explicit symbolic,
    scatter/classes={raw={black},gs={myblue},gt={mypurple},sae={myyellow},dune={mypink}},
    mark=diamond*, mark size=2.2pt, opacity=0.85,
] table[meta=family,x=x,y=y] {
x y family
1.000000 0.669097 raw
0.780804 0.827599 gs
0.833652 0.801716 gs
0.876731 0.756091 gs
0.540204 0.847095 gt
0.640777 0.807017 gt
0.675103 0.761261 gt
0.959318 0.787260 sae
0.946808 0.713285 sae
0.943825 0.830117 sae
0.943467 0.829752 sae
0.999998 0.695018 dune
};

\addplot[
    only marks, scatter, scatter src=explicit symbolic,
    scatter/classes={raw={black},gs={myblue},gt={mypurple},sae={myyellow},dune={mypink}},
    mark=triangle*, mark size=2.2pt, opacity=0.85,
] table[meta=family,x=x,y=y] {
x y family
1.000000 0.521753 raw
0.780804 0.839353 gs
0.833652 0.814640 gs
0.876731 0.783219 gs
0.540204 0.852462 gt
0.640777 0.815683 gt
0.675103 0.773939 gt
0.959318 0.829992 sae
0.946808 0.624519 sae
0.943825 0.848252 sae
0.943467 0.851023 sae
0.999998 0.680785 dune
};

\addplot[
    only marks, scatter, scatter src=explicit symbolic,
    scatter/classes={raw={black},gs={myblue},gt={mypurple},sae={myyellow},dune={mypink}},
    mark=*, mark size=2.2pt, opacity=0.85,
] table[meta=family,x=x,y=y] {
x y family
1.000000 0.547172 raw
0.780804 0.856117 gs
0.833652 0.840345 gs
0.876731 0.816350 gs
0.540204 0.862837 gt
0.640777 0.833647 gt
0.675103 0.795739 gt
0.959318 0.860986 sae
0.946808 0.670283 sae
0.943825 0.864904 sae
0.943467 0.866402 sae
0.999998 0.719783 dune
};

\addplot[
    only marks, scatter, scatter src=explicit symbolic,
    scatter/classes={raw={black},gs={myblue},gt={mypurple},sae={myyellow},dune={mypink}},
    mark=square*, mark size=2.2pt, opacity=0.85,
] table[meta=family,x=x,y=y] {
x y family
1.000000 0.608231 raw
0.780804 0.868975 gs
0.833652 0.851978 gs
0.876731 0.832127 gs
0.540204 0.858929 gt
0.640777 0.840912 gt
0.675103 0.816422 gt
0.959318 0.887522 sae
0.946808 0.744762 sae
0.943825 0.887959 sae
0.943467 0.889721 sae
0.999998 0.772874 dune
};
\end{axis}
\end{tikzpicture}
\end{sansmath}
  \end{minipage}
  \caption{\textbf{Embedding versus reconstruction on Gemma 3 270M-IT, layer~16.} Colors and markers follow \cref{fig:appendix:llm_threshold_ai}.}
  \label{fig:appendix:llm_threshold_emb}
\end{figure*}

\begin{table*}[t]
  \caption{\textbf{Monosemanticity \vsiclr~faithfulness on Gemma 3 270M-IT, layer~13.} We repeat the comparison with similar numbers of features at another layer. AI denotes AutoInterp, INT Intruder, EMB Embedding, and SIM Simulation. R$^2$ and next-token KL measure faithfulness. We report means and standard deviations across evaluation seeds 42, 43, and 44. KL is multiplied by 100 for easier reading.}
  \label{tab:appendix:gemma12}
  \centering
  \setlength{\tabcolsep}{5.0pt}
  \scriptsize
  \begin{tabularx}{\textwidth}{@{}rXS[table-format=1.3]S[table-format=1.3]S[table-format=1.3]S[table-format=1.3]S[table-format=1.3]S[table-format=1.3]@{}l@{}}
    \toprule
    & & \multicolumn{4}{c}{\bfseries Monosemanticity} & \multicolumn{2}{c}{\bfseries Faithfulness}\\
    \cmidrule(lr){3-6}\cmidrule(l){7-8}
    {\bfseries Model} & {\bfseries Method} &
    {\bfseries AI $(\uparrow)$} &
    {\bfseries INT $(\uparrow)$} &
    {\bfseries EMB $(\uparrow)$} &
    {\bfseries SIM $(\uparrow)$} &
    {\bfseries R$^2$ $(\uparrow)$} &
    {\bfseries KL $(\downarrow)$}\\
    \midrule

    \parbox[t]{2mm}{\multirow{15}{*}{\rotatebox[origin=c]{90}{Gemma 3 270M IT}}}
    & \textit{Backbone only} & \num{63.4}\tablestd{0.34} & \num{20.5}\tablestd{0.17} & \num{71.7}\tablestd{0.09} & \num{2.0}\tablestd{0.08} & \num{100.0}\tablestd{0.00} & \num{0.00}\tablestd{0.00} \\
    \cmidrule(l){2-8}
    & {\textcolor{black!55}{\textit{GemmaScope2}}} & \\
    & JumpReLU~\textcolor{black!55}{(Small)} & \num{80.2}\tablestd{0.08} & \num{32.1}\tablestd{0.53} & \num{82.4}\tablestd{0.31} & \num{6.8}\tablestd{0.16} & \num{80.6}\tablestd{0.05} & \num{11.99}\tablestd{0.10} \\
    & JumpReLU~\textcolor{black!55}{(Medium)} & \num{76.3}\tablestd{0.21} & \num{28.7}\tablestd{0.35} & \num{80.0}\tablestd{0.18} & \num{5.7}\tablestd{0.54} & \num{85.2}\tablestd{0.04} & \num{10.72}\tablestd{0.14} \\
    & JumpReLU~\textcolor{black!55}{(Large)} & \num{73.3}\tablestd{0.52} & \num{26.3}\tablestd{0.21} & \num{74.4}\tablestd{0.06} & \num{5.4}\tablestd{0.36} & \num{88.7}\tablestd{0.03} & \num{8.70}\tablestd{0.12} \\
    & Transcoder~\textcolor{black!55}{(Small)} & \num{81.9}\tablestd{0.23} & \num{34.1}\tablestd{0.85} & \num{83.0}\tablestd{0.28} & \num{6.4}\tablestd{0.35} & \num{73.6}\tablestd{0.08} & \num{25.92}\tablestd{0.29} \\
    & Transcoder~\textcolor{black!55}{(Medium)} & \num{78.2}\tablestd{0.28} & \num{30.9}\tablestd{0.21} & \num{79.9}\tablestd{0.14} & \num{5.9}\tablestd{0.52} & \num{77.4}\tablestd{0.05} & \num{24.42}\tablestd{0.27} \\
    & Transcoder~\textcolor{black!55}{(Large)} & \num{75.2}\tablestd{0.71} & \num{27.8}\tablestd{0.08} & \num{74.8}\tablestd{0.21} & \num{5.8}\tablestd{0.26} & \num{84.4}\tablestd{0.04} & \num{19.76}\tablestd{0.19} \\
    \cmidrule(l){2-8}
    & {\textcolor{black!55}{\textit{Self-Trained SAEs at EF25}}} & \\
    & Matching Pursuit SAE & \num{77.1}\tablestd{0.72} & \num{24.5}\tablestd{0.03} & \num{78.8}\tablestd{0.08} & \num{9.4}\tablestd{0.08} & \num{94.7}\tablestd{0.02} & \num{9.37}\tablestd{0.08} \\
    & Archetypal SAE & \num{80.0}\tablestd{0.37} & \num{31.5}\tablestd{0.21} & \num{81.2}\tablestd{0.40} & \num{2.9}\tablestd{0.25} & \num{85.8}\tablestd{0.16} & \num{25.46}\tablestd{0.32} \\
    & Matryoshka SAE & \num{81.1}\tablestd{0.74} & \num{29.5}\tablestd{0.27} & \num{80.8}\tablestd{0.23} & \num{10.0}\tablestd{0.52} & \num{92.6}\tablestd{0.08} & \num{13.68}\tablestd{0.15} \\
    & BatchTopK-SAE & \num{81.8}\tablestd{0.18} & \num{28.9}\tablestd{0.14} & \num{80.5}\tablestd{0.15} & \num{10.1}\tablestd{0.25} & \num{92.6}\tablestd{0.10} & \num{13.49}\tablestd{0.15} \\
    \cmidrule(l){2-8}
    & \tikz[remember picture,baseline] \node[inner sep=0pt,outer sep=0pt,minimum size=0pt] (llmgemma13beg) {};\ours \textit{(ours)} & \num{74.8}\tablestd{0.33} & \num{21.6}\tablestd{0.13} & \num{74.7}\tablestd{0.34} & \num{8.1}\tablestd{0.43} & \num{100.0}\tablestd{0.00} & \num{0.00}\tablestd{0.00} & {\rlap{\tikz[remember picture,baseline] \node[inner sep=0pt,outer sep=0pt,minimum size=0pt] (llmgemma13end) {};}} \\
    \bottomrule
  \end{tabularx}
  \begin{tikzpicture}[remember picture,overlay]
    \begin{scope}[on background layer]
      \fill[black,opacity=0.10]
        ([xshift=0,yshift=2.7ex]llmgemma13beg.north west)
        rectangle
        ([xshift=0,yshift=-1.4ex]llmgemma13end.south east);
    \end{scope}
  \end{tikzpicture}
\end{table*}

\begin{table*}[t]
  \caption{\textbf{Monosemanticity \vsiclr~faithfulness on Qwen3 0.6B, layer~21.} We repeat the comparison with similar numbers of features at another layer. AI denotes AutoInterp, INT Intruder, EMB Embedding, and SIM Simulation. R$^2$ and next-token KL measure faithfulness. We report means and standard deviations across evaluation seeds 42, 43, and 44. KL is multiplied by 100 for easier reading.}
  \label{tab:appendix:qwen21}
  \centering
  \setlength{\tabcolsep}{5.0pt}
  \scriptsize
  \begin{tabularx}{\textwidth}{@{}rXS[table-format=1.3]S[table-format=1.3]S[table-format=1.3]S[table-format=1.3]S[table-format=1.3]S[table-format=1.3]@{}l@{}}
    \toprule
    & & \multicolumn{4}{c}{\bfseries Monosemanticity} & \multicolumn{2}{c}{\bfseries Faithfulness}\\
    \cmidrule(lr){3-6}\cmidrule(l){7-8}
    {\bfseries Model} & {\bfseries Method} &
    {\bfseries AI $(\uparrow)$} &
    {\bfseries INT $(\uparrow)$} &
    {\bfseries EMB $(\uparrow)$} &
    {\bfseries SIM $(\uparrow)$} &
    {\bfseries R$^2$ $(\uparrow)$} &
    {\bfseries KL $(\downarrow)$}\\
    \midrule

    \parbox[t]{2mm}{\multirow{8}{*}{\rotatebox[origin=c]{90}{Qwen3 0.6B}}}
    & \textit{Backbone only} & \num{64.5}\tablestd{0.53} & \num{21.6}\tablestd{0.10} & \num{69.8}\tablestd{0.10} & \num{0.0}\tablestd{0.08} & \num{100.0}\tablestd{0.00} & \num{0.00}\tablestd{0.00} \\
    \cmidrule(l){2-8}
    & {\textcolor{black!55}{\textit{Self-Trained SAEs at EF25}}} & \\
    & Matching Pursuit SAE & \num{78.3}\tablestd{0.27} & \num{25.8}\tablestd{0.24} & \num{58.4}\tablestd{0.23} & \num{7.1}\tablestd{0.51} & \num{80.4}\tablestd{0.08} & \num{2.72}\tablestd{0.02} \\
    & Archetypal SAE & \num{69.8}\tablestd{0.26} & \num{24.3}\tablestd{0.10} & \num{58.3}\tablestd{0.26} & \num{-0.2}\tablestd{0.38} & \num{54.7}\tablestd{1.09} & \num{7.71}\tablestd{0.06} \\
    & Matryoshka SAE & \num{73.4}\tablestd{0.37} & \num{25.9}\tablestd{0.49} & \num{60.1}\tablestd{0.10} & \num{1.2}\tablestd{0.27} & \num{72.5}\tablestd{0.36} & \num{4.08}\tablestd{0.06} \\
    & BatchTopK-SAE & \num{73.2}\tablestd{0.40} & \num{25.4}\tablestd{0.19} & \num{59.1}\tablestd{0.27} & \num{1.8}\tablestd{0.14} & \num{72.9}\tablestd{0.37} & \num{3.98}\tablestd{0.06} \\
    \cmidrule(l){2-8}
    & \tikz[remember picture,baseline] \node[inner sep=0pt,outer sep=0pt,minimum size=0pt] (llmqwen21beg) {};\ours \textit{(ours)} & \num{74.4}\tablestd{0.55} & \num{22.6}\tablestd{0.16} & \num{68.5}\tablestd{0.58} & \num{4.9}\tablestd{0.42} & \num{100.0}\tablestd{0.00} & \num{0.00}\tablestd{0.00} & {\rlap{\tikz[remember picture,baseline] \node[inner sep=0pt,outer sep=0pt,minimum size=0pt] (llmqwen21end) {};}} \\
    \bottomrule
  \end{tabularx}
  \begin{tikzpicture}[remember picture,overlay]
    \begin{scope}[on background layer]
      \fill[black,opacity=0.10]
        ([xshift=0,yshift=2.7ex]llmqwen21beg.north west)
        rectangle
        ([xshift=0,yshift=-1.4ex]llmqwen21end.south east);
    \end{scope}
  \end{tikzpicture}
\end{table*}

\subsection{Qualitative results.}
We illustrate three Gemma layer-16 neurons and the identified subunits for which our AutoInterp evaluation saved descriptions and activating passages (\cref{tab:appendix:llm_qual_139,tab:appendix:llm_qual_331,tab:appendix:llm_qual_543}). The original neurons have AutoInterp scores of $0.64$, $0.79$, and $0.93$ for neurons 139, 331, and 543, respectively. This \ours run contains 24 subunits for neuron 139, 19 for neuron 331, and 24 for neuron 543. AutoInterp sampled features rather than exhaustively describing all of them; it evaluated eight, four, and six of these subunits, respectively. We show all of those evaluated subunits.
\begin{table*}[t]
  \caption{\textbf{Qualitative examples for Gemma layer-16 neuron 139.} The neuron description comes from the backbone-only AutoInterp evaluation; subunits descriptions come from the \ours evaluation. We show all eight subunits with saved descriptions and activating passages, out of 24 subunits in the checkpoint. Passages are shortened for readability; \ldots mark omitted context.}
  \label{tab:appendix:llm_qual_139}
  \centering
  \scriptsize
  \setlength{\tabcolsep}{5pt}
  \begin{tabularx}{\textwidth}{@{}lXX@{}}
    \toprule
    \textbf{Neuron ID} & \textbf{Generated feature description} & \textbf{Activating passage excerpt} \\
    \midrule
    Neuron 139 & text describing software development and troubleshooting & \texttt{display: flex; align-items: center}\newline ``Look, this is how it goes\ldots{} You attack me one at a time''\newline ``That's just how it was. There were no secrets on Main Street.'' \\
    \midrule
    Subunit 3742 & discussions of identity and social concepts like whiteness & ``whiteness becomes an object through the expression of anxiety''\newline ``Calling for whiteness to be seen\ldots''\newline ``whiteness studies as a political project'' \\
    Subunit 3745 & programming and scripting related terms and file paths & \texttt{\#!/usr/bin/expect -f}\newline \texttt{\#!/usr/bin/env python}\newline \texttt{\#!/usr/bin/env ruby} \\
    Subunit 3748 & discussions of the Middle East and related political or social issues & ``Egypt is key for a stable Middle East''\newline ``public safety initiatives in Latin America and the Middle East''\newline ``ISIL\ldots{} seized large parts of both Middle East countries'' \\
    Subunit 3749 & words related to biological processes and scientific terminology & ``cytokines (e.g. granulocyte-macrophage colony-stimulating factor)''\newline ``Hermaphrodites were considered gravid if they retained an embryonic\ldots''\newline ``reproductive output, fecundity and embryo volume'' \\
    Subunit 3751 & the concept of sequential or consecutive items or events & ``row-consecutive $p\times p$-minors''\newline ``a sequence of integers''\newline ``a mandatory consecutive term of 120 months'' \\
    Subunit 3755 & code elements like HTML tags and CSS classes within web development contexts & \texttt{\textless div class="navbar-collapse\ldots"\textgreater}\newline \texttt{\textless div class="collapse navbar-collapse"\ldots\textgreater}\newline \texttt{\textless div id="synopsis"\ldots\textgreater} \\
    Subunit 3756 & text related to font styles sizes and colors in web design or programming & \texttt{font: 10pt "Tw Cen MT Condensed Extra Bold"}\newline \texttt{font: 12pt "Niagara Engraved"}\newline \texttt{font: 14pt "Onyx"} \\
    Subunit 3760 & dates and times often associated with error messages or system logs & \texttt{Aug 30 06:36 backdoored-whois.deb}\newline \texttt{Mon Oct 17 09:44:39 UTC}\newline \texttt{[Tue Aug 16 16:26:11 2011] [error\ldots} \\
    \bottomrule
  \end{tabularx}
\end{table*}

\begin{table*}[t]
  \caption{\textbf{Qualitative examples for Gemma layer-16 neuron 331.} The neuron description comes from the backbone-only AutoInterp evaluation; subunit descriptions come from the \ours evaluation. We show all four subunits with saved descriptions and activating passages, out of 19 subunits in the checkpoint. Passages are shortened for readability; \ldots mark omitted context.}
  \label{tab:appendix:llm_qual_331}
  \centering
  \scriptsize
  \setlength{\tabcolsep}{5pt}
  \begin{tabularx}{\textwidth}{@{}lXX@{}}
    \toprule
    \textbf{Neuron ID} & \textbf{Generated feature description} & \textbf{Activating passage excerpt} \\
    \midrule
    Neuron 331 & text describing chemical processes and production & ``methanol vapors. Said vapors are toxic\ldots''\newline ``induction of the senescent physiological state\ldots{} knocking-down lncRNA-OIS1''\newline ``Response time can be an indicator of whether SQL query case is correct'' \\
    \midrule
    Subunit 8573 & technical text related to software development and error messages & \texttt{self::IF\_TYPE\_EPLRS}\newline \texttt{self::IF\_TYPE\_EON}\newline \texttt{ImportError: Could not import settings} \\
    Subunit 8575 & text related to XML encoding and UTF-8 character encoding & \texttt{<?xml version="1.0" encoding="UTF-8"?>}\newline \texttt{<xs:schema xmlns:\ldots}\newline \texttt{encoding='utf-8'} \\
    Subunit 8581 & musical artists, bands, albums, and songs, particularly debut releases and covers & ``Bikini Kill's self-titled Kill Rock Stars debut EP will be reissued''\newline ``debut album for BMG/RCA''\newline ``Madness later recorded the song for their cover album'' \\
    Subunit 8583 & scientific text describing cell types and biological processes & ``lncRNA-OIS1 in primary BJ cells''\newline ``embryonic fibroblast\ldots{} and keratinocyte''\newline ``senescence in WI-38 cells'' \\
    \bottomrule
  \end{tabularx}
\end{table*}

\begin{table*}[t]
  \caption{\textbf{Qualitative examples for Gemma layer- neuron 543.} The neuron description comes from the backbone-only AutoInterp evaluation; subunit descriptions come from the \ours evaluation. We show all six subunits with saved descriptions and activating passages, out of 24 subunits in the checkpoint. Passages are shortened for readability; \ldots mark omitted context.}
  \label{tab:appendix:llm_qual_543}
  \centering
  \scriptsize
  \setlength{\tabcolsep}{5pt}
  \begin{tabularx}{\textwidth}{@{}lXX@{}}
    \toprule
    \textbf{Neuron ID} & \textbf{Generated feature description} & \textbf{Activating passage excerpt} \\
    \midrule
    Neuron 543 & mathematical limits and infinity & ``$\ln 2 = \lim_{x\to 1^-}\ln(1+\cdots)$''\newline ``Roscoe's House of Chicken and Waffles and Dirty Bird's cornbreads\ldots''\newline ``Simpson ended up on the federal no-fly list'' \\
    \midrule
    Subunit 14088 & words related to distress or sounds of discomfort & ``her horrific, ear-splitting wail''\newline ``a loud, wailing howl'' \\
    Subunit 14101 & sports statistics and scores often involving hyphenated numbers & ``produced eight points during the run, with two long range triples''\newline ``completed just 52.5 percent of his passes''\newline ``4-for-4 from the charity stripe'' \\
    Subunit 14104 & measurements and dimensions expressed as numbers and units & ``dimensions of \texttt{3.875" x 2.5}\ldots''\newline ``an average width of 35 metres''\newline ``4 1/4 inches wide'' \\
    Subunit 14105 & code related to public class definitions and methods & \texttt{public class Problem\_16 \{ public long sumOfDigits(\ldots) \}}\newline \texttt{public class Hey<T>}\newline \texttt{public class Travels} \\
    Subunit 14107 & copyright notices and associated dates & \texttt{Copyright 2018--2020 Polyaxon}\newline \texttt{Copyright 2013 Google Inc.}\newline \texttt{Copyright 2015 go-termios} \\
    Subunit 14108 & mathematical notation and terminology & ``$\cdots^{d_1(j)} \times \mathbb{RP}^{d_2(j)}$ of the subspace''\newline ``$H^{2k-2}(N^{n-2};\mathbb{Z}/2)$''\newline ``$\Sigma_h^{\mathrm{reg}}$'' \\
    \bottomrule
  \end{tabularx}
\end{table*}

The neurons respond to several kinds of text that are difficult to capture with one description. Neuron 139 responds to both web styling and conversational prose; neuron 331 responds to chemical, biological, and software-related passages; and neuron 543 responds to mathematical notation as well as everyday prose. Their displayed subunits show more specific recurring patterns, including web styling, music releases, cell biology, sports scores, and mathematical notation. As in the vision examples, the split makes patterns within an original neuron easier to inspect.

\subsection{Conclusion}

To summarize, these experiments show that \ours %
can be readily applied to language models without adjustments to its algorithm. The same lossless construction applies beyond vision, preserving the model's next-token predictions. In particular at higher activation threshold evaluations, \ours achieves strong monosemanticity results.

\section{Implementation Details}\label{sec:appendix:impl_details}

\subsection{Sparse autoencoders}\label{sec:appendix:hp_sweep}

\paragraph{Sparsity target and training budget.} For each SAE and transcoder family, the target sparsity $K$ (active features per token) is fixed at $20$, following established practice in the SAE literature~\citep{Costa:2025:FFH, Pach:2025:SAL, Zaigrajew:2025:ICH}; Archetypal SAE uses a sparsity ratio of $10\%$ as in~\citet{Fel:2025:ASA}. All baselines are trained for $50$ epochs~\citep{Costa:2025:FFH, Fel:2025:ASA}.

\paragraph{Per-family grid sweep.} At the fixed sparsity above, we sweep the learning rate $\lambda \in \{10^{-4}, 5 \cdot 10^{-4}, 10^{-3}\}$ and the dead-feature regularization weight $w_{\mathrm{dead}} \in \{10^{-5}, 10^{-4}, 10^{-3}\}$ (which scales a re-animation term $\frac{1}{|\mathcal{F}_{\varnothing}|\,B}\sum_{s}\sum_{k\in\mathcal{F}_{\varnothing}}\tilde{z}_{sk}$ subtracted from the loss, where $\mathcal{F}_{\varnothing}$ denotes batch-dead features and $\tilde{z}_{sk}$ their pre-activations; maximizing this term encourages dormant features to activate) in all families. This dead-feature regularization is adapted from the \texttt{overcomplete} library. The Vanilla SAE additionally varies the $\ell_1$-sparsity weight $w_{\ell_1} \in \{10^{-3}, 5 \cdot 10^{-3}, 2 \cdot 10^{-2}\}$, and the JumpReLU SAE varies its sparsity-penalty weight $w_{\mathrm{sp}} \in \{10^{-4}, 10^{-3}, 10^{-2}\}$, giving $3^3 = 27$ configurations each. The remaining families -- TopK-SAE, BatchTopK-SAE, Matching Pursuit, Matryoshka SAE, and Archetypal SAE -- sweep only $\lambda \times w_{\mathrm{dead}} = 9$ configurations. In total this yields $99$ configurations per (backbone, expansion factor). The Transcoder uses the Vanilla architecture and sweeps the same $27$-cell $\lambda \times w_{\ell_1} \times w_{\mathrm{dead}}$ grid as the Vanilla SAE. All sweep runs use a single fixed training seed and the first of the three seeded token datasets used in \cref{sec:experiments:interp_faith}; the configuration selected per family is then applied to all three runs.

\paragraph{Selection criterion.} For each candidate, we compute the equally-weighted average of normalized reconstruction quality and sparsity,
\begin{equation}
    s = \tfrac{1}{2}\,\bigl(1 - \widetilde{\mathrm{MSE}}\bigr) + \tfrac{1}{2}\,\bigl(1 - \widetilde{\mathrm{L}_0}\bigr),
\end{equation}
where $\widetilde{\mathrm{MSE}}$ and $\widetilde{\mathrm{L}_0}$ are min--max normalized within each family, and $\mathrm{L}_0$ is the mean number of active features per token, divided by the total feature count $\mathrm{ef} \cdot d$ before normalization. The configuration with the highest $s$ is selected per family.

\subsection{Linear baseline implementation}

We implement the linear baselines as frozen encoder--dictionary pairs, enabling their evaluation with the same reconstruction and unit-level metrics as the SAE baselines.

We fit PCA~\citep{Pearson:1901:OLA} using scikit-learn's \texttt{PCA} implementation with its covariance eigendecomposition solver. For evaluation, we extract the fitted components and mean into a frozen linear encoder--dictionary pair. PCA is deterministic for a fixed token dataset and therefore has no initialization seed; we refit it only for each of the three seeded token datasets.

We use the \texttt{Semi-NMF}~\citep{Ding:2010:CAS} implementation from the \texttt{overcomplete} library because the target-layer activations are non-negative. Semi-NMF~\citep{Ding:2010:CAS} learns non-negative codes with an unconstrained dictionary using projected-gradient updates. To encourage sparse codes, we add an $\ell_1$ penalty to the code matrix. Because Semi-NMF~\citep{Ding:2010:CAS} maintains a code matrix for all fitting tokens on the GPU, we use a seeded subsample of $400{,}000$ tokens to remain within GPU memory limits. We fit Semi-NMF~\citep{Ding:2010:CAS} at expansion factor $8$, sweep its learning rate and sparsity penalty, and select the configuration using the same equally weighted reconstruction--sparsity criterion as for the SAE baselines.

\subsection{Compute hardware}\label{sec:appendix:hardware}

Different tasks run on different NVIDIA GPUs. The three vision--language judges of \cref{sec:appendix:coherence} (Qwen3-VL-30B-A3B-Instruct, Gemma-3-27B, InternVL3-14B) are served via vLLM on H200 GPUs. \ours checkpoint generation -- top-$k$ activation collection, contribution-vector clustering, and weight restructuring -- runs on H100 or A100 GPUs; a single full-pipeline \ours run on DINOv2-B/14 layer~$12$ at top-$k = 400{,}000$ takes approximately $26$--$30$ hours on a single A100. This runtime reflects a conservative, unoptimized implementation: neurons are processed sequentially on a single GPU, top-$k$ is set to $400{,}000$ (the high end of our ablation range), and \texttt{min\_cluster\_size}$=1000$ is chosen for cluster quality -- reducing it substantially shortens runtime. The cost is one-time and offline; its effect on inference is minimal, as shown in \cref{sec:appendix:throughput_results}. SAE and transcoder training, together with the downstream evaluation pipeline (classification accuracy, C-Score, $R^2$ reconstruction), runs on L40S GPUs. The throughput benchmark runs exclusively on a single H100.

\subsection{Software stack}\label{sec:appendix:software}

The codebase is implemented in PyTorch (BSD 3-Clause License) with torchvision (MIT License). Model loading uses \texttt{transformers} (Apache License 2.0), \texttt{timm}~\citep{Wightman:2019:PIM} (Apache License 2.0), and \texttt{torch.hub}. Clustering uses \texttt{hdbscan}~\citep{Campello:2013:DBC} (MIT License), with \texttt{cuML} (RAPIDS) (Apache License 2.0) accelerating the same algorithm at scale (the source of the residual numerical non-determinism noted in \cref{sec:appendix:stability_results}); both are installed separately from the vision environment. PCA uses \texttt{scikit-learn}'s \texttt{PCA} implementation (BSD 3-Clause License). The SAE, Semi-NMF and transcoder baselines reuse the implementations in the \texttt{overcomplete} library (MIT License). Auxiliary dependencies include \texttt{scipy} (Hungarian matching; BSD 3-Clause License), \texttt{umap-learn} (one ablation axis; BSD 3-Clause License), and \texttt{sentence-transformers}~\citep{Reimers:2019:SBS} (Apache License 2.0). The VLM judges are served through \texttt{vLLM} (Apache License 2.0).

The language-model evaluations use the SAEBench repository~\citep{Karvonene:2025:SAC}, with \texttt{TransformerLens} for model hooks and \texttt{SAE Lens} for published SAE checkpoints. The embedding-based evaluation uses \texttt{sentence-transformers}. We implement the Simulation scoring procedure locally rather than importing OpenAI's automated-interpretability library.

\subsection{Datasets}

All methods use ImageNet-1k~\citep{Russakovsky:2015:ILS} (standard train/validation split) with each backbone's native preprocessing pipeline; ImageNet-1k is used under its research-only terms.

\subsection{Models and evaluation checkpoints}

\Cref{tab:appendix:backbones,tab:appendix:evaluation_models} list the studied models, published baselines, and evaluation models, with their upstream identifiers, loading sources, and stated licenses.

\begin{table*}[t]
  \caption{\textbf{Models where we apply \ours.} We list the upstream checkpoint identifier and loading source for each model, together with its stated license. The DINOv2 entry covers the Small, Base, and Large variants.}
  \label{tab:appendix:backbones}
  \centering
  \scriptsize
  \setlength{\tabcolsep}{4pt}
  \begin{tabularx}{\textwidth}{@{}p{0.22\textwidth}Xp{0.12\textwidth}p{0.15\textwidth}@{}}
    \toprule
    \textbf{Model} & \textbf{Identifier} & \textbf{Source} & \textbf{License} \\
    \midrule
    DINOv2-S/B/L~\citep{Oquab:2024:DLR} & \path{facebookresearch/dinov2} (\path{vits14}, \path{vitb14}, \path{vitl14}) & \texttt{torch.hub} & Apache-2.0 \\
    AugReg ViT-B/16~\citep{Dosovitskiy:2021:AII,Steiner:2022:HTT} & \path{vit_base_patch16_224.augreg_in21k_ft_in1k} & \texttt{timm} & Apache-2.0 \\
    ConvNeXtV2-Base~\citep{Woo:2023:CVC} & \path{convnextv2_base.fcmae_ft_in22k_in1k} & \texttt{timm} & CC BY-NC 4.0 \\
    SigLIP2-B/16~\citep{Tschannen:2025:SMV} & \path{google/siglip2-base-patch16-224} & Hugging Face & Apache-2.0 \\
    LLaVA-OneVision 0.5B~\citep{Li:2025:LOE} & \path{llava-hf/llava-onevision-qwen2-0.5b-ov-hf} & Hugging Face & Apache-2.0 \\
    Gemma 3 270M-IT~\citep{Gemma:2025:GTR} & \path{google/gemma-3-270m-it} & Hugging Face & Gemma terms \\
    Qwen3 0.6B~\citep{Qwen:2025:QTR} & \path{Qwen/Qwen3-0.6B} & Hugging Face & Apache-2.0 \\
    \bottomrule
  \end{tabularx}
\end{table*}

\begin{table*}[t]
  \caption{\textbf{Published baselines and models used to evaluate.} We list checkpoint identifiers, loading sources, and stated licenses. For CLIP, the license follows the upstream project repository; for InternVL3-14B, it follows the model card's License section.}
  \label{tab:appendix:evaluation_models}
  \centering
  \scriptsize
  \setlength{\tabcolsep}{4pt}
  \begin{tabularx}{\textwidth}{@{}p{0.22\textwidth}Xp{0.12\textwidth}p{0.15\textwidth}@{}}
    \toprule
    \textbf{Model} & \textbf{Identifier} & \textbf{Source} & \textbf{License} \\
    \midrule
    Gemma Scope 2 270M-IT~\citep{McDougall:2025:GST} & \path{google/gemma-scope-2-270m-it} & Hugging Face & CC BY 4.0 \\
    CLIP ViT-B/16~\citep{Radford:2021:LTV} & \path{openai/clip-vit-base-patch16} & Hugging Face & MIT \\
    SigLIP2-SO400M/14 (384 px)~\citep{Tschannen:2025:SMV} & \path{google/siglip2-so400m-patch14-384} & Hugging Face & Apache-2.0 \\
    Qwen3-VL-30B-A3B-Instruct~\citep{Qwen:2025:QVL} & \path{Qwen/Qwen3-VL-30B-A3B-Instruct} & Hugging Face & Apache-2.0 \\
    Gemma 3 27B-IT~\citep{Gemma:2025:GTR} & \path{google/gemma-3-27b-it} & Hugging Face & Gemma terms \\
    InternVL3-14B~\citep{Zhu:2025:IEA} & \path{OpenGVLab/InternVL3-14B} & Hugging Face & MIT \\
    Gemma 3 12B-IT~\citep{Gemma:2025:GTR} & \path{google/gemma-3-12b-it} & Hugging Face & Gemma terms \\
    Llama 3.1 8B~\citep{LLama:2024:TLH} & \path{meta-llama/Llama-3.1-8B} & Hugging Face & Llama 3.1 license \\
    all-MiniLM-L6-v2~\citep{Reimers:2019:SBS} & \path{sentence-transformers/all-MiniLM-L6-v2} & Hugging Face & Apache-2.0 \\
    \bottomrule
  \end{tabularx}
\end{table*}

\subsection{Per-backbone disentanglement configurations}

The 1-indexed layer numbers used throughout the paper map to 0-indexed indices in our codebase as $\text{layer\_1idx} - 1$ (e.g.\ ``layer 12'' on DINOv2-B/14 is layer index $11$).

\paragraph{Matching the SAE expansion factor.} For each experiment we first set the expansion factor for the SAE and transcoder baselines, then match it with \ours by adjusting either the TopK activation budget or the HDBSCAN \texttt{min\_cluster\_size}: smaller \texttt{min\_cluster\_size} yields more, smaller clusters and a higher effective expansion factor. SAEs and transcoders use a fixed expansion factor of $8$ unless specified otherwise.

\paragraph{Per-backbone settings.} \Cref{tab:appendix:per_backbone_configs} lists the TopK and clustering deviations from the default (\texttt{min\_cluster\_size}~$=$~$1000$, \texttt{cluster\_selection\_method=leaf}, \texttt{min\_samples} at the library default, no UMAP; \cref{sec:appendix:clustering_hp}) for each (backbone, layer) configuration evaluated in \cref{sec:experiments:interp_faith,tab:generalization}.

\begin{table}[h]
  \caption{\textbf{Per-backbone \ours configurations.} Default clustering means \texttt{min\_cluster\_size}~$=$~$1000$ with all other settings as in \cref{sec:appendix:clustering_hp}. Effective EF is the per-neuron cluster count summed across neurons divided by the layer width, averaged over the three seeded token datasets.}
  \label{tab:appendix:per_backbone_configs}
  \centering
  \scriptsize
  \begin{tabularx}{\textwidth}{@{}Xllll@{}}
    \toprule
    \textbf{Backbone} & \textbf{Layer} & \textbf{TopK} & \textbf{Clustering} & \textbf{Eff.\ EF}\\
    \midrule
    AugReg ViT-B/16     & 12   & 100k & default & 7.49\\
    DINOv2-B/14         & 12   & 400k & default & 8 \\
    DINOv2-B/14         & 10   & 400k & default & 8.01 \\
    DINOv2-B/14         & 11   & 200k & default & 5.91 \\
    DINOv2-S/14         & 12   & 400k & default & 8.1 \\
    DINOv2-L/14         & 23   & 400k & \texttt{min\_cluster\_size}~$=$~$500$ & 7.90 \\
    SigLIP2-B/16        & 12   & 200k & default & 7.51 \\
    ConvNeXtV2-B        & last linear layer of the fourth stage & 500k & default & 7.58 \\
    DINOv2-B/14 (EF4)   & 12   & 60k  & default & 3.96 \\
    DINOv2-B/14 (EF16)  & 12   & 400k & \texttt{min\_cluster\_size}~$=$~$750$ & 11.41 \\
    Gemma 3 270M-IT      & 13 & 100k & \texttt{min\_cluster\_size}$=250$, \texttt{eom} & 26.56 \\
    Gemma 3 270M-IT      & 16 & 100k & \texttt{min\_cluster\_size}$=250$, \texttt{eom} & 25.66 \\
    Qwen3 0.6B           & 21 & 100k & \texttt{min\_cluster\_size}$=250$, \texttt{eom} & 28.13 \\
    \bottomrule
  \end{tabularx}
\end{table}

\subsection{Clustering hyperparameters}\label{sec:appendix:clustering_hp}

\ours uses HDBSCAN~\citep{Campello:2013:DBC} with \texttt{cluster\_selection\_method=leaf}, \texttt{min\_cluster\_size}~$=1000$, \texttt{min\_samples} at the library default, ARV unit-normalization enabled, and no UMAP dimensionality reduction. The ablations in \cref{sec:appendix:ablations_results} sweep each of these one at a time around this baseline.

\subsection{Seeds}

For SAE and transcoder baselines, training initialization uses seed $42$ for the headline run and $\{7, 1024\}$ for the cross-seed stability evaluation in \cref{sec:experiments:applications}. The three \emph{seeded token datasets} of \cref{sec:experiments:interp_faith} use seeds $\{42, 123, 999\}$; these select random subsets of $2$ tokens per ImageNet image and are independent of the SAE training seed.

\end{document}